\documentclass[runningheads]{llncs}

\usepackage{eccv}

\usepackage{eccvabbrv}

\usepackage{graphicx}
\usepackage{gensymb}
\usepackage{booktabs}
\usepackage{amssymb}
\usepackage{mathtools}
\usepackage{enumitem}

\usepackage{tikz}
\usetikzlibrary{arrows.meta}

\usepackage{tabularray}
\UseTblrLibrary{booktabs}
\usepackage[export]{adjustbox}
\usepackage{rotating}
\usepackage{makecell}
\usepackage{textcomp}
\usepackage{multirow}
\usepackage{arydshln}
\usepackage[percent]{overpic}

\usepackage[accsupp]{axessibility}  %

\usepackage{hyperref}

\usepackage{orcidlink}

\newcommand{\PAR}[1]{\vskip4pt \noindent{\bf #1~}}

\newcommand{\bI}{\mathbf{I}}

\newcommand{\bo}{\mathbf{o}}

\newcommand{\bt}{\mathbf{t}}

\newcommand{\rdt}{r_{\Delta\bt}}
\newcommand{\normt}{\lVert\Delta\bt\rVert}

\newcommand{\pose}{\boldsymbol{\xi}}

\newcommand{\ceqq}{\coloneqq}

\newcommand{\twodots}{\mathinner{\ldotp\ldotp}}

\DeclareMathOperator{\norm}{norm}

\definecolor{pastelred}{RGB}{235,119,119}
\definecolor{pastelorange}{RGB}{255,202,126}

\makeatletter
\newcommand{\setword}[2]{%
  \phantomsection
  #1\def\@currentlabel{\unexpanded{#1}}\label{#2}%
}
\makeatother

\providetoggle{foobool}

\begin{document}

\newcommand{\name}{AutoCompass\xspace}
\title{\name: Accurate Visual Localization on Public Maps by Learning from Weak Labels}

\titlerunning{AutoCompass}

\author{Javier Tirado-Garín\inst{1}%
\thanks{Work done during an internship at Niantic Spatial.}%
\and
Alan Savio Paul\inst{2} \and
Shuai Chen\inst{2} \and
Axel Barroso-Laguna\inst{2} \and
Tommaso Cavallari\inst{2} \and
Daniyar Turmukhambetov\inst{2} \and
Victor Adrian Prisacariu\inst{2,3} \and
Eric Brachmann\inst{2}}

\authorrunning{J.~Tirado-Garín et al.}

\definecolor{urlteal}{HTML}{2F6F73}
\definecolor{urlblue}{HTML}{3A5F8A}
\definecolor{urlindigo}{HTML}{4B4E8A}
\definecolor{urlplum}{HTML}{6A4C70}
\definecolor{urltol}{HTML}{e25655}
\definecolor{urlgray}{HTML}{666666}
\institute{%
  \vspace{-0.1cm}
  \textsuperscript{1}I3A, Universidad de Zaragoza
  \quad
  \textsuperscript{2}Niantic Spatial
  \quad
  \textsuperscript{3}University of Oxford%
  \\
  \href{https://nianticspatial.github.io/autocompass/}{%
  \textcolor{urlteal}{\nolinkurl{https://nianticspatial.github.io/autocompass/}}}%
}

\maketitle

\begin{abstract}

Neural map matchers estimate an image's 3-DoF pose relative to a 2D map. These models are trained on large-scale datasets of geo-referenced images, whose position and heading labels often contain noise that affects the trained models. To address this, we present \name, a supervision approach for training neural map matchers from inaccurate absolute pose labels. 
First, we show that heading labels are unnecessary: trained from raw GPS labels, models learn to predict accurate headings, automatically. 
Second, defining a tolerance region around raw GPS improves positional accuracy. 
Third, if available, our supervision uses relative poses between training images, obtained via SLAM or SfM, which provide a more accurate training signal.
Across driving and egocentric benchmarks, \name consistently outperforms counterparts trained with the usual strong reliance on absolute pose labels.

\end{abstract}

\begin{figure}
    \centering
    \includegraphics[width=0.99\linewidth]{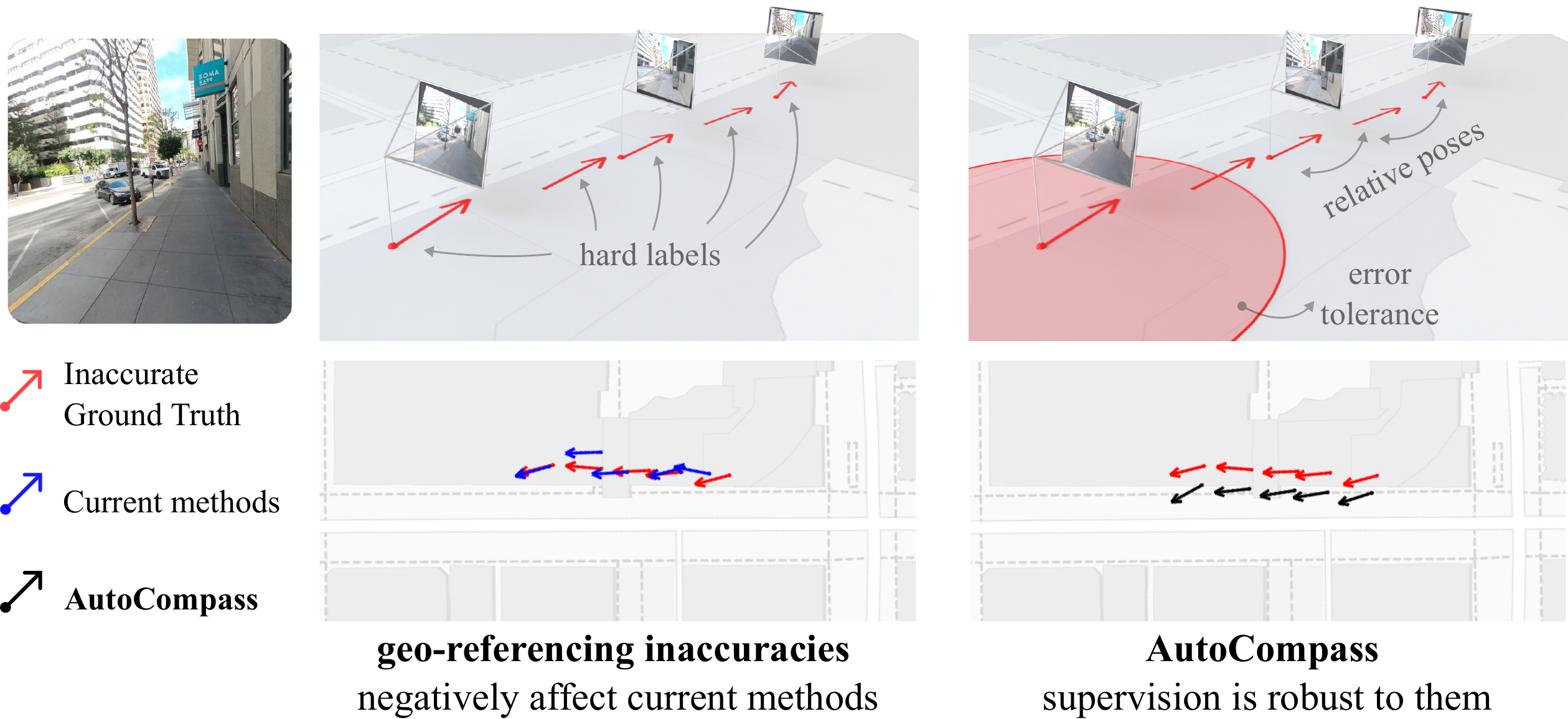}
    \vspace{-0.05cm}
    \caption{\textbf{Geo-localization datasets can be noisy.} Pseudo ground-truth (GT) poses of a training sequence from the MGL dataset \cite{sarlin2023orienternet} are offset by several meters: images captured on a sidewalk have poses inside a building. Methods trained with strong supervision, \eg \cite{sarlin2023orienternet}, learn and replicate these errors, while our proposed \name is robust to them. We instead treat the GT as \emph{weak} labels, modeling the unknown true pose as lying anywhere within a neighborhood of the GT label. We also learn from relative poses, which can be obtained from SLAM or SfM and are thus more accurate.}
    \label{fig:teaser}
\end{figure}

\section{Introduction}
\label{sec:intro}

To go anywhere, first we must know where we are. To find ourselves, we consult maps, often aided by satellite positioning systems such as GPS. 
GPS tells us where we are, at least up to a few meters under favorable conditions. However, its performance declines when signals are obstructed or deliberately jammed. In such cases, we can practice our archaic
skills of matching street names to maps.

Neural map matchers~\cite{sarlin2023orienternet, sarlin2023snap, Fervers2023uncertainty} offer an alternative: they infer a local bird's-eye-view (BEV) representation from an input image, and match it to a 2D map to estimate the 2D position and heading of the camera.
Compared to GPS, they are unaffected by multipath interference and signal jamming, and can be more accurate in urban environments, especially when estimates are fused sequentially.
Neural map matchers are also scalable: once trained, they rely on 2D maps, which providers, such as OpenStreetMap~\cite{OpenStreetMap}, make freely available for much of the planet.
This contrasts with 6-degree-of-freedom (DoF) visual localization~\cite{sarlin2019HFNet,Balntas18,brachmann2023ACE}, which typically relies on memory-intensive maps built from densely captured images, incurring high storage and computational costs at city scale and beyond.

However, one crucial limitation of neural map matchers is their reliance on large-scale geo-referenced training data. Prior work has leveraged hundreds of thousands of images annotated with absolute global positions and headings \cite{sarlin2023orienternet}. At this scale, annotation must be automated, making it inherently susceptible to noise and inaccuracies.
A common strategy is to reconstruct local image clusters using Structure-from-Motion (SfM)~\cite{schoenberger2016sfm, ullman1979interpretation, pan2024glomap, snavely2006}, and to fuse them with raw GPS measurements, assuming that random errors cancel out. Such pipelines do not address systematic biases and can be unreliable when image clusters are small, poorly conditioned, or sparsely connected. Consequently, publicly available geo-referenced image datasets inevitably contain pose inaccuracies, which negatively impact models trained on them (see example in Fig.~\ref{fig:teaser}).

With \name, we reduce neural map matchers' dependence on accurate geo-referenced camera pose labels for training.
We first remove the need for heading labels: we find that the inductive bias of existing architectures is enough for accurate heading predictions to emerge.
Next, to be robust to GT inaccuracies, we only assume that the true image position lies within a tolerance region of up to 20\,m around the GPS coordinates.
This makes our approach robust, \eg, to GPS errors of the magnitude commonly observed with receivers in egocentric devices~\cite{Krishnan2025lamaria,engel2023aria}.
Finally, we find that relative camera poses, despite providing only local (and thus weaker) constraints, significantly improve the accuracy of the downstream learned absolute pose distribution. While geo-referencing at scale remains as a challenging problem, relative poses can be reliably estimated via SfM or Simultaneous Localization And Mapping (SLAM)~\cite{davison2007monoslam, mur2015orb, campos2021orb3, mur2017orb2, boche2025okvis2, opensfm, Engel17}.

Despite relying on weaker supervision than previous approaches, \name comes out on top in terms of accuracy, due to the aforementioned issues with geo-referenced pseudo ground-truth. \name is agnostic to the underlying architecture and can be applied to existing architectures without modification.

\noindent We summarize our \textbf{contributions} as follows:
\begin{itemize}
    \item[$\circ$] An absolute pose loss that adds significant error tolerance to geo-referenced position labels, and does not require any geo-referenced heading labels.
    \item[$\circ$] Using our absolute pose loss, we demonstrate that a neural map matcher can be trained from unoptimized GPS position labels, alone.
    \item[$\circ$] Various relative pose losses tailored to practical scenarios, \eg non-metric poses or approximately geo-referenced ones. These losses significantly improve the accuracy of existing single-image supervised approaches
\end{itemize}

Our evaluation across multiple datasets shows that neural map matchers, trained with our proposed supervision, even when using only raw GPS labels, outperform baselines trained with strong supervision on geo-referenced image poses. 
Using relative pose supervision, \name sets a new state of the art in visual localization on 2D public maps.

\section{Related work}

\subsection{Ground-to-ground visual localization}
Traditional ground-to-ground visual relocalization operates in pre-mapped environments and relies on building and storing 3D representations from ground-level data. Existing methods range from multi-stage pipelines, such as combining image retrieval~\cite{Zamir10, Arandjelovic14, torii201524, arandjelovic2016netvlad, Weyand16, Gordo16, kapture2020, berton2022rethinking, izquierdo2024optimal, izquierdo2024close, berton2025megaloc,deng2026_unipr3d} with feature matching~\cite{sarlin2019HFNet, Lynen2019largescale, sarlin2020superglue, sarlin21pixloc,morlana2023reuse, barroso2024matching, panek2022meshloc, leroy2024grounding, edstedt2024roma, liu2025air, edstedt2025roma} or relative pose regression~\cite{Balntas18, dong2025reloc3r, deng2025reloc, barroso2026scene}, to more implicit approaches that rely on networks to memorize scenes, such as scene coordinate regression (SCR)~\cite{Shotton13, Cavallari_2017_CVPR, Brachmann17, Brachmann18, brachmann2020dsacstar, brachmann2023ACE, wang2024glace, wang2024hscnet++, jiang2025r,bian2025scrpriors} or absolute pose regression (APR)~\cite{Kendall15, Kendall16, Kendall17, Wu17, Naseer17, Brahmbhatt18, Shavit21multiscene, Moreau21, chen2022dfnet, Shavit22PAE}. While most ground-to-ground methods focus on estimating 6-DoF camera poses, they face inherent trade-offs between map scale, mapping and inference efficiency, and localization accuracy. For example, the most scalable structure-based solutions~\cite{sarlin2019HFNet, sattler2016efficient, Sattler12} require explicitly building and maintaining a large reference database via SfM, which imposes substantial storage requirements and increases matching complexity as the dataset grows. On the other hand, SCR~\cite{brachmann2023ACE, bruns2025ace} or APR methods~\cite{chen2024marepo} can efficiently map individual scenes and scale by training on large numbers of small-to-medium-sized scenes to increase map coverage. However, none of the existing solutions can avoid the substantial costs of collecting dense ground data. Another premise of ground-to-ground methods is that accurate mapping data is essential. Prior studies~\cite{brachmann2021limits, chen2024neural} have shown that the quality of the pseudo ground-truth highly impacts localization results. Differently, neural map matchers like OrienterNet \cite{sarlin2023orienternet} use lightweight maps (taking about 200KB for a 128x128m area) that are free and publicly available, and do not require densely captured ground-level mapping images. In this work, our contributions reduce the strong reliance of such methods on often noisy geo-referenced ground truth.

\subsection{Cross-view visual localization}

Cross-view visual localization estimates 3-DoF camera poses (\ie, 2D position and orientation) by localizing images in 2D map representations derived from aerial/satellite imagery or planimetric maps such as OpenStreetMap (OSM). Since these map sources are typically georeferenced and available at large scale, cross-view methods enable wide-area localization without collecting and building dense image-based maps.

Early work on 2D map-based visual localization focused on controlled environments, where the map structure is relatively simple and reliable. In indoor scenes, \eg, robust pose estimation can be achieved from floor plans \cite{howard2021lalaloc, howard2022lalaloc++, grader2025supercharging, Min_2022_CVPR, chen2024f3loc}. More recently, C3PO \cite{huang2025c3po} adapts DUSt3R point maps \cite{wang2024dust3r} for cross-view prediction between images and floor plans, further improving indoor localization performance. Beyond indoor settings, cross-view localization has also been explored in large-scale outdoor environments. Some methods target continental-scale localization \cite{lindenberger2025scaling, fervers2024statewide}, achieving errors on the order of hundreds of meters. A prominent line of work uses aerial or satellite imagery as the reference map for matching \cite{lentsch2023slicematch,Fervers2023uncertainty,xia2026loc,xia2025fg,sarlin2023snap}. While these methods can be highly robust and meter-level accurate, they often require substantial storage resources \cite{sarlin2023orienternet}.

To reduce this dependence on dense imagery, recent methods have investigated localization using lightweight, widely available 2D maps. A seminal OSM-based approach, OrienterNet \cite{sarlin2023orienternet}, performs neural matching by cross-correlating learned BEV features from the query image with learned map features. Following this direction, OSMLoc \cite{liao2024osmloc} leverages foundation models to inject strong semantic and geometric priors into its learning pipeline, while DiffVL \cite{gao2025diffvl} reformulates visual localization as a GPS-denoising task using diffusion models. Complementary approaches further improve accuracy by leveraging additional inputs at inference time, such as multiple images \cite{wu2024maplocnet,camiletto2024u} or LiDAR point clouds \cite{zhou2025seglocnet}.

\subsection{Weakly-supervised cross-view visual localization}

A major challenge in cross-view localization is the need for accurate GT pose labels. To reduce this dependence, several works have explored weakly supervised learning. C-BEV~\cite{fervers2023cbev} proposes a trainable retrieval system based on BEV features extracted from panoramic images, replacing traditional descriptor-based matching. It shows that camera pose estimation can emerge from 360\degree-BEV embeddings without explicit pose supervision. GeoDistill~\cite{tong2025geodistill} also requires panoramas, using a teacher-student framework in which a teacher model with access to full panoramas supervises a student learning from perspective crops.

Other works relax the supervision requirements under different assumptions. Shi \etal~\cite{shi2024weakly} weakly supervise the position of the ground-level camera, while assuming a strong orientation prior at both training and inference time. Xia \etal~\cite{xia2024adapting} address datasets with heterogeneous ground-truth quality, adapting training to account for varying levels of informativeness in the GT labels.

In contrast, \name assumes access only to perspective images at both training and inference time, together with noisy GPS measurements or relative poses for supervision. We show that this limited weak supervision is enough to outperform strongly supervised approaches.

\begin{figure}[t]
    \newcommand{\dotarrow}[2][blue]{%
  \mathrel{%
    \tikz[baseline=-0.6ex]{
      \draw[#1, line width=0.5pt, -{Stealth[length=1.6mm]}] (0,0) -- (#2,0);
      \fill[#1] (0,0) circle[radius=0.5mm];
    }%
  }%
}

    \centering
    \includegraphics[width=1.\linewidth]{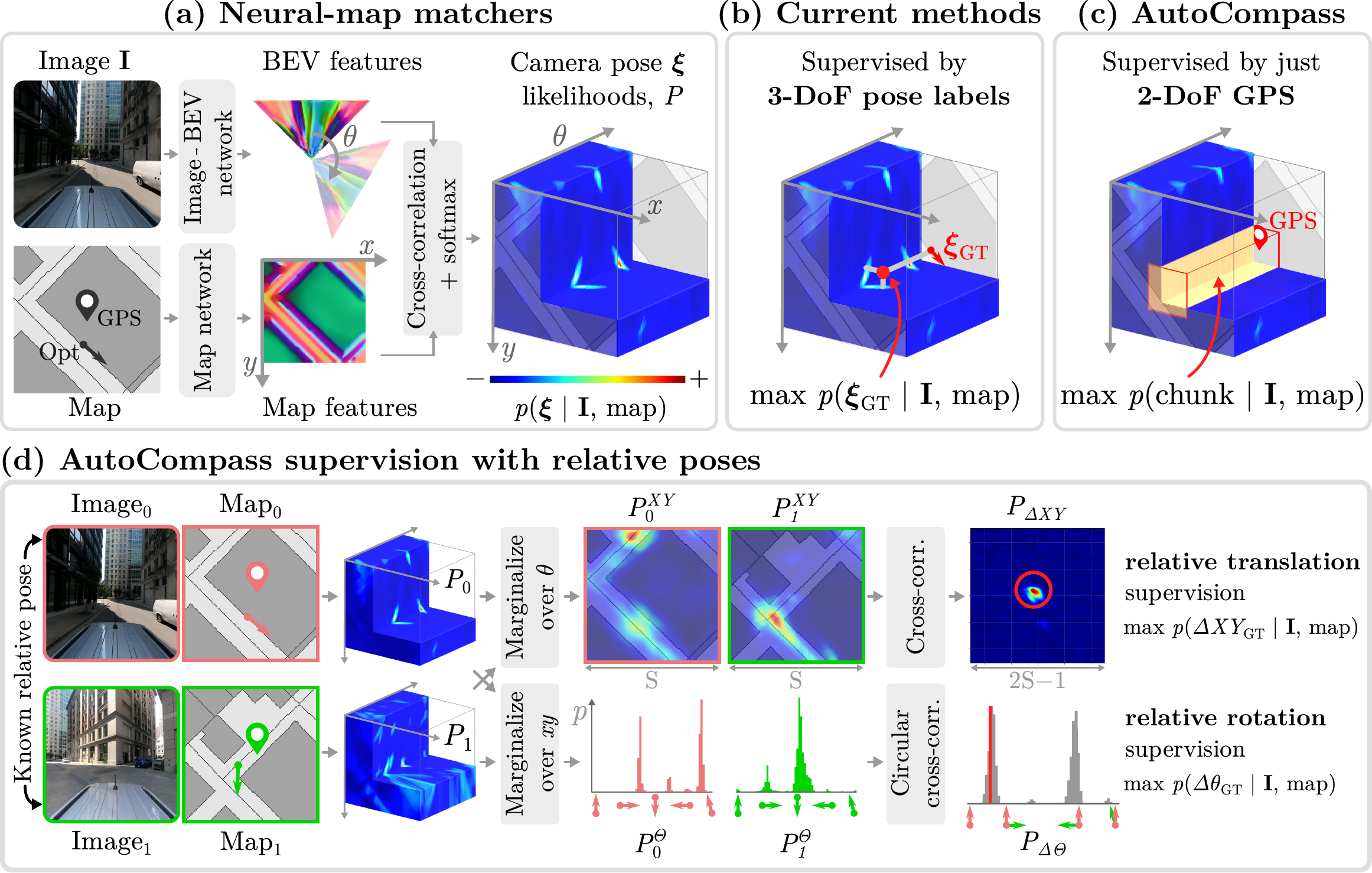}
    \caption{\textbf{Robust visual localization via weak supervision.}
    (a) We adopt neural-map matching \cite{sarlin2023orienternet} to estimate a distribution over discrete camera poses.
    (b) Current methods supervise this distribution under the assumption of accurate 3-DoF pose labels (Opt.), which can break when GPS-based geo-referencing is noisy, as in the example shown.
    (c, d) We instead propose weak-supervision strategies that are robust to such inaccuracies.
    The simplest strategy, shown in (c), requires only noisy 2-DoF GPS labels, as it maximizes the probability assigned to a GPS-centered spatial chunk, thus only assuming that the true pose lies within this chunk.
    (d) We also learn from relative poses obtained from SfM, which are robust to offsets in geo-referenced poses (see \cref{fig:losses_invariances}).}
    \label{fig:prob_dists}
\end{figure}

\section{Method}

\PAR{Task}
Given a gravity-aligned, calibrated query image $\bI$ and a rasterized $H \times W$ local map tile from OpenStreetMap~\cite{OpenStreetMap}, our goal is to estimate the 3-DoF pose $\pose = (x,y,\theta)$ corresponding to the image $\bI$ in the map. The pose consists of a planar position $(x,y) \in \mathbb{R}^2$ (expressed in the geo-referenced map-tile coordinate frame) and a heading $\theta \in [0, 360)\degree$. For simplicity, we assume a square tile and denote its size by $S$ (\ie, $H = W = S$). In our experiments, $S=128$\,m.

\PAR{Setting} We adopt a neural map-matching approach~\cite{sarlin2023orienternet} that discretizes the pose space and estimates a categorical probability distribution:
\begin{equation}\label{eq:prob_volume}
    P\coloneqq p(\pose_{u,v,k}\mid \bI,\mathrm{map})~,   
\end{equation}
over a set of candidate poses $\{\pose_{u,v,k}\}$, where $u,v\in\{0,\twodots,S-1\}$ and $k\in\{0,\twodots,N-1\}$ correspond to headings $\theta_k = 360k/N$\degree. We use $P[\pose]$ to denote the estimated probability of the pose $\pose$. In practice, the network estimates an $S\times S\times N$ tensor of unnormalized scores via cross-correlation of learned $S\times S$ map features and $N$ rotated versions of bird's-eye-view (BEV) features extracted from $\bI$.
The scores are then normalized using softmax over all discrete poses $\pose_{u,v,k}$, yielding $p(\pose_{u,v,k}\mid \bI,\mathrm{map})$ as shown in \cref{fig:prob_dists}. During inference, $\arg\max_{\pose_{u,v,k}} P[\pose_{u,v,k}]$ is taken as the pose estimate. This setting is flexible and commonly adopted in related tasks such as satellite-based cross-view localization \cite{fervers2023cbev,Fervers2023uncertainty}.

\PAR{Motivation} Current approaches \cite{sarlin2023orienternet,liao2024osmloc,gao2025diffvl} assume perfectly geo-referenced ground-truth to supervise their 3-DoF pose estimates. However, geo-referencing is a challenging problem at scale, where manual verification is impractical. Open platforms such as Mapillary \cite{mapillary} help address this by jointly optimizing poses from multiple cameras that share visual observations. Although additional heuristics can be used to detect inaccuracies, such as agreement between GPS measurements and optimized poses \cite{sarlin2023orienternet,ho2024mapitanywhere}, they do not guarantee the complete removal of incorrect estimates, and these errors can propagate to models trained on them  (\cref{fig:teaser}). 
To address this problem, we propose several supervision techniques capable of learning from potentially inaccurate data labels.
Our proposals just require noisy 2-DoF GPS coordinates (\cref{subsec:chunk}) or relative poses (\cref{subsec:train_relposes}) obtained, \eg, from SfM. Our main strategies are depicted in \cref{fig:prob_dists}.

\subsection{Learning accurate 3-DoF poses from 2D labels}\label{subsec:chunk}

\PAR{Automatic heading angle} To supervise the probability volume predicted by neural map matchers, a common strategy \cite{sarlin2023orienternet,liao2024osmloc,Fervers2023uncertainty} is to minimize the negative log-likelihood (NLL) at the ground-truth pose $\pose_{(u, v, k)_\mathrm{GT}}$:
\begin{equation}
    \mathcal{L}_{\pose} = -\log P\!\left[\pose_{(u, v, k)_\mathrm{GT}}\right]~.
\end{equation}
This supervision requires ground-truth labels for heading angles.
However, the matching step between the learned map and BEV features provides a strong geometric inductive bias that we found can be exploited to predict the orientation. Intuitively, BEV features encode the scene semantics captured by the camera, as do the map features, so the similarity between the two feature sets is maximized at the correct orientation, provided that the geometry and semantics are estimated correctly. As a result, the heading angle can be supervised implicitly by 
maximizing the likelihood at the ground-truth position $(u_\mathrm{GT}, v_\mathrm{GT})$ of the resulting probability distribution after marginalizing the heading probabilities:
\begin{equation}\label{eq:loss_xy}
    \mathcal{L}_{\pose_{XY}} = -\log \sum_{k=0}^{N-1} P\!\left[\pose_{(u_\mathrm{GT}, v_\mathrm{GT}, k)}\right]~,
\end{equation}
and the model ``automatically'' learns useful patterns for predicting the heading.

A related observation was found by C-BEV \cite{fervers2023cbev}, but under stronger constraints: $360\degree$ panoramic ground images and both positive and negative aerial tile samples contrasted against each query panorama during training. In contrast, our simple approach uses only single perspective images during training. Despite having a narrow field of view, this is sufficient to predict heading with accuracy on-par with or better than methods supervised with heading labels.

\PAR{Learning from GPS} In favorable conditions, when not affected by adverse effects such as non-line-of-sight and multipath, GPS can provide meter-level accurate geo-referenced position estimates, with errors $<8$\,m in 95\% of cases \cite{gps}. However, GPS errors are often biased \cite{peretic2025statistical,williams2004error} and can become substantially larger under adverse conditions. Directly training on these measurements can thus encourage models to learn patterns that explain and replicate these errors.

Therefore, to improve over direct GPS supervision with \cref{eq:loss_xy}, we ``chunk'' the probability distribution over position by marginalizing within a $\pm r$ offset around the GPS coordinates, in addition to marginalizing over orientation. Concretely, we maximize the likelihood of the chunk that contains the GPS label:
\begin{equation}\label{eq:loss_xy_chunk}
    \mathcal{L}_{\mathrm{GPS},~\mathrm{chunk}} =
    -\log \sum_{i=-r}^{r}\sum_{j=-r}^{r}\sum_{k=0}^{N-1}
    P\!\left[\pose_{(u_\mathrm{GPS}+i,\, v_\mathrm{GPS}+j,\, k)}\right]~,
\end{equation}
where $r$ is the chunk ``radius''. In essence, we encourage the model to place probability mass anywhere within the $\pm r$ m neighborhood of the GPS measurement, again relying on the strong geometric inductive bias of the model. 
We experimentally show that this loss not only reduces the impact of GPS errors, but also improves over baseline models trained on optimized absolute 3-DoF poses. 

As shown in 
\cref{tab:supp_extra_ablations}, 
$r$ is not a sensitive parameter, as similar performance is obtained for $r\in[5, 20]$\,m. 
The loss in \cref{eq:loss_xy_chunk} is thus robust, \eg, to GPS errors of the magnitude commonly observed with receivers in egocentric devices~\cite{Krishnan2025lamaria,engel2023aria}.

\begin{figure}[t]
  \def\thumbH{1.1cm}
  \def\plotH{2.35cm}
    \definecolor{myorange}{RGB}{242,169,59}
    \centering
    \begin{tblr}{
        colspec={*5{c}},
        vspan=even,
        colsep=1.5pt,
        cell{2}{3,6}={c=2}{c},
        cell{2}{2-Z}={r=2}{c},
        row{2}={belowsep=0pt},
        row{3}={abovesep=0pt},
        column{1,4}={ rightsep+=2.5pt },
        column{2,5}={ leftsep+=2.5pt },
        vline{2,5}={2-3}{gray8},
        vline{2,5}={2}{abovepos=-2},
        vline{2,5}={3}{belowpos=-2},
        stretch=0,
    }
     Images 
     & Shifted GT \raisebox{-.1\width}{\includegraphics[width=2.7mm]{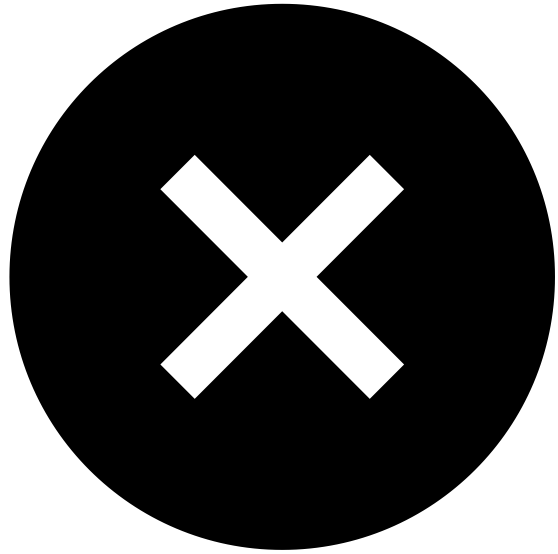}}
     & $\to$  & Effect on loss 
     & Rotated GT \raisebox{-.1\width}{\includegraphics[width=2.7mm]{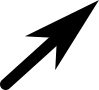}}
     & $\to$ & Effect on loss \\
          \includegraphics[height=\thumbH,cfbox=red 1pt 0pt]{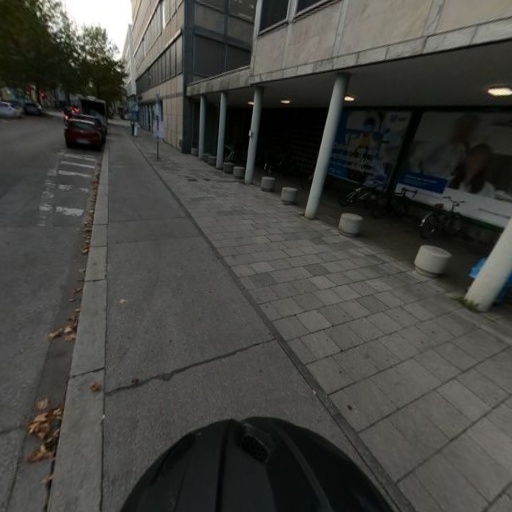} 
        & \includegraphics[height=\plotH]{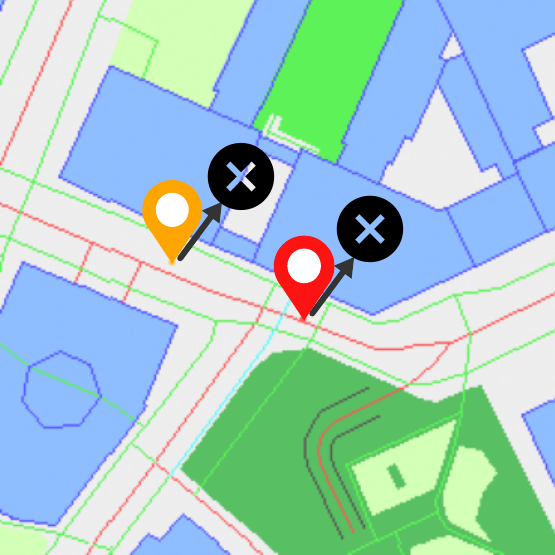} 
        & \includegraphics[height=\plotH]{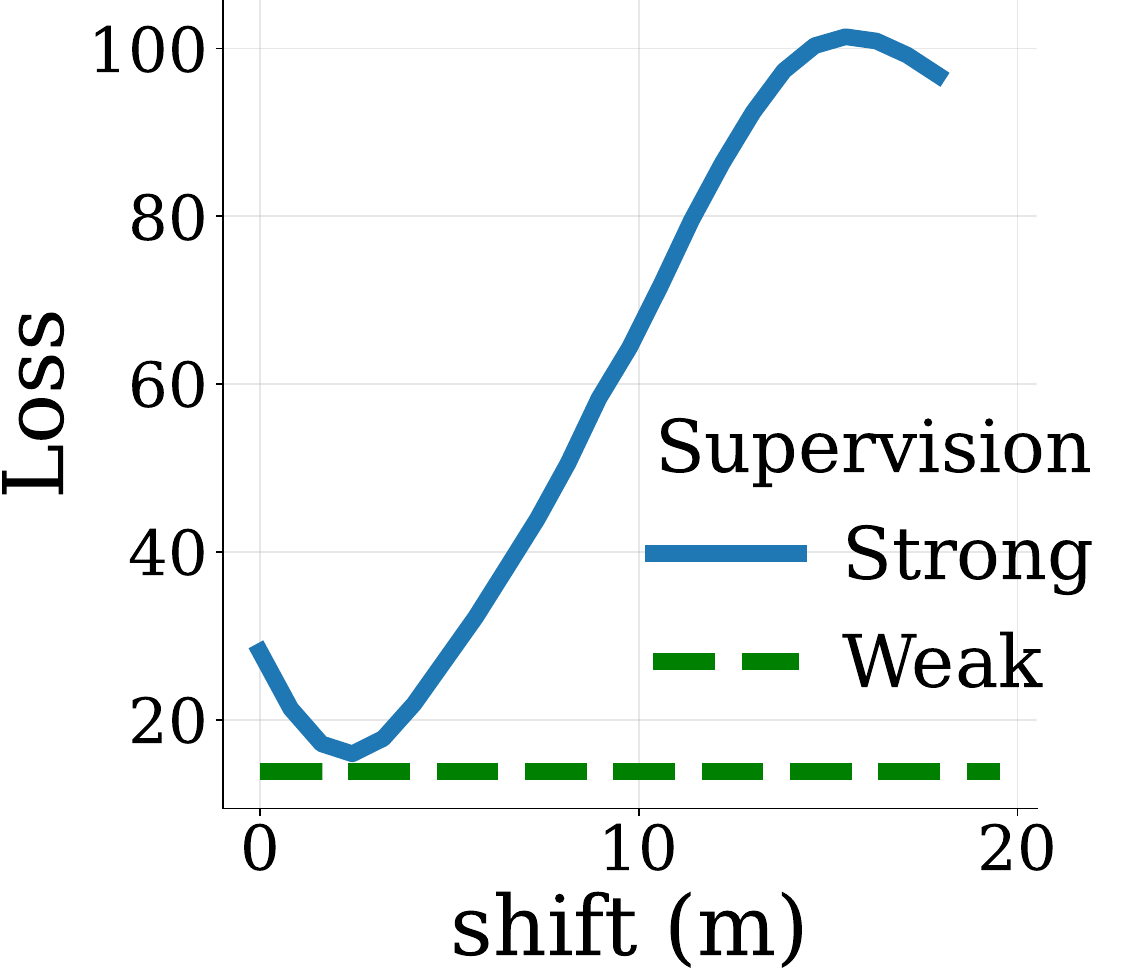} 
        && \includegraphics[height=\plotH]{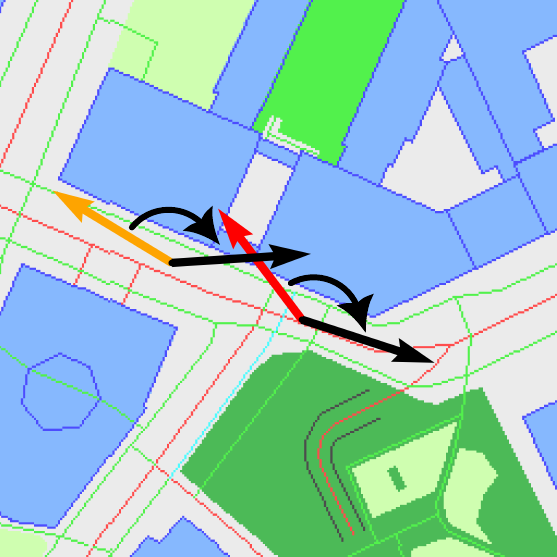} 
        & \includegraphics[height=\plotH]{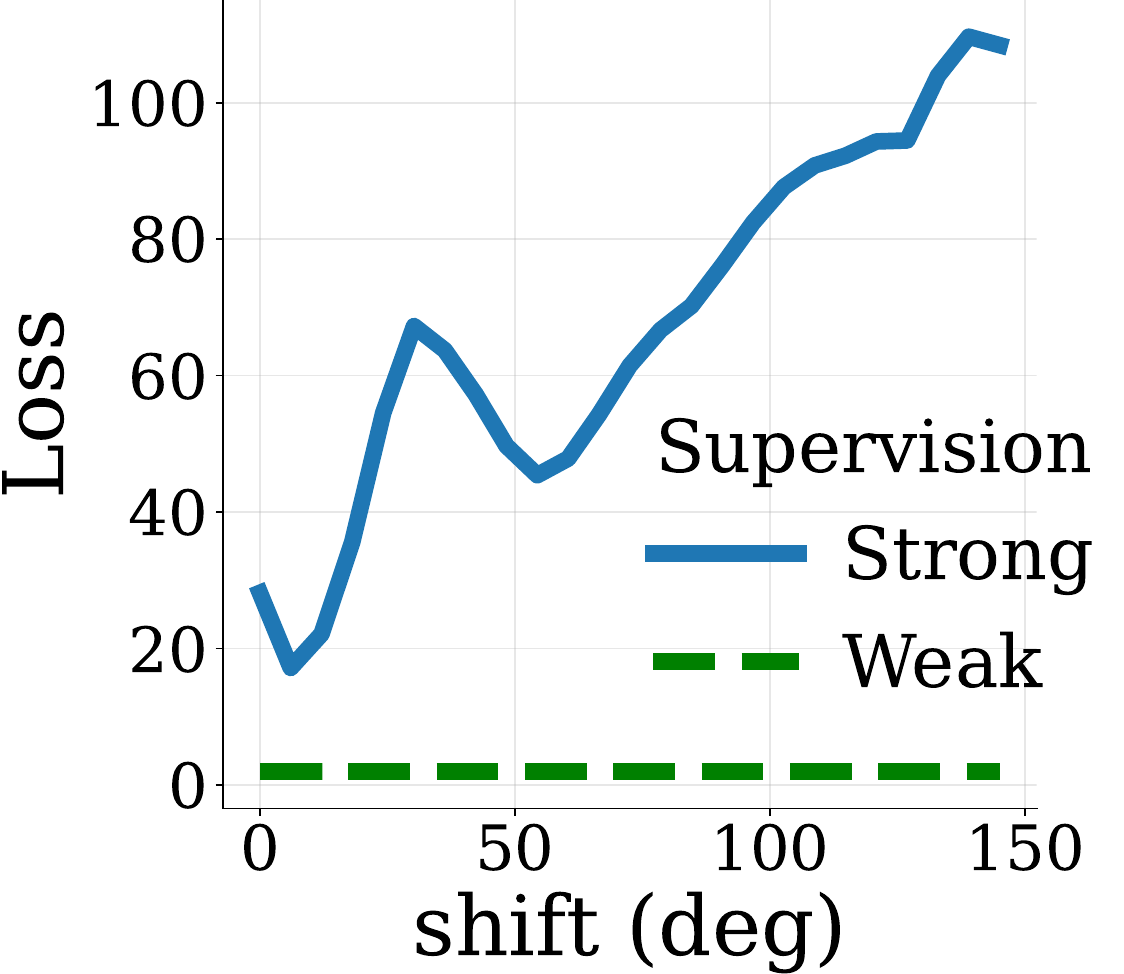} 
        \\
        \includegraphics[height=\thumbH,cfbox=myorange 1pt 0pt]{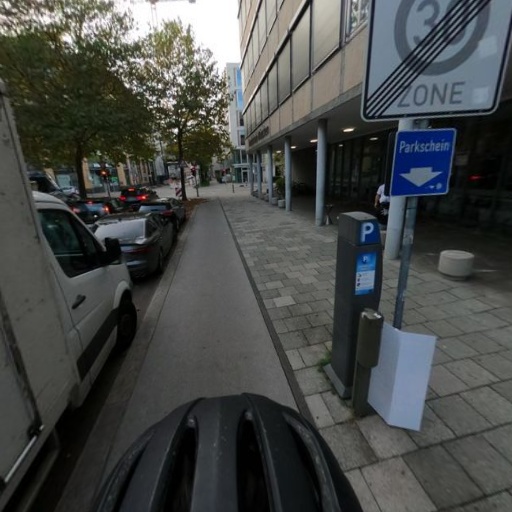} \\
    \end{tblr}
    \caption{\textbf{Robustness to ground-truth errors.} \name' weak supervision is insensitive to GT's translation (middle) and rotation (right) offsets shared by datapoints involved in our losses (\cref{eq:loss_relrot,eq:loss_reltrans}) since we just use their relative information. 
    In contrast, strong supervision \cite{sarlin2023orienternet,liao2024osmloc} is not robust and heavily penalizes these same errors (see loss curves), which forces networks to learn patterns that explain them.}
    \label{fig:losses_invariances}
\end{figure}

\subsection{Learning accurate 3-DoF poses from relative poses}\label{subsec:train_relposes}

Relative poses, \eg obtained from SLAM~\cite{mur2017orb2,campos2021orb3} or SfM~\cite{schoenberger2016sfm,pan2024glomap},
are constrained by multi-view observations whose noise and outliers are well handled by robust optimization techniques~\cite{triggs2000bundle,raguram2013usac}
and off-the-shelf systems \cite{campos2021orb3,engel2023aria,schoenberger2016sfm,opensfm,pan2024glomap}. 
We show that such relative poses can be used directly for training, without explicit geo-referencing, as long as a coarse 2D location (\eg, from GPS) is available to sample an appropriate map tile covering a datapoint. %
To this end, and as shown in \cref{fig:prob_dists}, we supervise the rotation and translation marginals of the relative-pose distribution between pairs of datapoints, which allows us to weight them according to their discretization level. These marginals are computed from the independently predicted absolute-pose distributions of each datapoint, denoted by $P_0$ and $P_1$ for datapoints $0$ and $1$, respectively. 

\PAR{Relative rotation}
We first obtain the categorical distribution over the $N$ discrete \emph{absolute} headings by marginalizing over positions:
\begin{equation}\label{eq:rot_probs}
    P_i^{\Theta}[k] \;\coloneqq\; \sum_{u=0}^{S-1}\sum_{v=0}^{S-1} P_i\!\left[\pose_{u,v,k}\right],
    \qquad i\in\{0,1\},\; k\in\{0,\dots,N-1\}~.
\end{equation}
Let $\Delta k \in \{0,\dots,N-1\}$ denote a discrete relative rotation, corresponding to
$\Delta\theta_{\Delta k} = 360\,\Delta k/N\degree$, then the 
distribution over relative rotations is the \emph{circular cross-correlation} of the two absolute heading distributions:
\begin{equation}\label{eq:relrot_circxcorr}
    P_{\Delta\Theta}[\Delta k]
    \;\coloneqq\;
    \sum_{k=0}^{N-1}
    P_0^{\Theta}[k]\;
    P_1^{\Theta}\!\Big[(k+\Delta k)\bmod N\Big]~,
\end{equation}
\ie, we sum the joint probabilities of absolute headings with the same relative heading, $\Delta k$, under the assumption that $P_0^{\Theta}$ and $P_1^{\Theta}$ are independent.%

For supervision, we minimize the NLL of the distribution at the ground-truth relative rotation $\Delta k_{\mathrm{GT}}\ceqq\Delta\theta_{\mathrm{GT}}N/360\degree$:
\begin{equation}\label{eq:loss_relrot}
    \mathcal{L}_{\Delta\Theta} \;=\; -\log P_{\Delta\Theta}\!\left[\Delta k_{\mathrm{GT}}\right]~,
\end{equation}
and linearly interpolate the log-probabilities to achieve sub-bin precision\footnote{This corresponds to minimizing the cross-entropy between a ground-truth categorical distribution with soft labels (interpolation weights) and the predicted one.}. 

\PAR{Relative translation}
We obtain the categorical distribution over the \emph{absolute} 2D position on each tile by marginalizing the predicted volume over headings:
\begin{equation}\label{eq:trans_probs}
    P_i^{XY}[u,v] \;\coloneqq\; \sum_{k=0}^{N-1} P_i\!\left[\pose_{u,v,k}\right],
    \qquad i\in\{0,1\},\; u,v\in\{0,\dots,S-1\}~.
\end{equation}
To relate these local coordinates to a common (geo-referenced) frame, let $\bo_i\in\mathbb{R}^2$ denote the known geo-referenced coordinates of 
the tile's origin (\ie the top-left corner of the map tile used for datapoint $i$). Then, each discrete absolute translation can be written as 
$\bt_i(u,v) = \bo_i + \begin{bmatrix}u & v\end{bmatrix}^\top$, with $i\!\in\!\{0,1\}$,
and the corresponding relative translation between two candidates is thus given by
\begin{equation}\label{eq:reltrans_decomp}
    \bt_{01}
    \;\coloneqq\;
    \bt_1(u_1,v_1) - \bt_0(u_0,v_0)
    \;=\;
    \underbrace{(\bo_1-\bo_0)}_{\text{constant}}
    \;+\;
    \underbrace{\begin{bmatrix}u_1-u_0 &&& v_1-v_0\end{bmatrix}^\top}_{(\Delta u,\Delta v)}~.
\end{equation}
Since the tile origins $\bo_0,\,\bo_1$ are constants, the only random quantity to model is the \emph{discrete relative shift} $(\Delta u,\Delta v)$ expressed in local tile coordinates.
We obtain their probabilities by summing the joint probabilities of all absolute positions that induce that same shift:
\begin{equation}\label{eq:reltrans_linxcorr}
    P_{\Delta XY}[\Delta u,\Delta v]
    \;\coloneqq\;
    \sum_{u=0}^{S-1}\sum_{v=0}^{S-1}
    P_0^{XY}[u,v]\;
    P_1^{XY}[u+\Delta u,\; v+\Delta v]~,
\end{equation}
where $P_1^{XY}[\cdot,\cdot]$ is treated as zero outside $\{0,\dots,S-1\}^2$. %
\cref{eq:reltrans_linxcorr} is a \emph{linear} (2D) cross-correlation between two absolute translation distributions, producing a $(2S-1)\times(2S-1)$ categorical distribution over relative shifts.

For supervision, we minimize the NLL at the ground-truth relative shift
\begin{equation}\label{eq:loss_reltrans}
    \mathcal{L}_{\Delta XY} \;=\; -\log P_{\Delta XY}\!\left[\Delta u_{\mathrm{GT}},\Delta v_{\mathrm{GT}}\right]~,
\end{equation}
and bilinearly interpolate the log-probabilities to 
achieve sub-bin precision.

\PAR{Relative distance}
As shown in \cref{fig:losses_invariances}, supervision with \cref{eq:loss_relrot,eq:loss_reltrans} is invariant to offsets in absolute-pose labels. 
However, a limitation of the relative translation loss in \cref{eq:loss_reltrans} is that it requires expressing the GT relative translation in the same (geo-referenced) coordinate system as the map tiles, so that it can be mapped to a GT relative shift $(\Delta u_{\mathrm{GT}},\Delta v_{\mathrm{GT}})$ in \cref{eq:reltrans_decomp}. 
To unlock the usage of relative translations expressed in an arbitrary coordinate system, we can instead supervise the probability distribution of their norms. %

Starting from the discrete relative-shift distribution $P_{\Delta XY}$ (\cref{eq:reltrans_linxcorr}), we can associate each shift $(\Delta u,\Delta v)$ with the norm of the corresponding translation, as the tile origins $\bo_0,\,\bo_1$ are geo-referenced and known:
\begin{equation}
    \norm(\Delta u, \Delta v)
    \ceqq
    \lVert(\bo_1-\bo_0) + \begin{bmatrix}\Delta u & \Delta v\end{bmatrix}^\top\rVert,
\end{equation}
We then obtain the categorical distribution over (discrete) norms by marginalizing $P_{\Delta XY}$ over all shifts that yield the same norm value:
\begin{equation}\label{eq:norm_probs}
    P_{\lVert \Delta \bt \rVert}[d]
    \;\coloneqq\;
    \sum_{\Delta u,\Delta v:\; \norm(\Delta u,\Delta v)=d}
    P_{\Delta XY}[\Delta u,\Delta v]~.
\end{equation}
$P_{\lVert \Delta \bt \rVert}$ is non-smooth and not appropriate for interpolation at the ground-truth norm (as shown in
\cref{fig:supp_prob_dists}). 
Thus, and similarly to the chunk-based marginalization of \cref{subsec:chunk}, we marginalize the raw distribution into contiguous bins and then maximize the likelihood of the bin that contains the ground-truth norm:
\begin{equation}\label{eq:loss_chunked_norm}
    \mathcal{L}_{\lVert \Delta \bt \rVert}
    \;\coloneqq\;
    -\log \sum_{d\in\mathcal{C}(r_{\mathrm{GT}})}
    P_{\lVert \Delta \bt \rVert}[d],
    \qquad
    \mathcal{C}(r_{\mathrm{GT}})\ceqq\{\,d \mid \lvert d-d_{\mathrm{GT}}\rvert \le r_{\Delta \bt}\,\},
\end{equation}
where $r_{\Delta \bt}$ represents the half-bin width. The resulting distribution is smoother and %
robust to errors in the relative translation norms. 
As shown in 
\cref{tab:supp_extra_ablations}, 
$\rdt$ is not a sensitive parameter. In practice we use a conservative $r_{\Delta \bt}=5\,$m.

\subsection{Implementation}
We build on OrienterNet~\cite{sarlin2023orienternet}. For most experiments, we keep the same architecture and hyperparameters in order to isolate the effect of our contributions. %

\PAR{Training dataset}
We use the Mapillary Geo-Localization (MGL) dataset~\cite{sarlin2023orienternet}, containing 760k images from 12 cities across Europe and the US. 
At the time of writing, data from Amsterdam (72k images) is no longer available.
We train \name and baselines on the remaining 11 cities.
Images are captured by handheld devices or cameras mounted on cars or bikes, and come with OSM data and geo-referenced 6-DoF poses obtained by fusing SfM and GPS. 

\PAR{Relative-pose supervision}
We sample pairs of datapoints that were optimized within the same SfM cluster. 
We reject pairs whose camera centers are more than 100\,m from each other to avoid large relative baselines potentially affected by drift. Otherwise, we do not restrict the type of relative motion. 
We sample OSM map tiles for each datapoint independently.

\PAR{Training details}
We use the GPS coordinates to sample the map tile. 
We center the tile at the GPS location with a random translation and a random rotation to avoid overfitting. For comparisons to OrienterNet, we use the same U-Net-based architecture for the image and map branches, with ResNet-101 \cite{he2016deep} and VGG-19 \cite{Simonyan15vggn} backbones, respectively. 
We also experiment with a DINOv2~\cite{oquab2024dinov2} image encoder inspired by \cite{liao2024osmloc}.
We train with the Adam optimizer \cite{kingma2014adam} and a batch size of 12. We typically train for 500k steps on two 40\,GB NVIDIA A100 GPUs ($\sim$3 days).
More implementation details can be found in 
\cref{sec:supp_training_details}.

\section{Experiments}

We evaluate our contributions on benchmarks with accurate, geo-referenced GT.
For the main evaluations in this section, we consider two training strategies:
\begin{enumerate}[noitemsep,topsep=0pt]
    \item \textbf{raw GPS}: uses $\mathcal{L}_{\mathrm{GPS},~\mathrm{chunk}}$ (\cref{eq:loss_xy_chunk}) with a chunk radius of $r\!=\!5$\,m, and
    \item \textbf{w/ rel. poses}: uses $ 0.1\mathcal{L}_{\Delta\Theta} + \mathcal{L}_{\Delta XY}$ (\cref{eq:loss_relrot}, \cref{eq:loss_reltrans}) as loss.
\end{enumerate}
We show that: (1) 2-DoF GPS labels, when combined with chunk marginalization (\emph{raw GPS}), are sufficient to outperform baselines trained with 3-DoF labels; and (2) relative-pose supervision (\emph{w/ rel. poses}) yields a new state of the art.
We evaluate additional weak-supervision strategies tailored to practical scenarios in 
\cref{sec:supp_quantitative}, 
which yield similar conclusions and \eg show that similar performance is obtained with $r\in[5,20]$m.
Qualitative results are shown in 
\cref{fig:qualitative,sec:supp_qualitative}.

\PAR{Baselines} Our main baseline is OrienterNet \cite{sarlin2023orienternet}. 
We compare against its pretrained checkpoint, trained on a previous version of MGL \cite{sarlin2023orienternet} containing 12 cities. 
We also compare against the pretrained checkpoints of OSMLoc \cite{liao2024osmloc}, which combines OrienterNet with a DINOv2 backbone \cite{oquab2024dinov2} and additional losses to guide its depth predictions \cite{lihe2024depthanything} and similarity between BEV and map features. 
Both baselines \cite{sarlin2023orienternet,liao2024osmloc} are trained with strong supervision on geo-referenced labels, and we report their results after rerunning the evaluation with their released code and checkpoints.
We additionally include two baselines: OrienterNet retrained on the 11 currently available cities of the MGL dataset, \ie the same data used to train \name; and OrienterNet retrained with a DINOv2 image backbone, as in OSMLoc.
Finally, for context, we also report results for state-of-the-art methods that match ground images to satellite maps \cite{xia2024ccvpe,xia2026loc,lentsch2023slicematch}.

\begin{table}[t]

\newcommand{\slic}{SliceMatch \cite{lentsch2023slicematch}}
\newcommand{\ccvp}{CCVPE \cite{xia2024ccvpe}}
\newcommand{\loct}{Loc$^2$ \cite{xia2026loc}}
\newcommand{\onet}{OrienterNet \cite{sarlin2023orienternet}}
\newcommand{\oloc}{OSMLoc \cite{liao2024osmloc}}

\centering
\resizebox{\textwidth}{!}{%
\begin{tblr}{
    colspec={lcll*{9}{c}},
    colsep=8pt,
    stretch=0pt,
    cell{1}{1-4}={r=2}{l},
    cell{1,2}{5,8,11}={c=3}{c},
    cell{3}{1,4}={r=3}{l},
    cell{3}{2}={r=3}{c},
    cell{6}{1,2}={r=2}{l},
    cell{8,12}{1,2,4}={r=4}{l},
    column{1,2,3} = {leftsep=2pt,rightsep=2pt},
    column{6,9,12} = {leftsep=0.5pt,rightsep=0.5pt},
    cell{Z}{5-Z}={font=\bfseries},
    cell{Y}{5-X}={cmd=\underline},
    cell{W}{Y-Z}={cmd=\underline},
}
\toprule
Map & \shortstack[c]{Training \\ dataset} & Method & \shortstack[c]{Image \\ encoder} & Lateral [m] & & & Longitudinal [m] & & & Orientation [$^\circ$] & & \\
&&&& R@1/3/5 $\uparrow$ & & & R@1/3/5 $\uparrow$ & & & R@1/3/5 $\uparrow$ & & \\
\midrule
Satellite & KITTI & 
   \slic & N/A & 24.0 & - & 72.9 &  7.2 & - & 33.1 & 31.7 & 31.7 & 31.7 \\
&& \ccvp &     & 23.4 & - & 60.5 & 11.8 & - & 42.1 &  3.1 & - & 14.6 \\
&& \loct &     & 13.6 & - & 50.9 & 14.0 & - & 50.7 &  2.5 & - & 12.8 \\
\midrule
OSM & \shortstack[c]{MGL \\ (12 cities)} &
   \onet & ResNet-101 & 42.5 & 74.8 & 83.2 & 21.7 & 49.8 & 61.0 & 20.3 & 50.9 & 65.1 \\
&& \oloc & DINOv2-B   & 49.9 & 82.5 & 87.0 & 25.7 & 56.4 & 66.5 & 22.4 & 56.2 & 72.7 \\
\midrule[gray5,dashed]
OSM & \shortstack[c]{MGL \\ (11 cities)} &
   \onet                     & ResNet-101 & 37.7 & 67.3 & 77.4 & 17.6 & 42.6 & 53.7 & 15.7 & 42.1 & 57.1 \\
&& \textbf{\name} \\
&& $\hookrightarrow$ raw GPS &            & 43.1 & 78.8 & 85.6 & 26.8 & 56.1 & 65.1 & 21.1 & 53.0 & 69.2 \\
&& $\hookrightarrow$ w/ rel. poses &      & 56.6 & 83.2 & 87.8 & 33.2 & 59.5 & 66.0 & 28.9 & 64.4 & 75.9 \\
\midrule[gray5,dashed]
OSM & \shortstack[c]{MGL \\ (11 cities)} &
   \onet                      & DINOv2-B & 48.7 & 86.0 & 91.3 & 24.7 & 62.4 & 73.3 & 29.8 & 70.2 & 83.3 \\
&& \textbf{\name} \\
&& $\hookrightarrow$ raw GPS &           & 62.3 & 88.3 & 92.2 & 33.7 & 66.6 & 75.1 & 30.5 & 69.3 & 82.6 \\
&& $\hookrightarrow$ w/ rel. poses &     & 70.8 & 91.1 & 94.2 & 34.1 & 70.8 & 77.6 & 37.9 & 78.5 & 88.5 \\
\bottomrule
\end{tblr}
}
\caption{\textbf{Results on KITTI \cite{geiger2012kitti}}. We report position and orientation error recalls at different thresholds. Best results among methods using OSM are marked \textbf{bold}, and second best are \underline{underlined}. The current version of the MGL dataset \cite{sarlin2023orienternet} is missing one of the 12 original cities. For a fair comparison, we retrain our baseline, OrienterNet \cite{sarlin2023orienternet}, on the reduced training set. When trained on raw 2D GPS labels, \name outperforms OrienterNet trained on geo-referenced 3-DoF poses. When supervising with relative poses, \name achieves best results across all OSM methods, even those trained on more data. Using a DINOv2 backbone further improves the performance.}
\label{tab:kitti_main}
\end{table}

\subsection{Driving scenarios}\label{subsec:exp_driving}
\PAR{Dataset} We use the ``Test2'' split defined by \cite{shi2022beyond} on the KITTI dataset \cite{geiger2012kitti}, which contains driving sequences in residential and road areas captured by cameras mounted on a car. It provides accurate RTK-corrected, geo-referenced ground-truth, and we use the rasterized OSM tiles provided by OrienterNet \cite{sarlin2023orienternet}.

\PAR{Setup} We follow standard protocols \cite{sarlin2023orienternet,xia2026loc}: we compute lateral (perpendicular) and longitudinal (parallel) position errors relative to the driving direction, as well as orientation errors, and report recall at 1/3/5\,m/\degree.
These protocols also restrict the search to a $40\!\times\!40$\,m area whose center matches that of the tile and that is randomly sampled within $\pm20$m of the GT position.
Following \cite{xia2026loc}, we assume no orientation prior, while the evaluation in 
\cref{sec:supp_quantitative} 
follows \cite{sarlin2023orienternet}, and restricts orientations to $\pm10\degree$ from the GT orientation.
See 
\cref{sec:supp_eval_details} 
for more details.

\PAR{Results}
As can be seen in \cref{tab:kitti_main}, \name consistently outperforms strongly supervised baselines \cite{sarlin2023orienternet,liao2024osmloc}, even when trained with just raw GPS measurements. Supervision with relative poses further improves accuracy, making \name the best-performing method, including those trained on the full MGL dataset and satellite-to-ground approaches trained on KITTI. Using a stronger DINOv2-based image encoder further improves the performance.

\begin{table}[t]
\newcommand{\gnss}{GPS}
\newcommand{\oloc}{OSMLoc-S \cite{liao2024osmloc}}
\newcommand{\olocb}{OSMLoc-B \cite{liao2024osmloc}}
\newcommand{\onet}{OrienterNet \cite{sarlin2023orienternet}}
\newcommand{\oner}{OrienterNet$^*$ \cite{sarlin2023orienternet}}
\newcommand{\auto}{\textbf{\name}}

\newcommand{\kiti}{KITTI \cite{geiger2012kitti}}
\newcommand{\oxfr}{\shortstack[c]{Oxford \\ Day-and-Night \cite{wang2025oxford}}}
\newcommand{\lama}{LaMAria \cite{Krishnan2025lamaria}}

\definecolor{best}{RGB}{178,255,178}
\definecolor{secd}{RGB}{255,235,158}

\centering
\resizebox{1\linewidth}{!}{%
\begin{tblr}{
    rowsep=0.pt,
    colspec={cl*{14}{c}}, 
    colsep=2pt, 
    column{1,7,13}={rightsep=6pt}, 
    column{10}={rightsep=12pt}, 
    cell{1}{1-4}={r=2}{l},
    cell{1}{4}={r=2}{c},
    cell{1,2}{5,8,11,14}={c=3}{c}, 
    cell{3}{1}={r=11}{c, m, cmd=\rotatebox{90}}, 
    cell{14}{1}={r=10}{c, m, cmd=\rotatebox{90}}, 
    cell{4}{3}={r=2}{c, m}, 
    cell{6}{3,4}={r=4}{c, m}, %
    cell{10}{3,4}={r=4}{c, m}, %
    cell{14}{3}={r=2}{c, m}, 
    cell{16}{3,4}={r=4}{c, m}, %
    cell{20}{3,4}={r=4}{c, m}, %
    cell{13}{5-12,14-16}={font=\bfseries},
    cell{12}{12,13}={font=\bfseries},
    cell{12}{5-10,14-16}={cmd=\underline},
    cell{9}{11}={cmd=\underline},
    cell{13}{13}={cmd=\underline},
    cell{10}{16}={cmd=\underline},
    cell{23}{5-14}={font=\bfseries},
    cell{22}{13}={font=\bfseries},
    cell{19}{15}={font=\bfseries},
    cell{20}{16}={font=\bfseries},
    cell{22}{5-10}={cmd=\underline},
    cell{19}{11,12,14,16}={cmd=\underline},
    cell{15}{15}={cmd=\underline},
    cell{14}{16}={cmd=\underline},
}
\toprule
    & Method & \shortstack{Training \\ MGL version} & \shortstack{Image \\ Encoder} & XY [m] &-&-& Orientation [\degree]&-&- & XY (seq) [m] &-&-& Orient. (seq) [\degree]&-&- \\
\cmidrule[lr]{5-7} \cmidrule[lr]{8-10} \cmidrule[lr]{11-13} \cmidrule[lr]{14-16} 
    & & & &  R@1/3/5\,\textuparrow & - & - & R@1/3/5\,\textuparrow & - & - & R@1/3/5\,\textuparrow & - & - & R@1/3/5\,\textuparrow & - & - \\
\midrule
\lama 
& \gnss                      & -         & -          & 6.0 & 37.2 & 65.2 & - & - & - & 9.3 & 49.3 & 74.7 & 16.0 & 59.5 & 77.2\\
\cline{2-16}
& \onet                      & 12 cities & ResNet-101 & 2.7 & 12.6 & 20.1 & 5.0 & 14.6 & 22.1 & 36.5 & 77.2 & 80.2 & 48.7 & 81.5 & 84.5 \\
& \olocb                     &           & DINOv2-B   & 6.7 & 26.3 & 37.1 & 10.1 & 27.0 & 39.0 & 59.6 & 93.2 & 94.3 & 69.2 & 92.8 & 94.0 \\
\cline[dashed]{2-16}
& \onet                      & 11 cities & ResNet-101 & 1.9 & 10.9 & 18.2 & 4.8 & 13.4 & 20.7 & 35.8 & 76.4 & 84.6 & 50.8 & 83.8 & 89.0 \\
& \auto \\
& $\hookrightarrow$ raw GPS  &           &            & 3.7 & 16.5 & 26.3 & 7.3  & 20.0 & 28.7 & 47.9 & 85.6 & 86.4 & 61.8 & 89.0 & 91.7 \\
& $\hookrightarrow$ w/ rel. poses &      &            & 8.1 & 25.2 & 33.0 & 10.5 & 26.4 & 35.5 & 72.6 & 92.1 & 92.6 & 77.8 & 92.8 & 94.0\\
\cline[dashed]{2-16} 
& \onet                      & 11 cities & DINOv2-B   &  9.1 & 32.5 & 43.6 & 11.4 & 30.4 & 43.3 & 57.5 & 92.9 & 93.7 & 70.8 & 94.8 & 95.0\\
& \auto \\
& $\hookrightarrow$ raw GPS   &           &           & 10.9 & 40.8 & 52.3 & 14.5 & 37.7 & 51.8 & 70.5 & 95.1 & 95.6 & 82.1 & 95.0 & 95.0\\
& $\hookrightarrow$ w/ rel. poses   &    &            & 20.9 & 49.2 & 56.3 & 19.5 & 47.0 & 59.5 & 86.2 & 95.1 & 95.3 & 87.8 & 96.0 & 96.0 \\
\midrule \midrule
\oxfr
& \onet                      & 12 cities & ResNet-101 & 10.4 & 37.2 & 52.2 & 15.6 & 38.6 & 51.7 & 77.5 & 96.1 & 99.5 & 70.5 & 95.7 & 99.9 \\
& \olocb                     &           & DINOv2-B   & 15.5 & 49.9 & 63.9 & 21.5 & 49.9 & 63.1 & 84.6 & 98.9 & 99.7 & 78.3 & 99.3 & 99.8 \\
\cline[dashed]{2-16} 
& \onet                      & 11 cities & ResNet-101 & 9.8  & 36.3 & 50.5 & 14.6 & 35.1 & 48.5 & 78.6 & 96.2 & 99.6 & 73.7 & 97.0 & 99.4 \\
& \auto   \\
& $\hookrightarrow$ raw GPS  &           &            & 10.4 & 36.8 & 52.2 & 17.8 & 41.6 & 54.4 & 76.6 & 94.8 & 98.4 & 69.5 & 93.0 & 98.4\\
& $\hookrightarrow$ w/ rel. poses &      &            & 13.2 & 42.9 & 56.4 & 19.0 & 43.8 & 56.5 & 86.8 & 99.4 & 99.4 & 80.6 & 99.7 & 99.9 \\
\cline[dashed]{2-16} 
& \onet                      & 11 cities & DINOv2-B   & 16.5 & 53.6 & 66.6 & 20.1 & 48.5 & 63.2 & 84.9 & 98.3 & 99.9  & 78.6 & 99.0 & 100.0 \\
& \auto   \\
& $\hookrightarrow$ raw GPS   & &                     & 16.8 & 55.1 & 68.6 & 22.9 & 52.4 & 66.9 & 83.2 & 99.0 & 100.0 & 73.2 & 99.0 & 99.4 \\
& $\hookrightarrow$ w/ rel. poses   &    &            & 20.8 & 60.7 & 73.5 & 24.7 & 57.0 & 71.7 & 90.5 & 99.9 & 100.0 & 83.2 & 99.2 & 99.4 \\
\bottomrule
\end{tblr}
}
    \caption{\textbf{Results on egocentric sequences \cite{Krishnan2025lamaria, wang2025oxford}} We show position and orientation recall at different thresholds. Best results in \textbf{bold} across OSM-based methods, second best \underline{underlined}. Table center shows results based on single frame estimates, right side shows results after sequential fusion of 50 frames. \name supervised with relative poses and using a DINOv2 backbone performs best across both datasets.} 
    \label{tab:sequ_results_main}
\end{table}

\subsection{Egocentric sequences}\label{subsec:exp_ar}

\PAR{Datasets} We use the LaMAria \cite{Krishnan2025lamaria} and Oxford Day-and-Night (ODN) \cite{wang2025oxford} datasets, both captured with Aria Glasses \cite{engel2023aria}. LaMAria provides km-long sequences in Zurich with cm-accurate ground-truth poses obtained by recording survey-grade control points at regular intervals. While LaMAria includes natural motions representative of AR-device use, the motions required to record survey control points are artificial (mostly looking at the ground), which makes the benchmark harder in an unintended way. We manually remove these images to retain only those reflecting normal device usage (details in 
\cref{sec:supp_eval_details}).

ODN's poses are estimated via multi-sensor fusion, including GPS and visual-inertial measurements, and are optimized across multiple SLAM sessions with loop closures.
The poses are locally highly accurate (cm-level \cite{wang2025oxford}), but the geo-referencing can have meter-level errors. Thus, we compute a robust per-sequence alignment between each method’s estimates and ground-truth (see 
\cref{sec:supp_eval_details}). 
We uniformly subsample both datasets to 5k evaluation images.

\PAR{Setup} We compute position and orientation error norms and report recall at 1/3/5\,m/\degree. We render $128\!\times\!128$\,m tiles for each datapoint, use no orientation prior and restrict the position search to a $64\!\times\!64$\,m area. In LaMAria, we center the tile and search area at the GPS coordinates. 
In ODN, GPS labels were released only recently, so we randomly sample the center within $\pm32$\,m of the ground-truth.

\begin{figure}[t]

\def\plotH{1.15cm}
\newcommand{\annotimg}[4][]{%
  \adjustbox{valign=m}{%
    \begin{overpic}[#1,height=\plotH]{#2}
      \put(#3){%
        \makebox(0,0)[lb]{%
          \tikz[overlay]{
            \node[
              anchor=south west,
              fill=white,
              fill opacity=0.6,
              text opacity=1,
              inner sep=1pt,
              rounded corners=1pt
            ] {\scriptsize #4};
          }%
        }%
      }%
    \end{overpic}%
  }%
}

    \centering
    \begin{tblr}{
        colspec={*{6}c},
        colsep=1pt, 
        rowsep=1pt, 
        row{1,2,4}={belowsep=0.5pt},
        row{2,3,5}={abovesep=0.5pt},
        row{Y}={belowsep=0pt},
        row{Z}={abovesep=0pt},
        stretch=0pt,
        cell{2-5}{1}={}{halign=c, valign=m,cmd={\scriptsize\rotatebox[origin=c]{90}}},
    }
    & Images \cite{wang2025oxford, Krishnan2025lamaria} 
    & Input Map & Neural Map & Features Norm & Probabilities \\
    \name
    & \includegraphics[valign=m,height=\plotH]{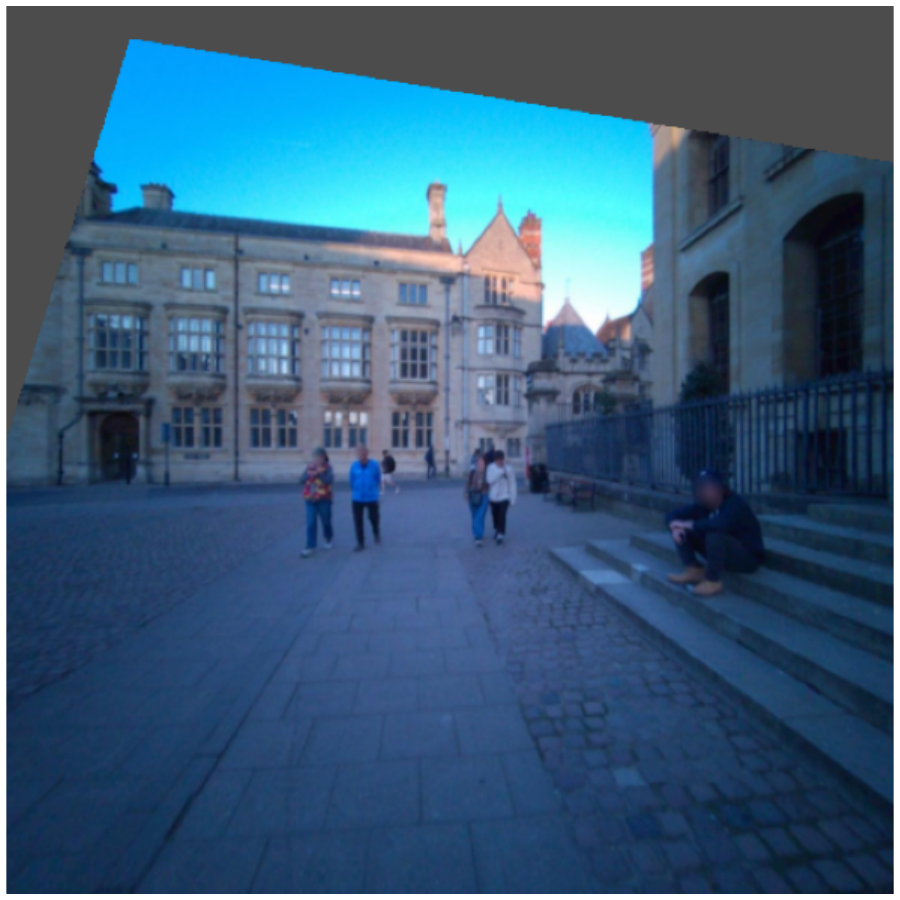} 
    & \includegraphics[valign=m,height=\plotH]{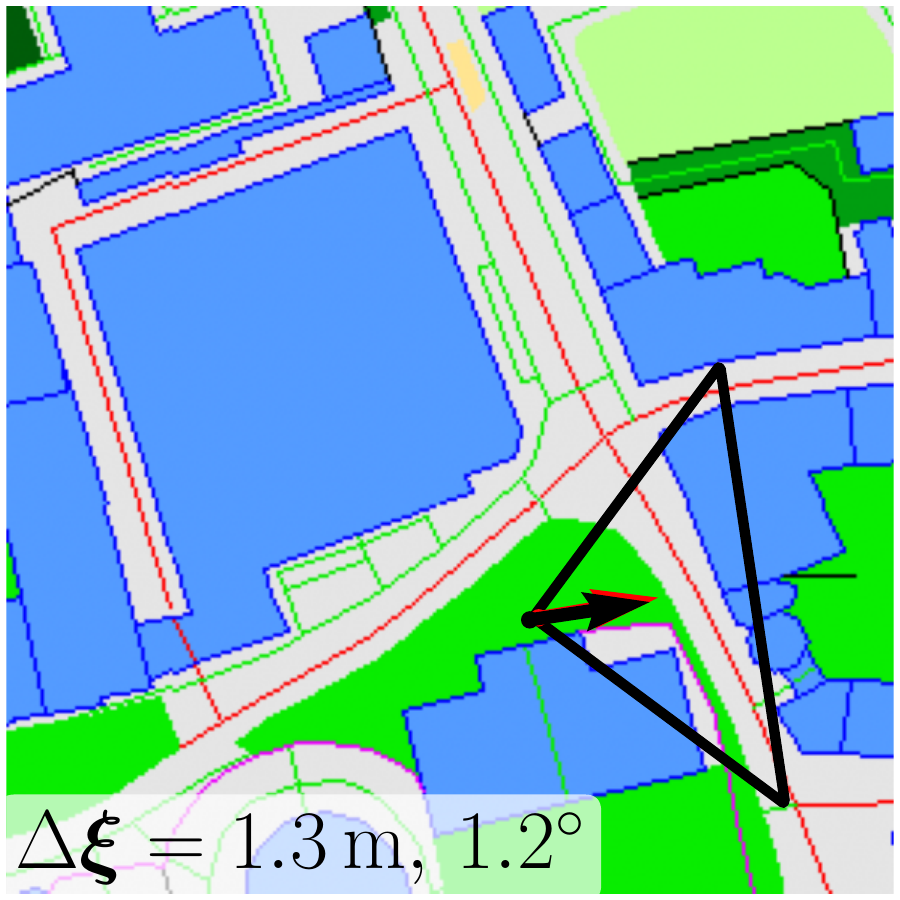} 
    & \includegraphics[valign=m,height=\plotH]{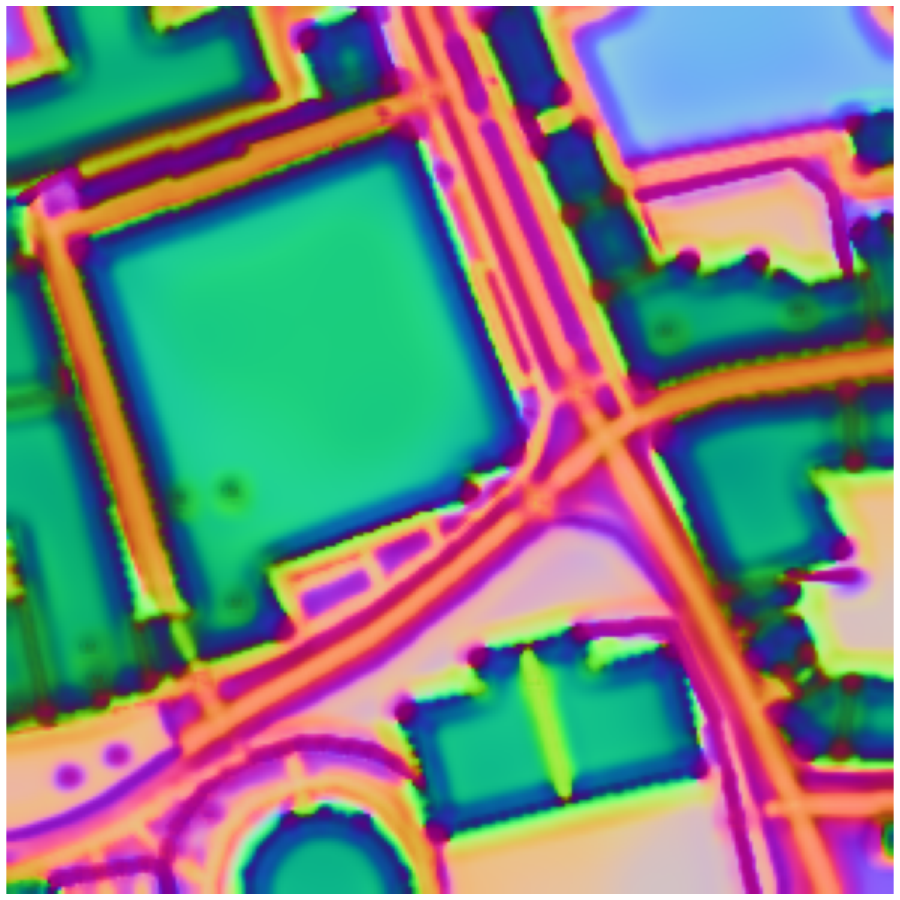} 
    & \includegraphics[valign=m,height=\plotH]{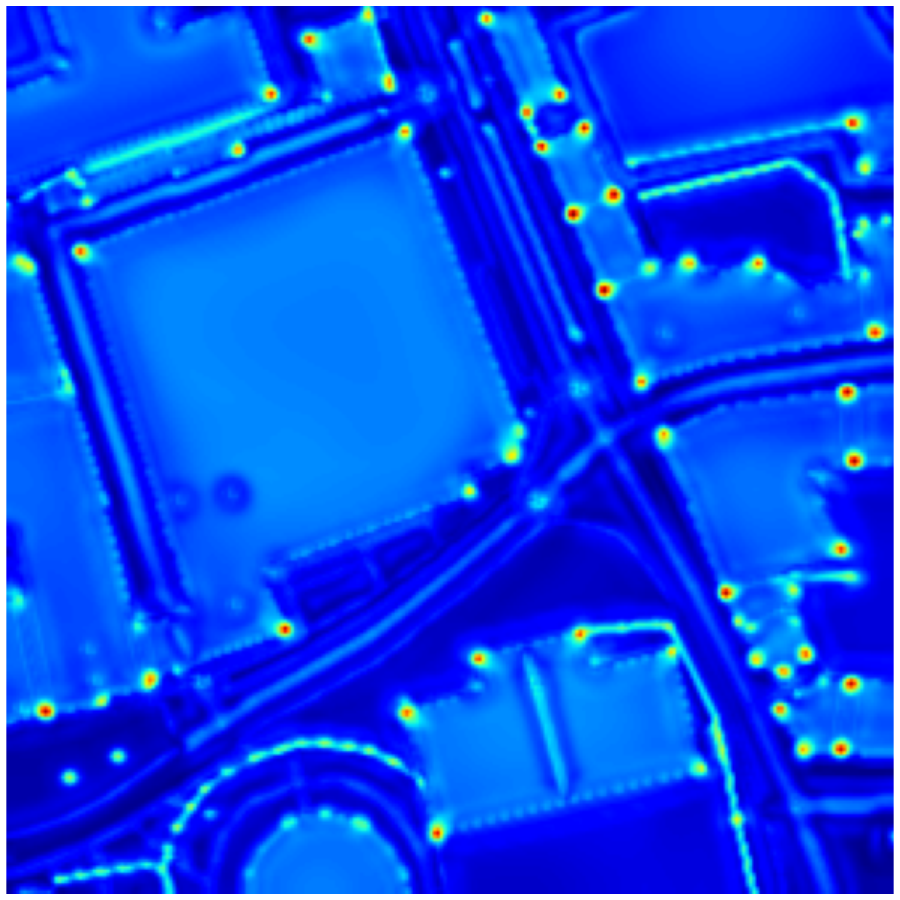} 
    & \includegraphics[valign=m,height=\plotH]{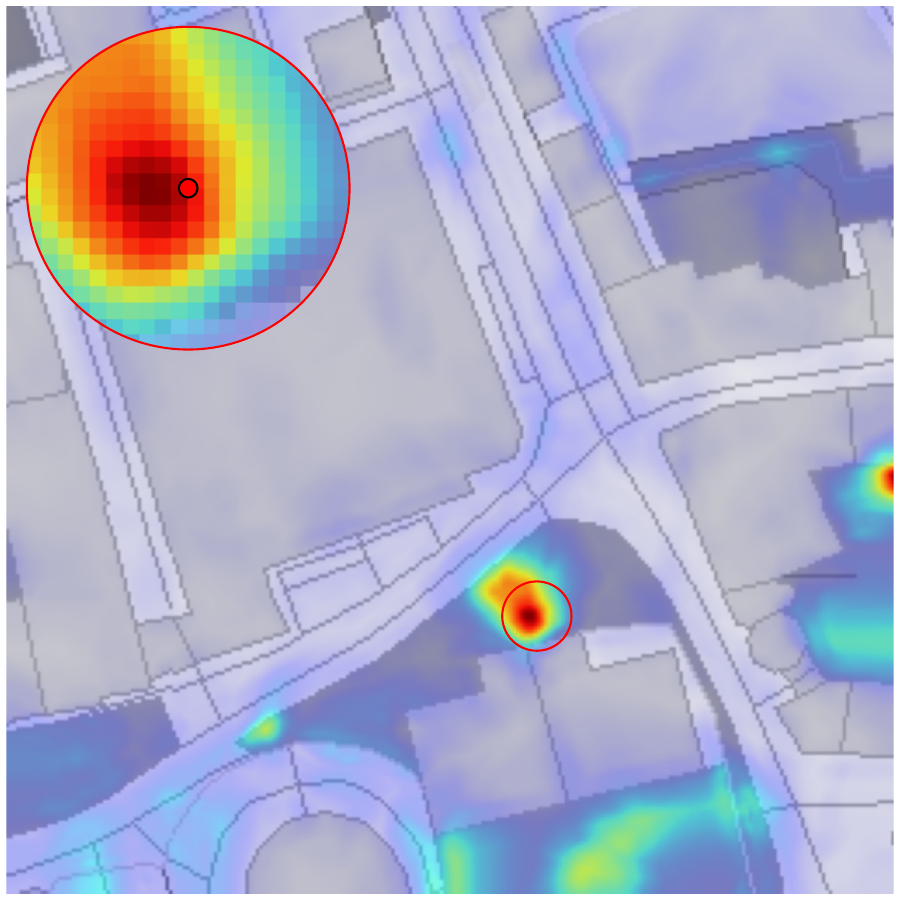} \\
    OrienterNet \cite{sarlin2023orienternet}
    & \includegraphics[valign=m,height=\plotH]{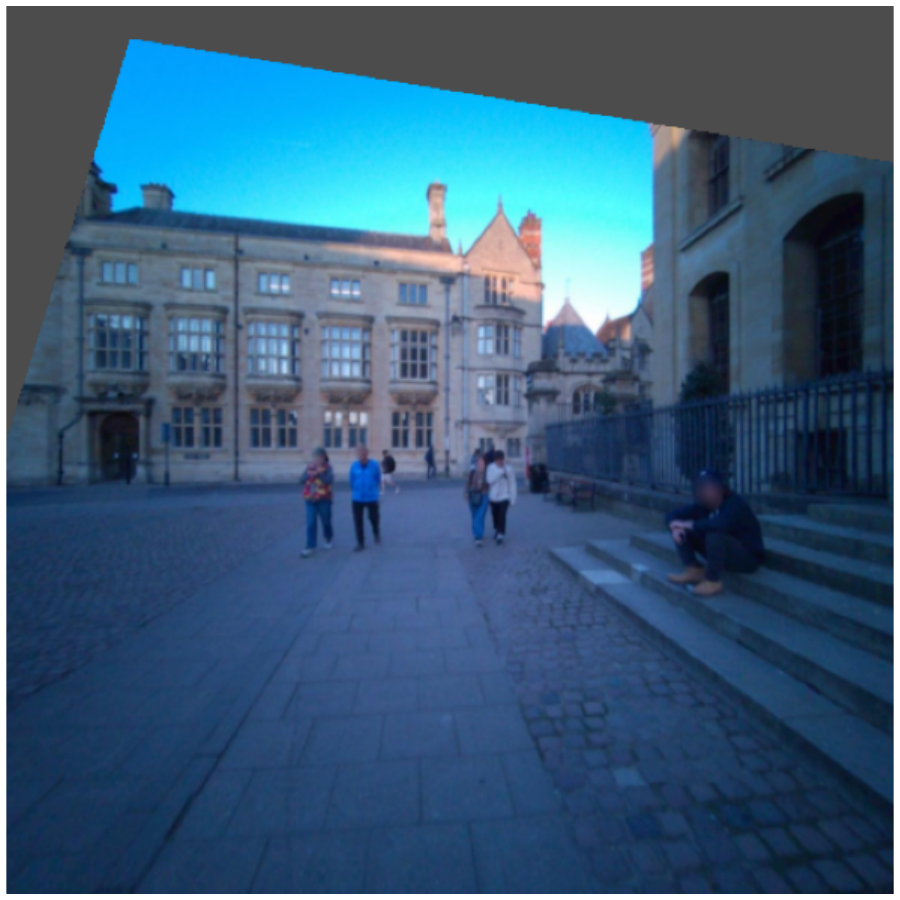} 
    & \includegraphics[valign=m,height=\plotH]{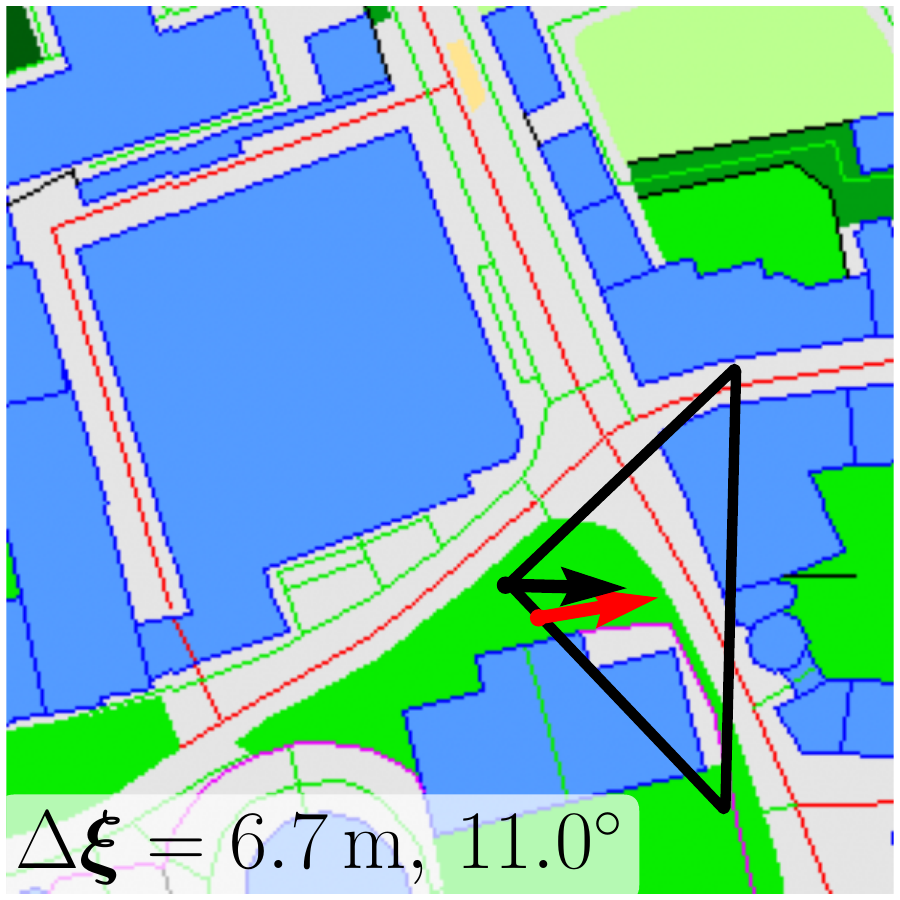} 
    & \includegraphics[valign=m,height=\plotH]{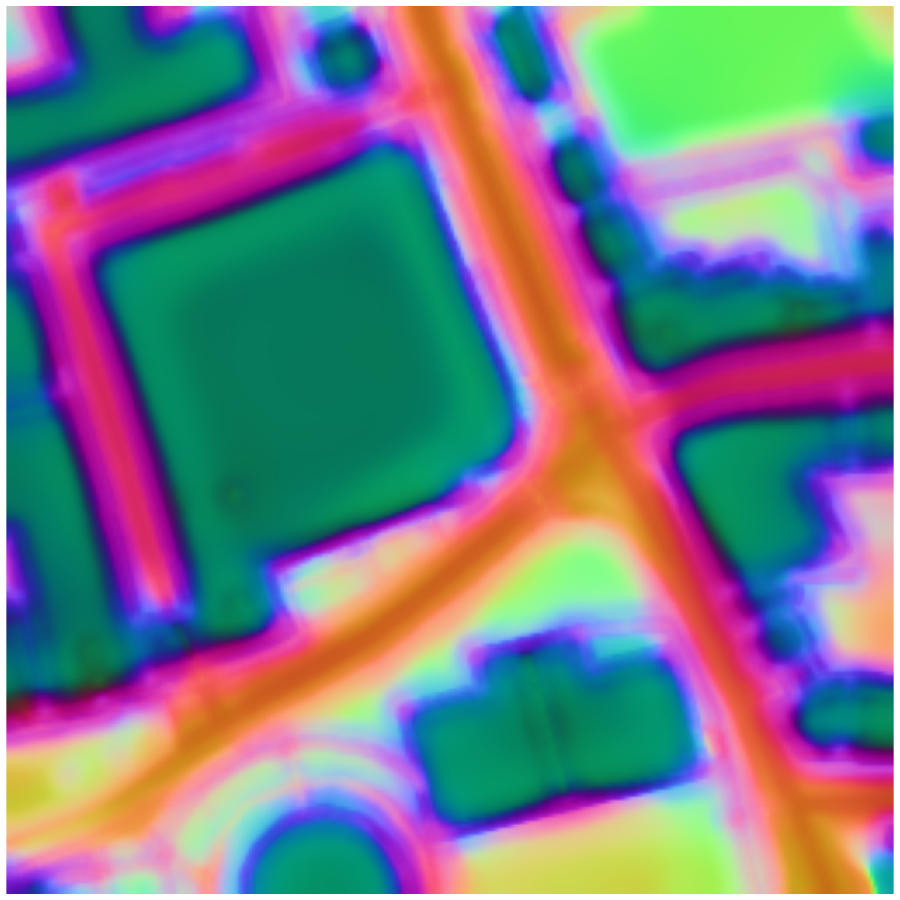} 
    & \includegraphics[valign=m,height=\plotH]{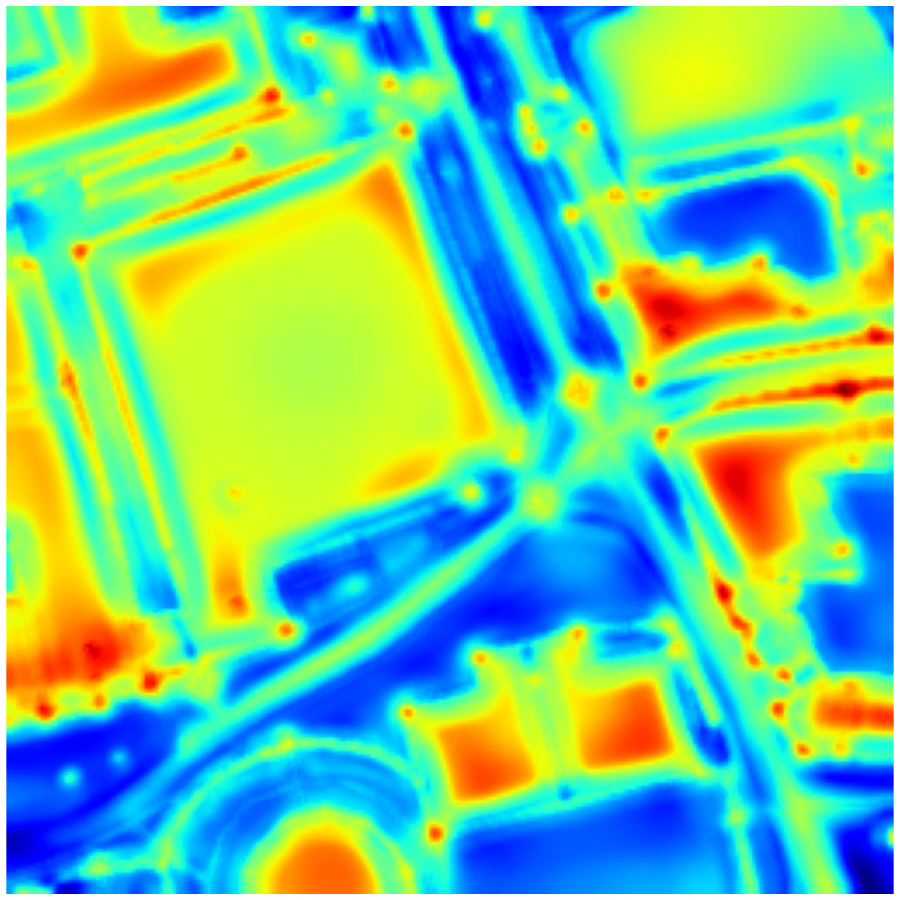} 
    & \includegraphics[valign=m,height=\plotH]{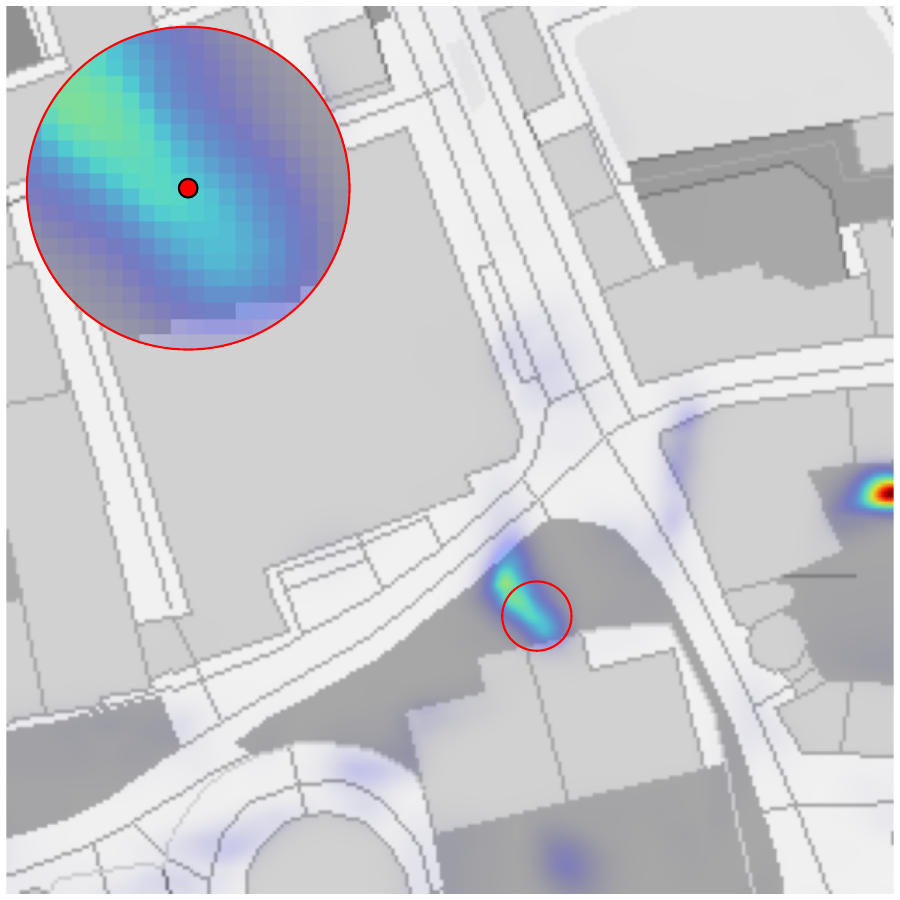} \\
    \name
    & \includegraphics[valign=m,height=\plotH]{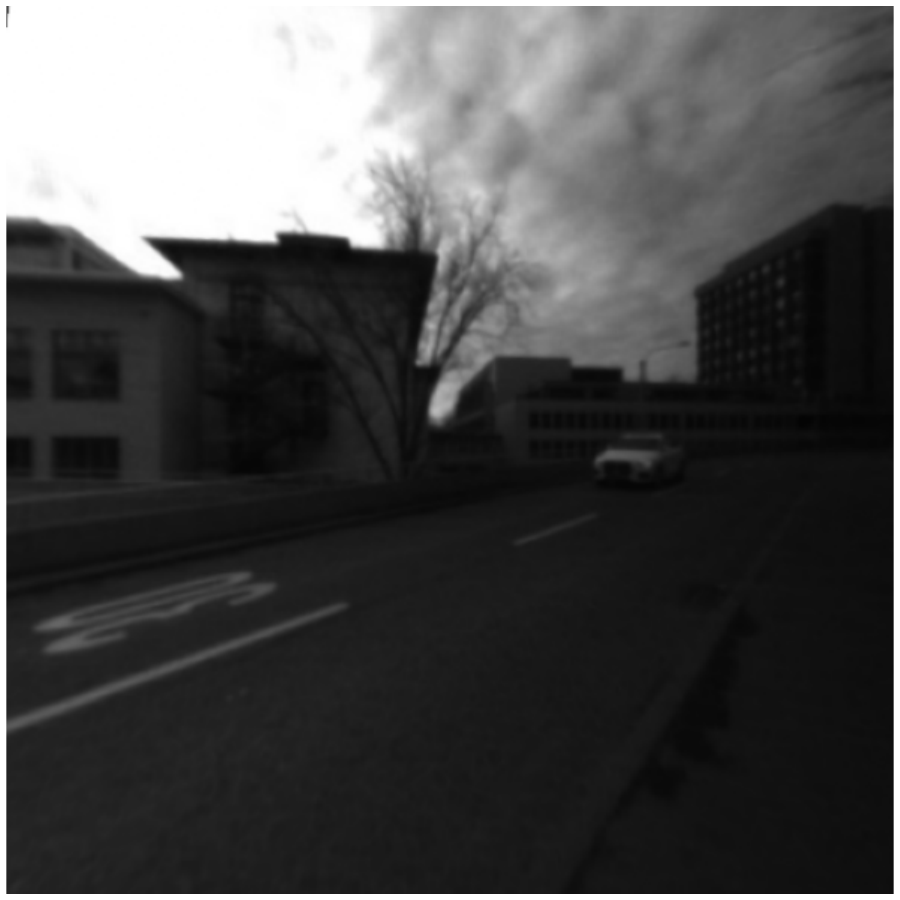} 
    & \includegraphics[valign=m,height=\plotH]{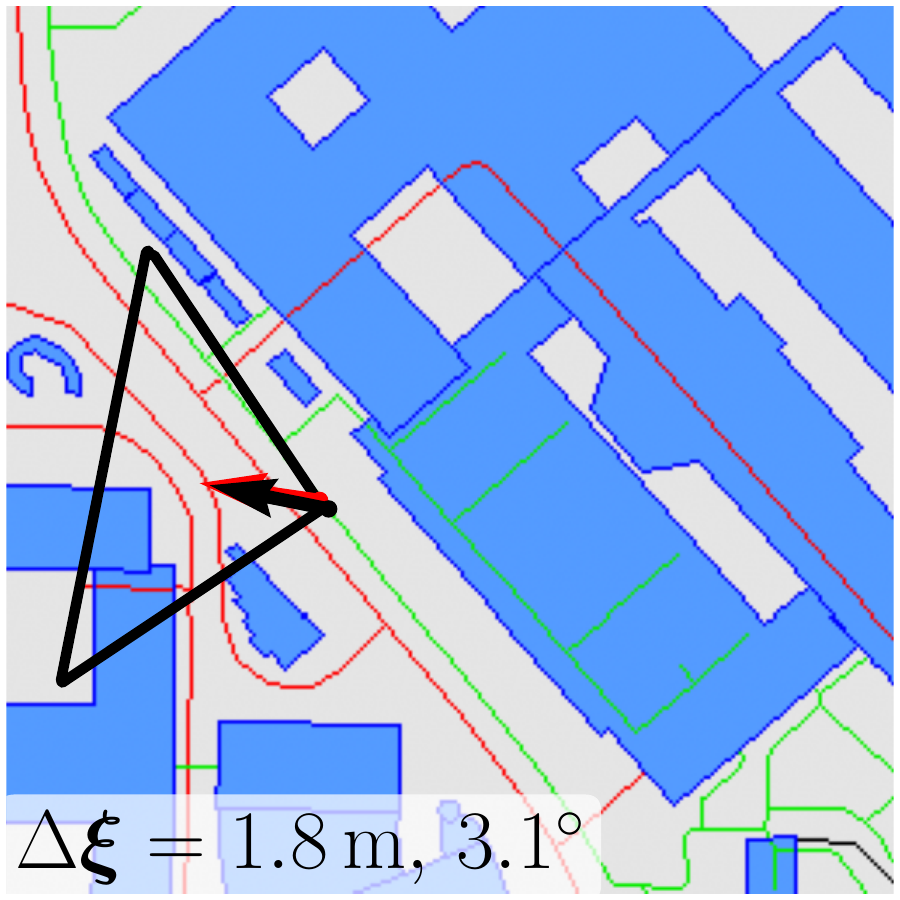} 
    & \includegraphics[valign=m,height=\plotH]{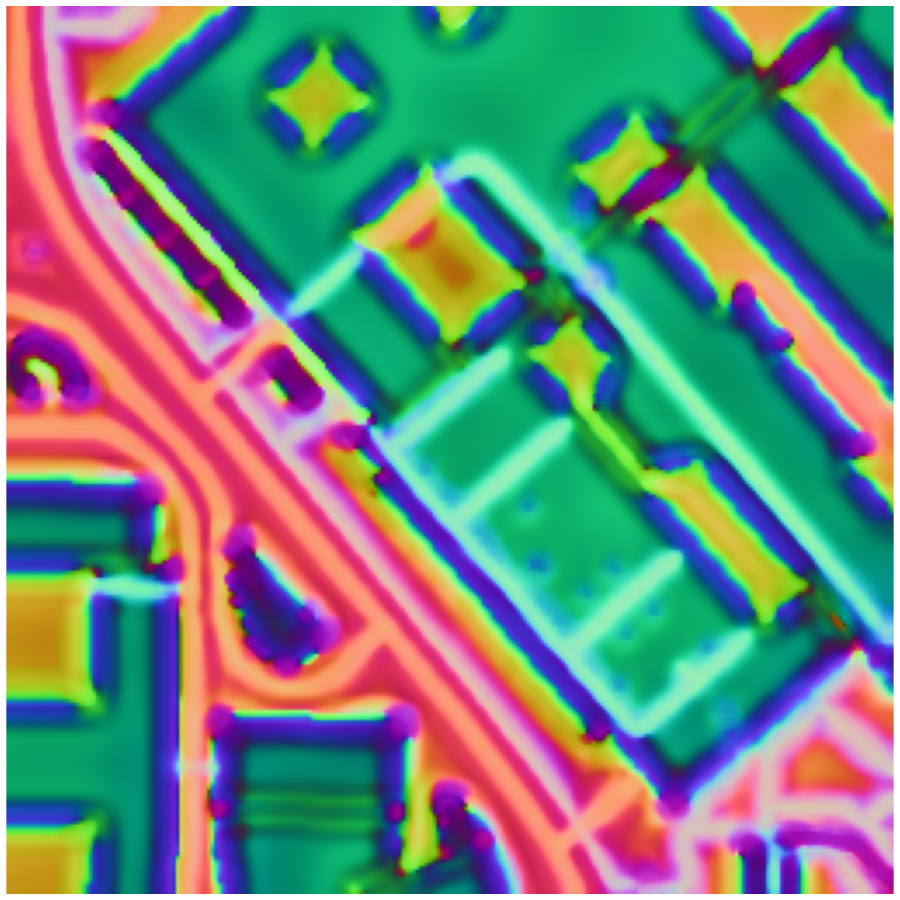} 
    & \includegraphics[valign=m,height=\plotH]{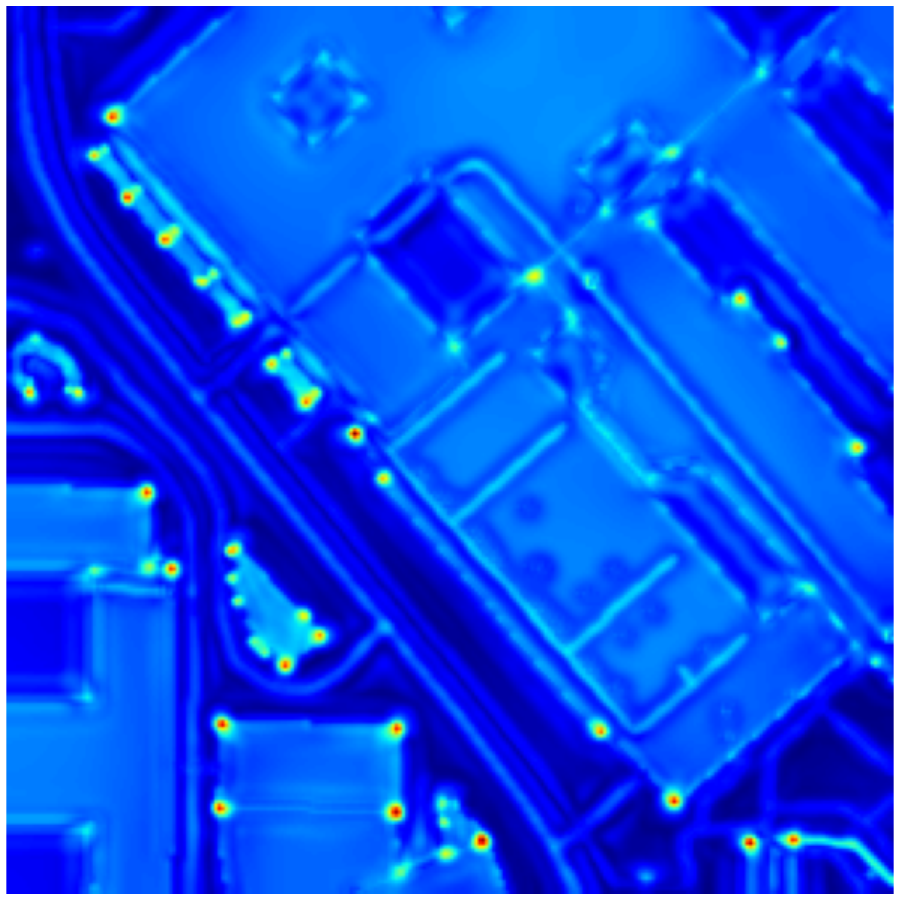} 
    & \includegraphics[valign=m,height=\plotH]{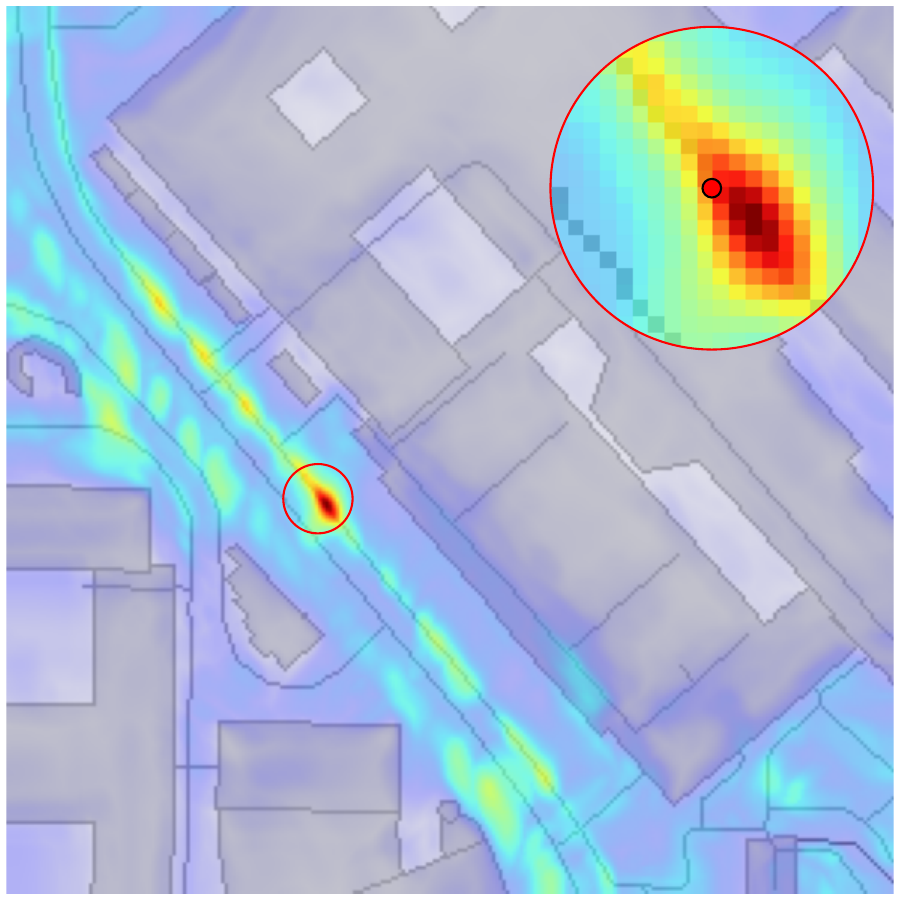} \\
    OrienterNet \cite{sarlin2023orienternet}
    & \includegraphics[valign=m,height=\plotH]{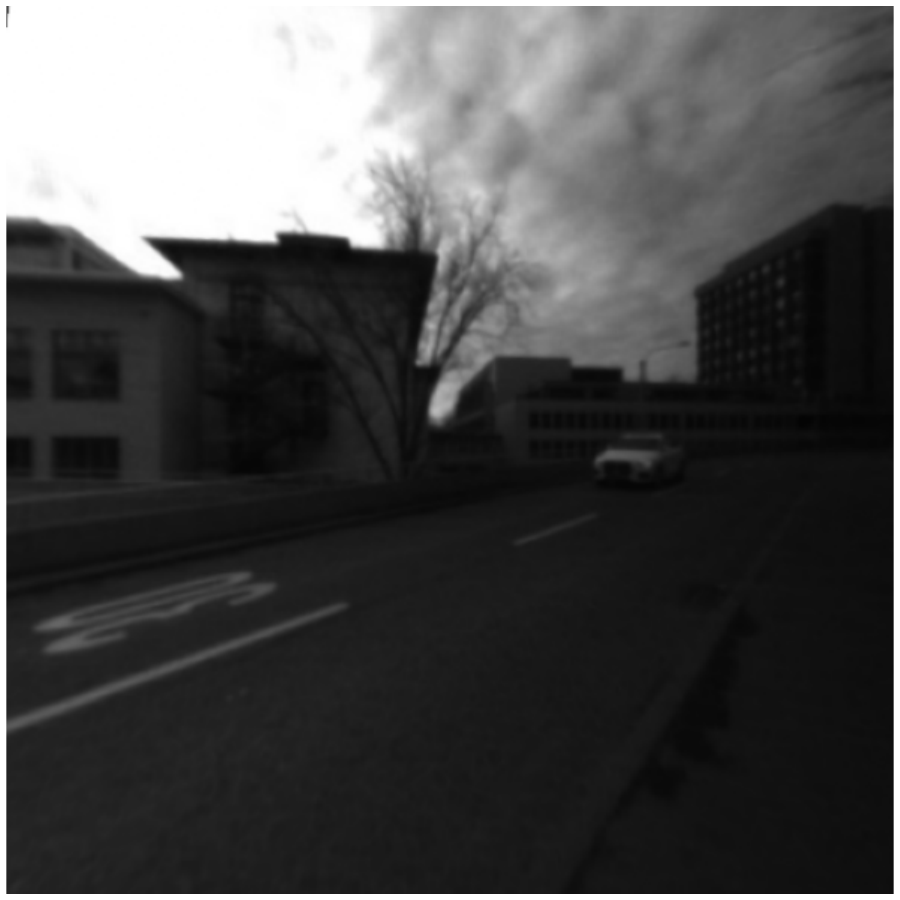} 
    & \includegraphics[valign=m,height=\plotH]{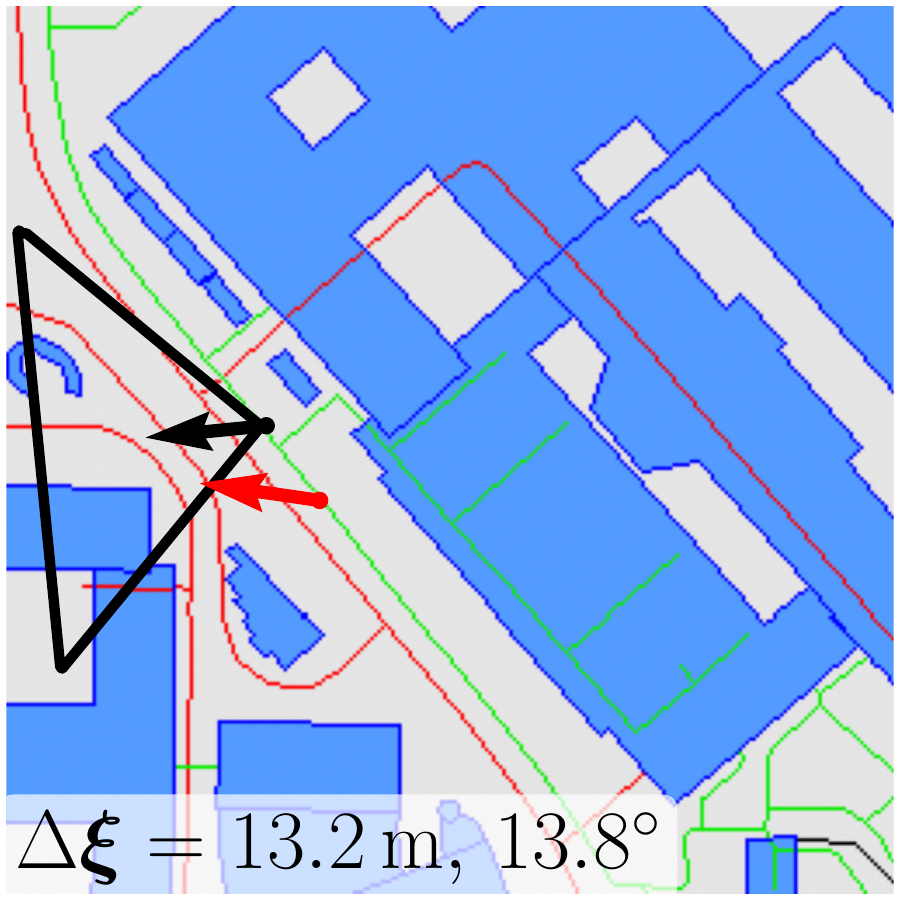} 
    & \includegraphics[valign=m,height=\plotH]{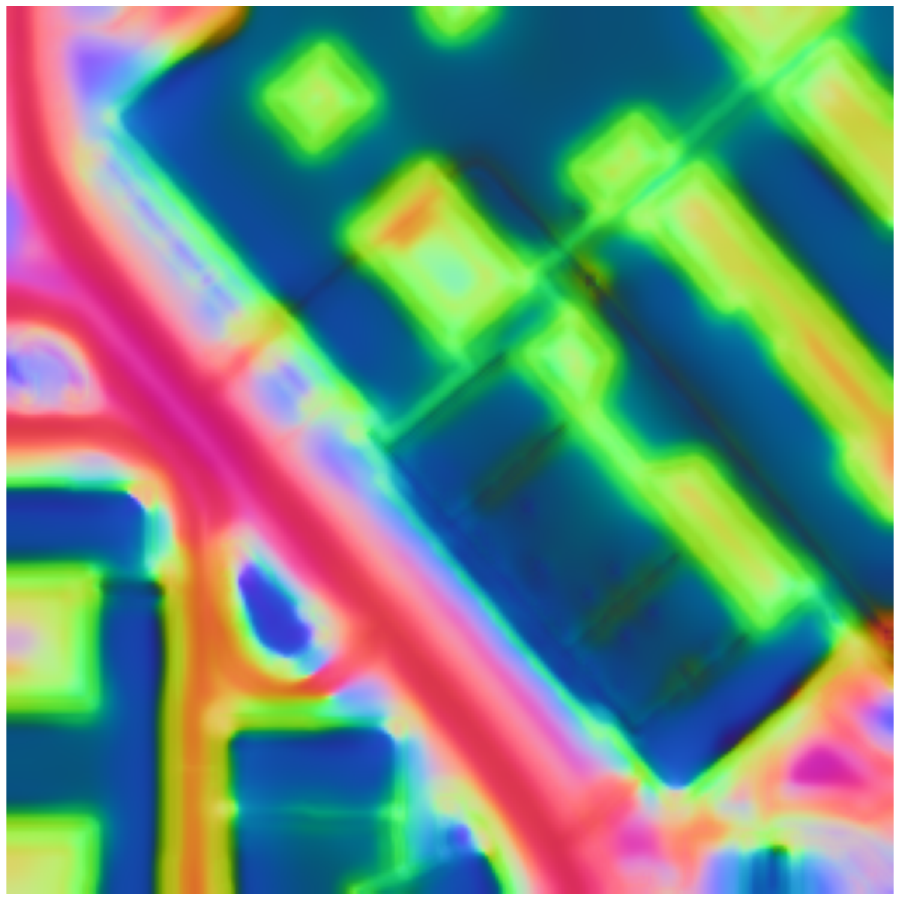} 
    & \includegraphics[valign=m,height=\plotH]{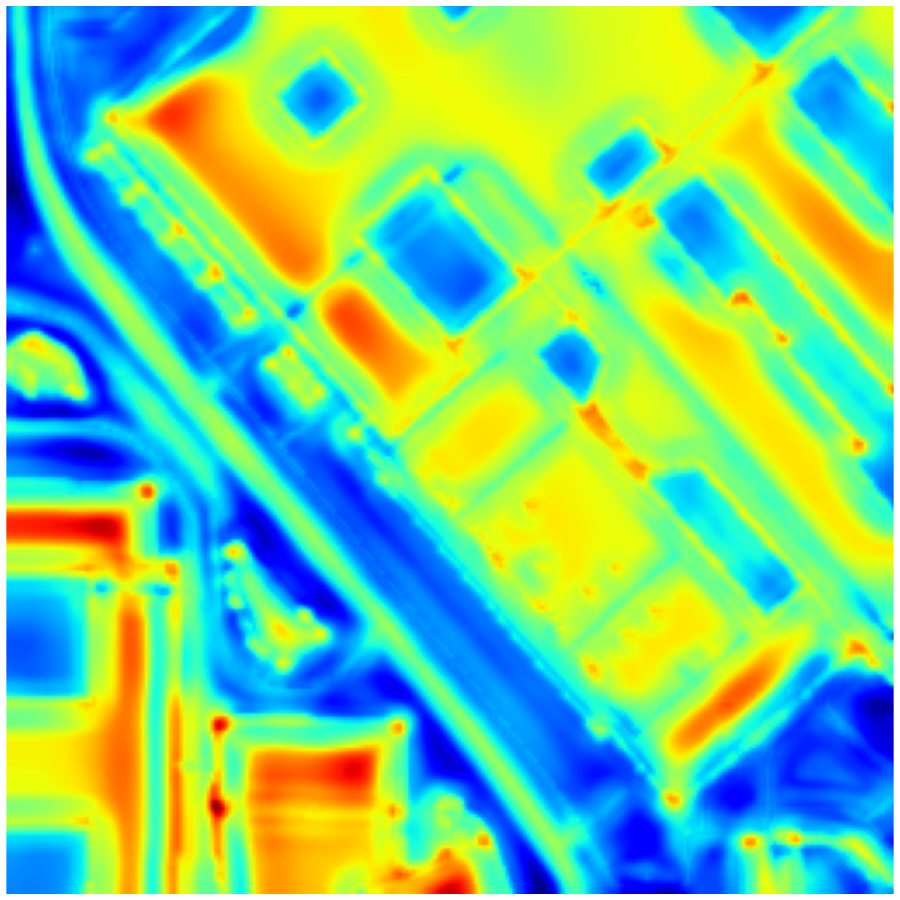} 
    & \includegraphics[valign=m,height=\plotH]{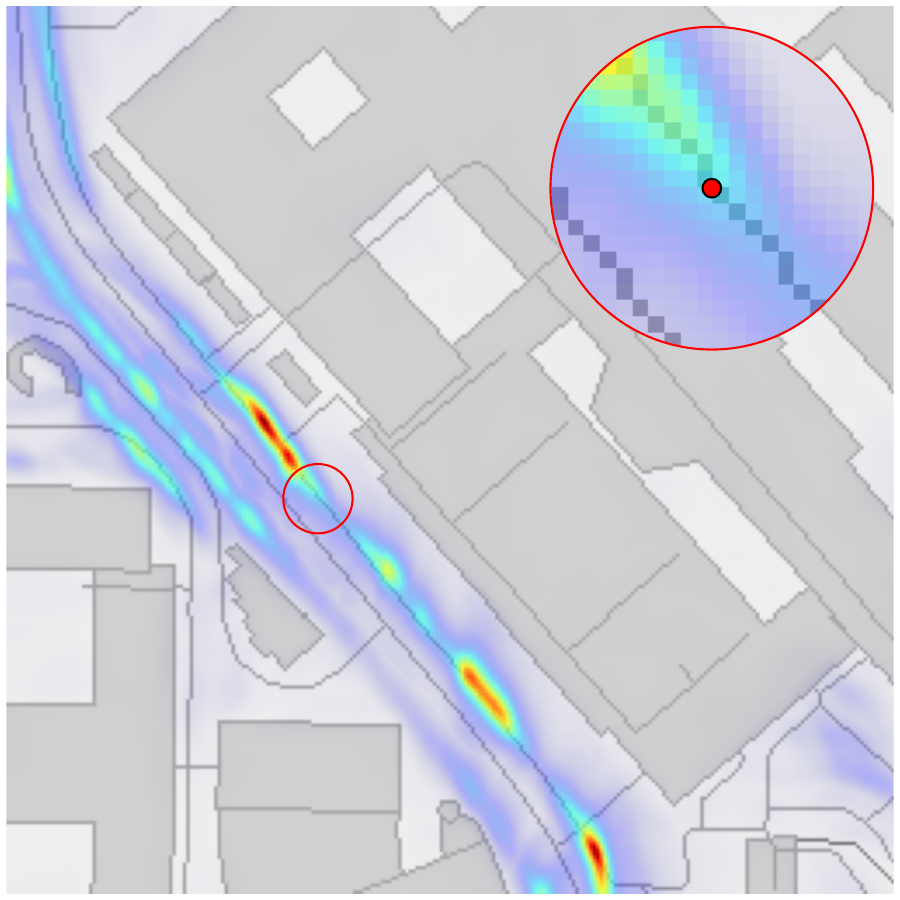}
    \end{tblr}
    \vspace{-0.3cm}
    \caption{\textbf{Qualitative results.} To accommodate inaccuracies in the training GT, a strongly supervised OrienterNet \cite{sarlin2023orienternet} 
    learns smooth map features, and focuses on large regions useful for coarse localization, such as entire buildings (see feature norms).
    In contrast, \name learns much sharper features, focusing on keypoints visible in the images, such as building corners, more useful for fine-grained visual localization.%
    }
    \label{fig:qualitative}
\end{figure}

\PAR{Results} \cref{tab:sequ_results_main} shows that \name generalizes well to urban scenes not seen during training and outperforms alternative approaches, even when they are trained on more data. 
Qualitative comparisons to OrienterNet are shown in \cref{fig:qualitative}. 
Using DINOv2 as the image backbone consistently improves over using a ResNet-101 backbone.
Notably, on LaMAria, \name improves over GPS at strict error thresholds while achieving comparable performance at coarse error thresholds. In the following, we show that sequential fusion of single-view estimates yields a clear advantage over GPS in both accuracy and robustness.

\subsection{Sequential evaluation}\label{subsec:exp_seq}
Given relative pose estimates $\hat{\pose}_{j,i}$ between views $j$ and $i$
(\eg, from VIO/SLAM \cite{mur2017orb2,campos2021orb3,engel2023aria}),
we can aggregate multi-view information for higher accuracy.
Like~\cite{sarlin2023orienternet}, we compute the posterior $p(\pose_0\,|\,\mathbf{I}_{0:t}, \mathrm{map}_{0:t}, \hat{\pose})$ over the pose $\pose_0$
at the first time step, given images $\mathbf{I}_{0:t}$, map tiles $\mathrm{map}_{0:t}$, and relative poses $\hat{\pose}$ up to $t$.

\PAR{Setup} We use the same splits as in the previous single-view benchmarks, sample contiguous datapoints in time, and vary the sequence length. We keep the same search radius for the poses as in their single-view evaluations. 
Recall metrics correspond to error differences between the ground-truth and the aligned input trajectory with the transform derived from the smoothed estimate of $\pose_0$ \cite{sarlin2023orienternet}.

\begin{figure}[t]
\scriptsize

\newcommand{\legendline}[2]{\tikz[baseline=-0.6ex]\draw[#1,line width=0.9pt] (0,0)--(1.2em,0) node[pos=1,anchor=west] {#2};}
\newcommand{\legendonly}[1]{\tikz[baseline=-0.6ex]\draw[#1,line width=0.9pt] (0,0)--(1.2em,0);}

\centering

\begin{tblr}{
  colspec={cccc},
  colsep=1pt,
  rowsep=0pt,
  cell{1,2}{1}={c=3}{c},
  cell{5}{1}={c=3}{c},
  hspan=even, 
  cell{3}{4}={r=2}{c},
  column{3} = { rightsep+=10pt },
}
{\scriptsize
\legendonly{black}\,Recall@1m/\degree
\quad
\legendonly{black,dashed}\,Recall@5m/\degree}
&&& {\scriptsize Average Recall} \\
{\scriptsize
\legendline{orange}{OrienterNet \cite{sarlin2023orienternet}}
\legendline{gray2}{GPS}
\legendline{blue}{Ours (rel. poses)}
\legendline{ForestGreen}{Ours (GPS)}} 
&&& 
{\scriptsize Across Benchmarks} \\
\includegraphics[height=4.3cm]{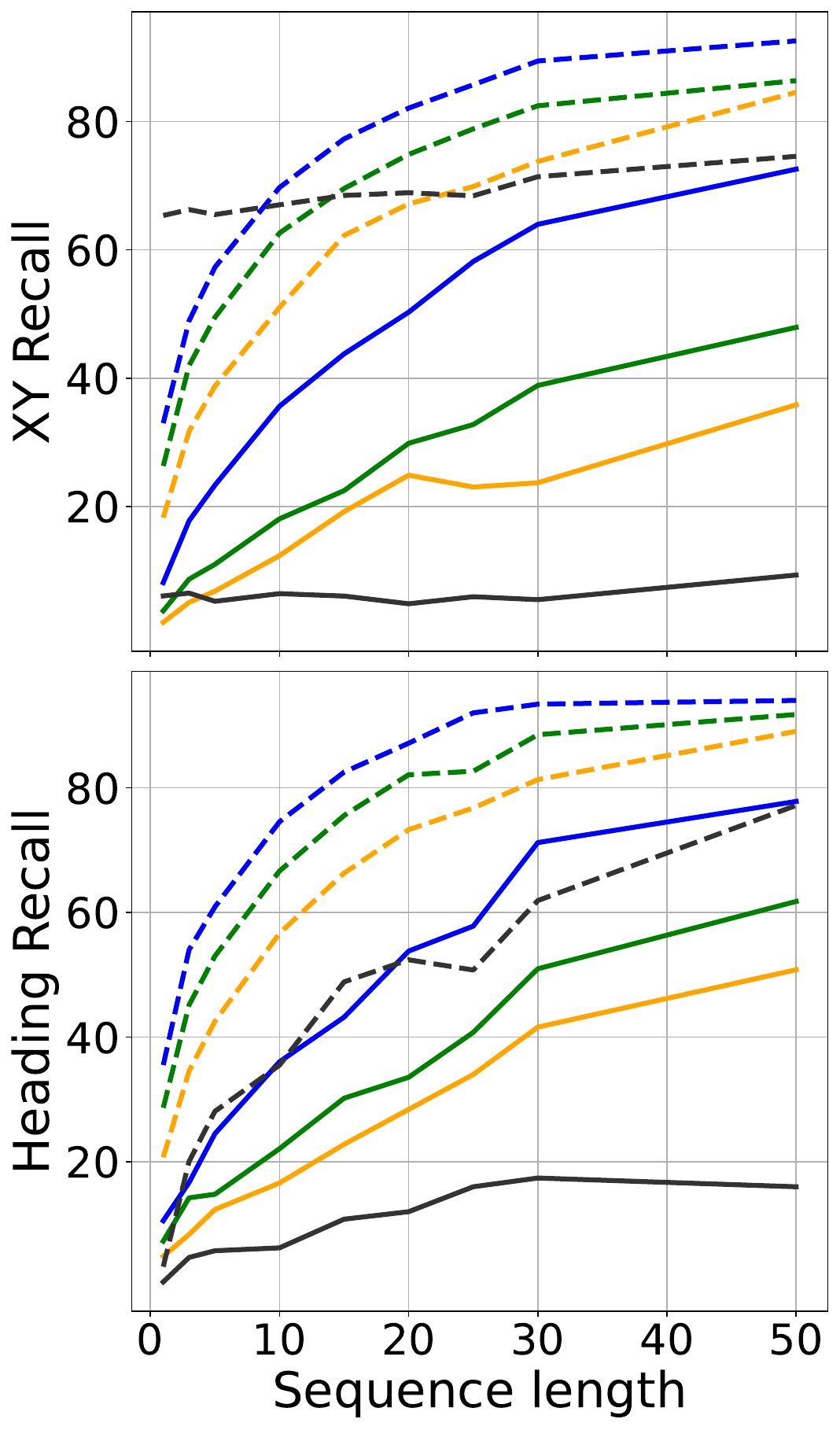} &
\includegraphics[height=4.3cm]{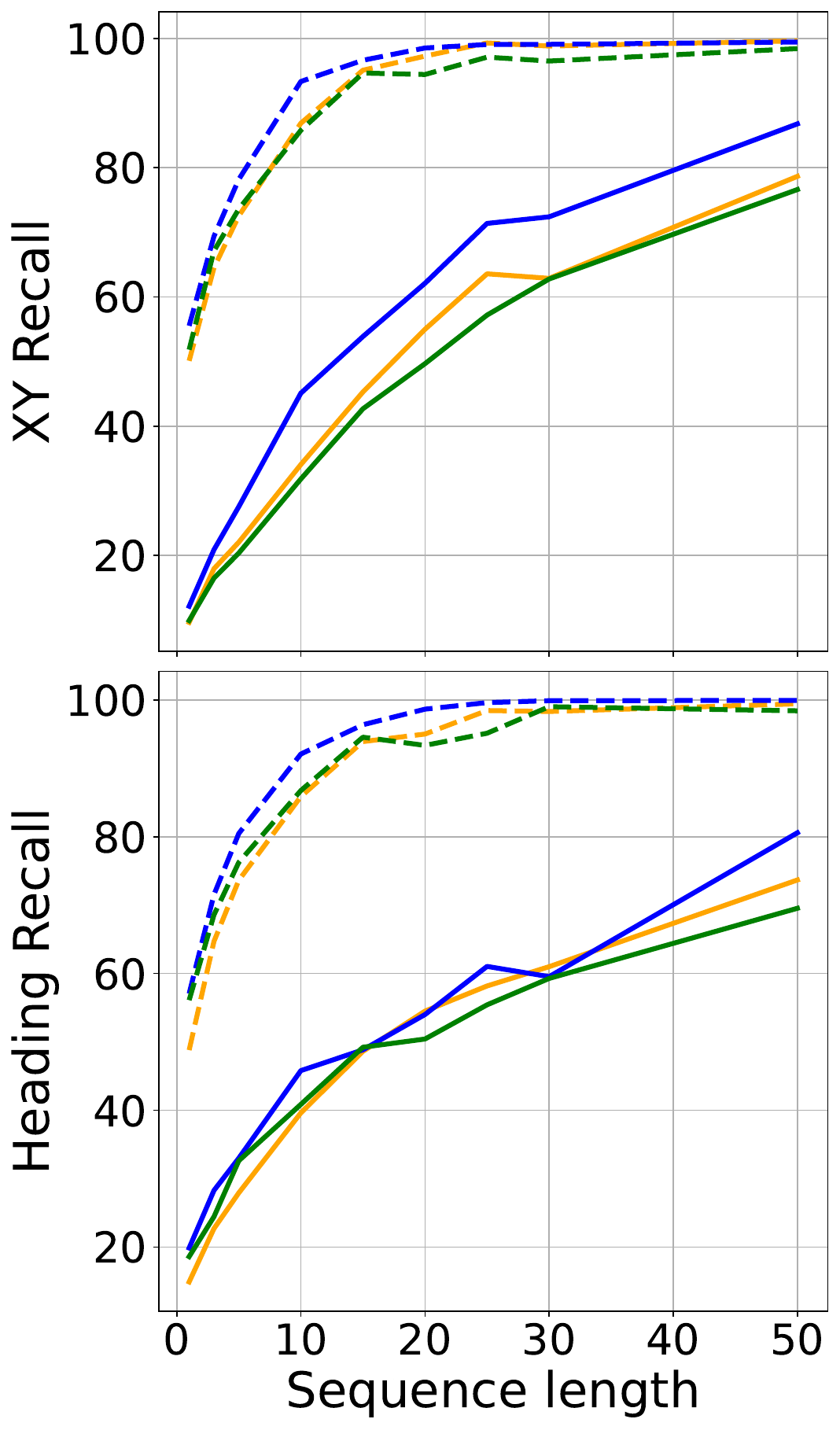} &
\includegraphics[height=4.3cm]{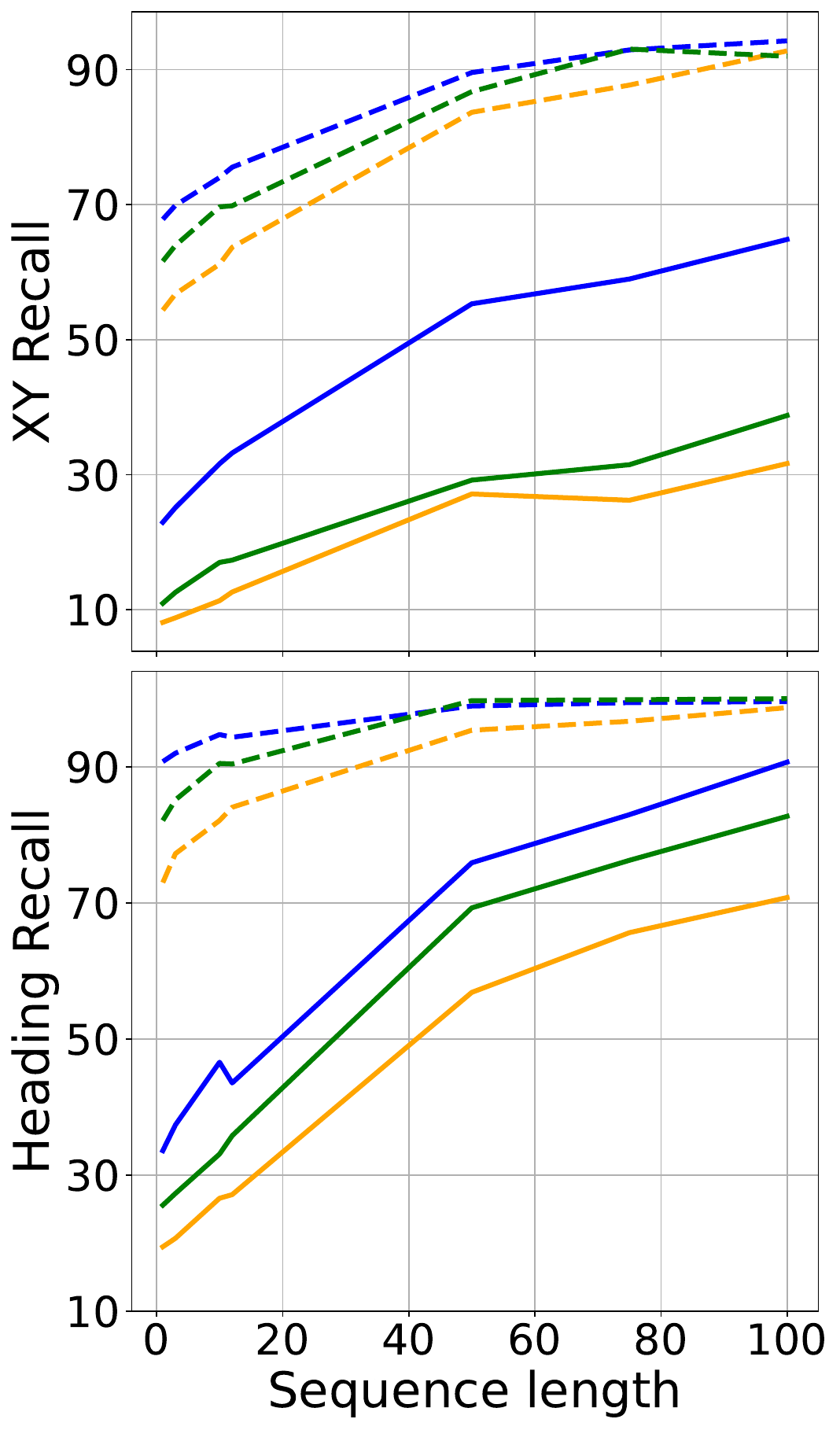} &
\includegraphics[height=4.3cm]{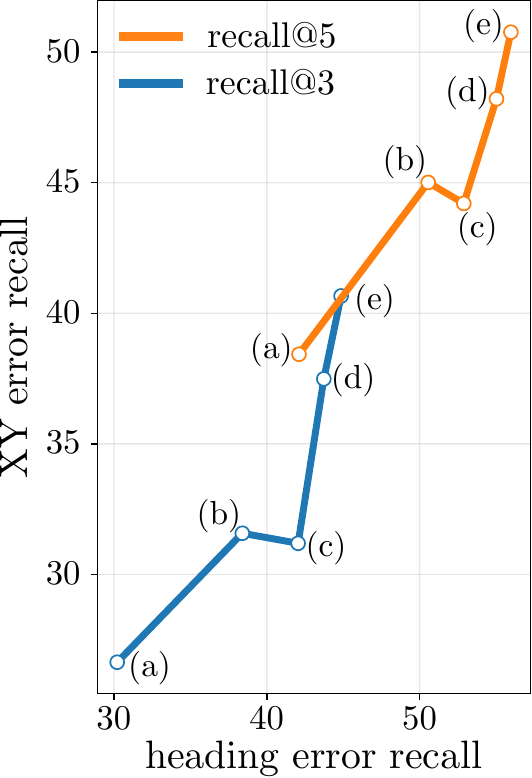} \\
LaMAria \cite{Krishnan2025lamaria} &
ODN \cite{wang2025oxford} &
KITTI \cite{geiger2012kitti} &
Ablation (single-view) \\
{\footnotesize \textbf{Accuracy vs.~Sequence Length}}
&&&
{\footnotesize \textbf{Supervision Impact}}
\end{tblr}
\caption{
\textbf{Left:} \name accuracy scales well with sequence length and improves faster than OrienterNet~\cite{sarlin2023orienternet}.
\textbf{Right:} Weak labels outperform OrienterNet and their performance correlates with label informativeness. Each point corresponds to
(a) OrienterNet,
(b) GPS labels,
(c) GPS labels with non-metric relative poses,
(d) metric relative poses, and
(e) approximately geo-referenced relative poses.
The losses for (b)--(e), respectively, are
$\mathcal{L}_{\mathrm{GPS},\mathrm{chunk}}$, see \cref{eq:loss_xy_chunk};
$\mathcal{L}_{\mathrm{GPS},\mathrm{chunk}} + 0.5\mathcal{L}_{\Delta\Theta}$, see \cref{eq:loss_relrot};
$0.1\mathcal{L}_{\Delta\Theta} + \mathcal{L}_{\lVert \Delta \bt \rVert}$, see \cref{eq:loss_chunked_norm};
and
$0.1\mathcal{L}_{\Delta\Theta} + \mathcal{L}_{\Delta XY}$, see \cref{eq:loss_reltrans}.
All results use a ResNet-101 backbone, with chunk sizes $r = 20$\,m in \cref{eq:loss_xy_chunk} and $\rdt = 5$\,m in \cref{eq:loss_chunked_norm}.
}
\label{fig:ablation}
\end{figure}

\PAR{Results} The results in \cref{tab:sequ_results_main} and \cref{fig:ablation} (Left) confirm that the more accurate single-view estimates achieved with our supervision are beneficial for sequential fusion. On the challenging LaMAria~\cite{Krishnan2025lamaria}, fusing under 10 frames allows us to surpass the high recall@5m of GPS, which benefits less due to its biased measurements (as similarly found in~\cite{sarlin2023orienternet} and exemplified in 
\cref{fig:supp_lama_filt}). 
A retrained OrienterNet on the same data requires over 20 frames to achieve this. 
This behavior is consistent across different egocentric~\cite{wang2025oxford} and driving~\cite{geiger2012kitti} benchmarks.

\subsection{Analyzing supervision types}\label{subsec:exp_ablation}
As shown in \cref{fig:ablation} (Right), different types of weak labels can train accurate visual localization models. GPS provides noisy, potentially erroneous location estimates, to which we are robust by using a tolerance radius $r$ (\cref{eq:loss_xy_chunk}) of up to $20$\,m. By comparison, relative poses provide a more accurate training signal, with performance improving as more information becomes available: non-metric relative poses already improve rotation accuracy by supervising the relative rotation, while position accuracy improves when using metric or coarsely geo-referenced relative translations, with the latter yielding the best overall performance.

\section{Conclusion}
We have presented \name, a novel approach for 3-DoF visual localization on 2D maps. 
\name tackles the problem of inaccurate geo-referenced training data via weak supervision, and requires only 2D GPS labels or relative poses, easier to obtain at scale than accurate geo-referenced 3-DoF poses.
\name is robust to several types of geo-referencing errors, including GPS noise and absolute pose offsets. 
Experimentally, \name consistently outperforms strongly supervised methods across driving and egocentric benchmarks.

\section*{Acknowledgements}
The authors sincerely thank Zirui Wang for generously providing assistance with the Oxford Day-and-Night benchmark.
Javier Tirado-Garín is funded by the scholarship FPU21/04468.

\bibliographystyle{splncs04}
\bibliography{main}

\clearpage

\setcounter{section}{0}
\renewcommand\thesection{\Alph{section}}

\renewcommand{\theHsection}{supp.\Alph{section}}
\renewcommand{\theHsubsection}{supp.\Alph{section}.\arabic{subsection}}
\renewcommand{\theHsubsubsection}{supp.\Alph{section}.\arabic{subsection}.\arabic{subsubsection}}

\section*{Supplementary Material}

\cref{sec:supp_training_details,sec:supp_eval_details} provide additional training and evaluation details, respectively, and \cref{sec:supp_quantitative,sec:supp_qualitative} provide additional quantitative and qualitative results.

\section{Additional training details}\label{sec:supp_training_details}

\PAR{Tile sampling}
The Mapillary Geo-Localization (MGL) dataset \cite{sarlin2023orienternet}, used for training, provides two types of geo-referenced position coordinates for each input image: (1) the \emph{raw} GPS coordinates, and (2) refined coordinates obtained through SfM by fusing multi-view visual observations with the raw GPS measurements. During training, we use the raw GPS coordinates to center $128\times128$\,m OSM tiles. Following the official OrienterNet checkpoint \cite{sarlin2023orienternet}, we randomly perturb each tile center within $\pm48$\,m and apply random flips and rotations in $90\degree$ increments to reduce overfitting, augmenting images and poses accordingly. This preprocessing is performed independently for each tile.

\begin{figure}[t]
    \centering
    \includegraphics[width=1\linewidth]{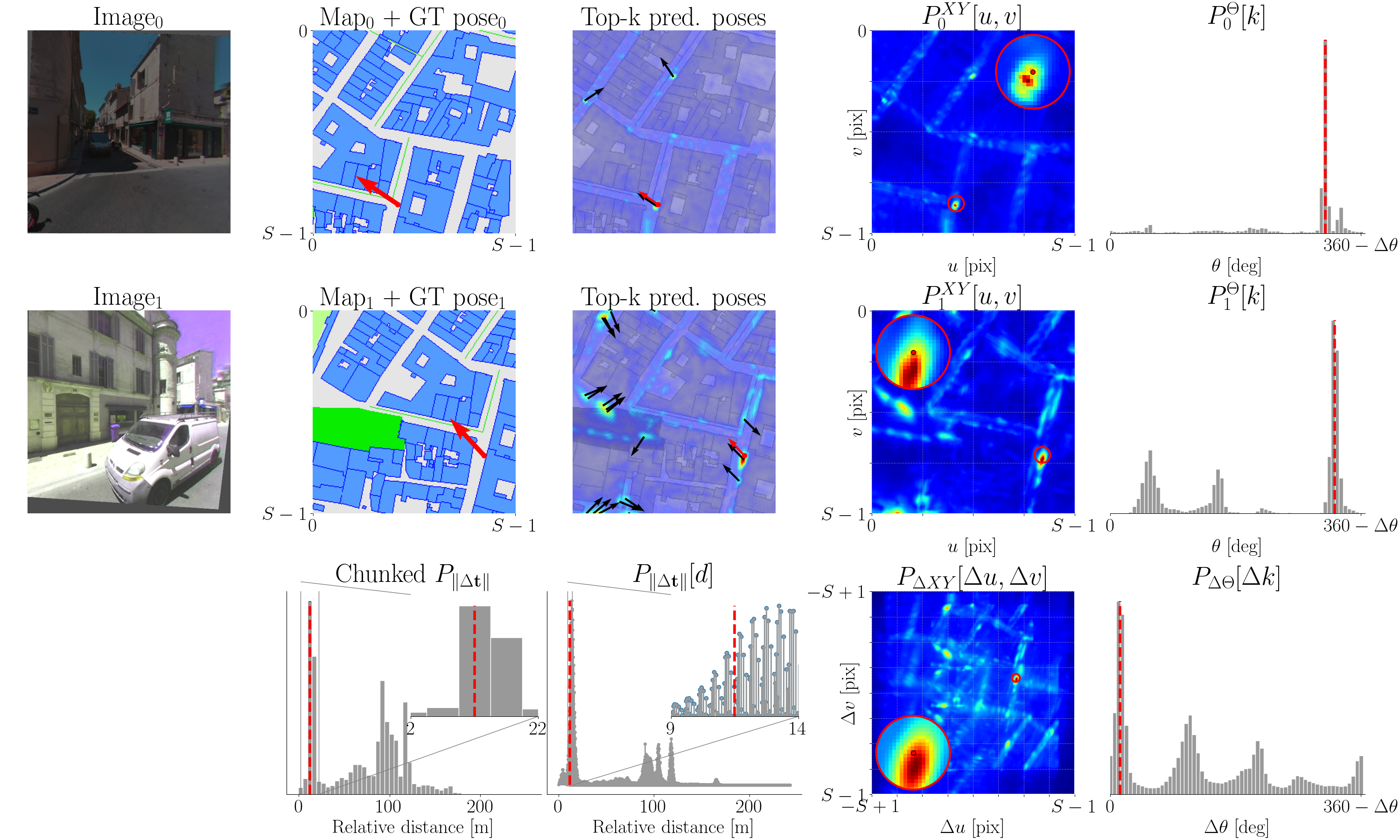}
    \caption{
    \textbf{Probability distributions.}
    We show a pair of training datapoints together with their predicted absolute translation ($P_i^{XY}[u,v]$, \cref{eq:trans_probs}) and rotation ($P_i^{\Theta}[k]$, \cref{eq:rot_probs}) distributions. The last row shows the relative pose distributions derived from the pair: the relative-distance distribution $P_{\lVert \Delta \bt \rVert}[d]$ (\cref{eq:norm_probs}), its chunked version, the relative-shift distribution $P_{\Delta XY}[\Delta u,\Delta v]$ (\cref{eq:reltrans_linxcorr}), and the relative-rotation distribution $P_{\Delta\Theta}[\Delta k]$ (\cref{eq:relrot_circxcorr}). Note that the raw distance distribution is non-smooth, which makes interpolation difficult. We therefore use its chunked version in \cref{eq:loss_chunked_norm}, which also adds robustness to errors in the ground-truth (GT) relative distances. For each distribution, we mark in red the values corresponding to the GT poses.}
    \label{fig:supp_prob_dists}
\end{figure}

\PAR{Relative poses}
For relative-pose supervision, we sample pairs of data points that were jointly optimized within the same SfM cluster. We identify these clusters using the \texttt{merge\_cc} attribute in the Mapillary metadata\footnote{\url{https://www.mapillary.com/developer/api-documentation}}. Since the two input tiles in a pair are augmented independently, we undo the corresponding flip and rotation augmentations for each predicted probability volume before computing the relative-pose probability distributions (\cref{subsec:train_relposes}). This ensures that the resulting joint probabilities are defined in a consistent, non-augmented coordinate frame. Examples of relative-pose probability distributions supervised during training are shown in \cref{fig:supp_prob_dists}.

\PAR{Images}
Following \cite{sarlin2023orienternet}, we apply color jitter augmentation to the input images and resize them to $512\!\times\!512$ when using a ResNet \cite{he2016deep} image backbone. 
Most images in MGL are square perspective crops from panoramic images with a $90\degree$ field of view (FoV), resulting in an average focal length of approximately 256 pixels. When using a DINOv2 \cite{oquab2024dinov2} backbone, we follow the same procedure but resize images to $518\times518$ so that their dimensions are divisible by the patch size used in DINOv2.

\PAR{Architecture and hyperparameters}
Following \cite{sarlin2023orienternet}, we use $N=64$ discrete rotations during training and keep the same architecture design, including a $64.5\times32$\,m BEV extent with a resolution of $0.5$\,m/pixel for both the BEV and the map tile. For our checkpoints, as well as our retrained OrienterNet baselines using a ViT backbone pretrained by DINOv2, we fine-tune this backbone with a learning rate of $10^{-7}$. We combine it with a DPT decoder~\cite{ranftl2021dpt}, trained from scratch with a learning rate of $10^{-5}$. We use gradient clipping with a maximum norm of $1$ and keep the same U-Net with VGG-19 map encoder architecture and learning rate, $10^{-4}$, as in our other checkpoints.

\begin{figure}[t]
    \def\mapH{5.7cm}
    \def\imW{2cm}
    \centering
    \begin{tblr}{
        colspec={*{3}c},
        colsep=1pt,
        row{Z}={cmd={\scriptsize}},
    }
        \includegraphics[height=\mapH]{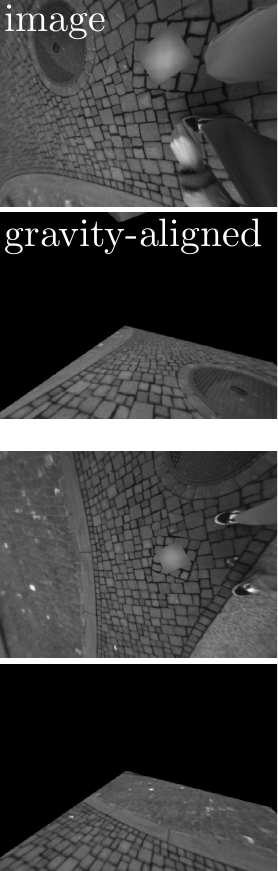}
        & \includegraphics[height=\mapH]{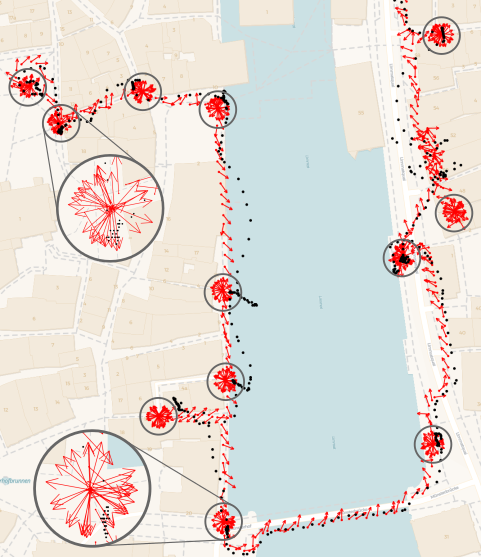} 
        & \includegraphics[height=\mapH]{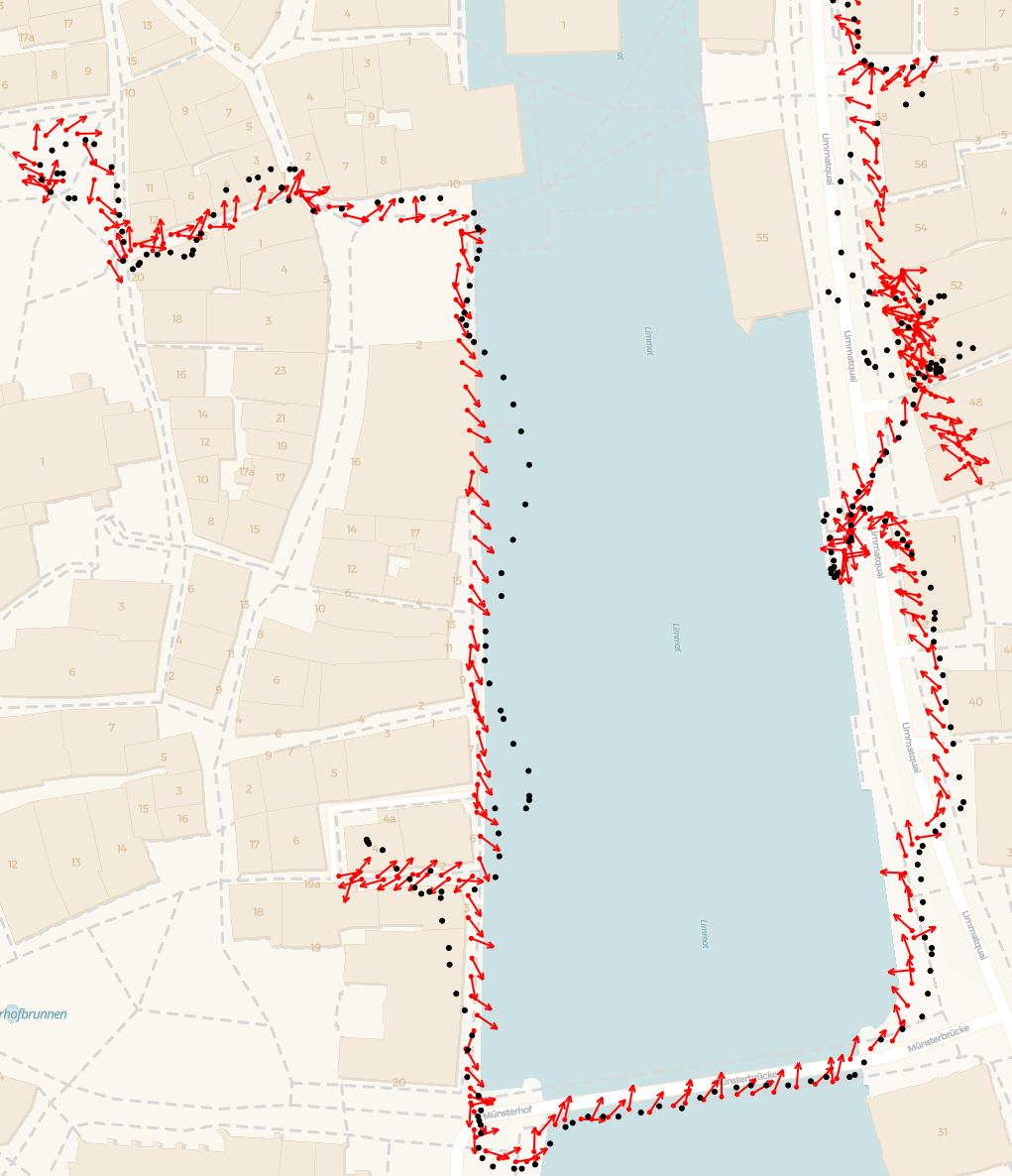} \\
        Control points & \texttt{sequence\_1\_20} & Filtered \texttt{sequence\_1\_20}
    \end{tblr}
    \caption{\textbf{Evaluation on LaMAria \cite{Krishnan2025lamaria}.} We exclude clips recorded while circling around ground control points (see ground-truth poses indicated with red arrows~\includegraphics[width=2.25mm]{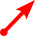}), whose locations correspond to that of the blurred AprilTags~\cite{Edwin2011apriltags} appearing in the images. The LaMAria benchmark also provides GPS measurements (shown as black dots,~$\bullet$), which, as can be seen, are biased with consistent errors over time.}
    \label{fig:supp_lama_filt}
\end{figure}

\begin{figure}[t]
    \centering
    \begin{tblr}{
        colspec={c},
        width=\linewidth,
        rowsep=1pt,
        columns={leftsep=0pt,rightsep=0pt},
    }
    Ground-truth \includegraphics[width=2mm]{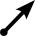} ~~~ Aligned ground-truth \includegraphics[width=2mm]{figures/qual/red_arrow.png} \\
    \includegraphics[width=1\textwidth]{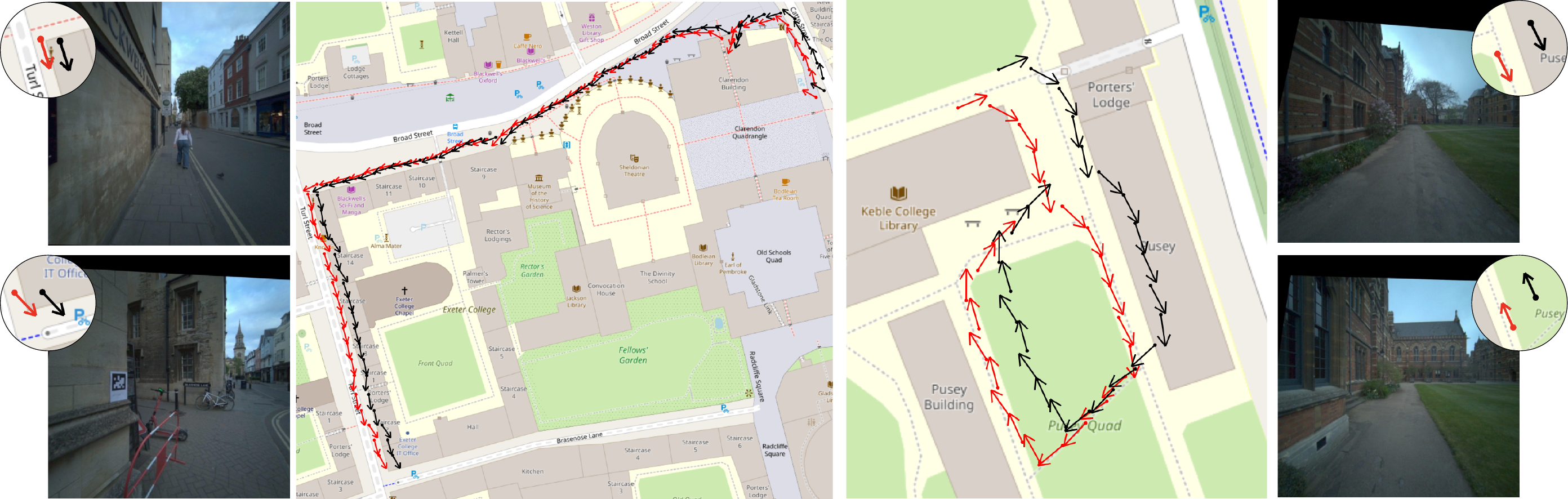}
    \end{tblr}
    \caption{\textbf{Ground-truth alignment in Oxford Day \& Night.} We perform a per-sequence, per-model alignment between the ground truth and the model estimates. This alignment is computed robustly (via \cref{eq:supp_tls,eq:supp_tlad}) to account for potential outliers in the model estimates. As shown in these examples, the resulting alignments yield more accurate reference poses with respect to which error metrics are computed.}
    \label{fig:supp_oxford_align}
\end{figure}

\section{Additional evaluation details}\label{sec:supp_eval_details}

\PAR{General}
During evaluation, OrienterNet \cite{sarlin2023orienternet}, OSMLoc \cite{liao2024osmloc}, and \name use $N=256$ discrete rotations. To match the image resolution and perspective distortion seen during training, we resize images so that their focal length is 256 pixels. We also crop LaMAria \cite{Krishnan2025lamaria} and Oxford Day-and-Night (ODN) \cite{wang2025oxford} images to $512\!\times\!512$ pixels, or to $518\!\times\!518$ pixels when using DINOv2.

\PAR{KITTI}
KITTI images have an aspect ratio of $\sim3.3{:}1$ (W:H), which is out of distribution with respect to the $1{:}1$ aspect ratio seen during training. For this reason, and for DINOv2-based models, after resizing images so that their focal length is 256 pixels, we center-pad them with zeros to the image size seen during training ($518\!\times\!518$ pixels). ResNet-based models benefit less from this padding, so for them we only resize images so that the focal length is 256 pixels. We believe that the stronger sensitivity of DINOv2-based models to aspect-ratio changes is due to the learned absolute positional encodings in DINOv2 \cite{oquab2024dinov2}, which can lead to worse generalization across aspect ratios not seen during training.

\PAR{LaMAria}
We use the eleven LaMAria training sequences for which pseudo-dense GT is available and more than one control point is observed for geo-referencing, \ie with \#CPs$>1$. We use this pseudo-dense GT for evaluation. As mentioned in the main paper (\cref{subsec:exp_ar}), we manually filter these sequences to exclude clips corresponding to the capture of ground control points. To do so, we identify the first and last frames of each surveying clip and exclude all images in between. \cref{fig:supp_lama_filt} shows an example of this filtering for one full sequence.

LaMAria's pseudo-dense GT poses are expressed with respect to the LV95 projected coordinate system, which uses grid north, whereas the ENU topocentric coordinate system we use, following OrienterNet \cite{sarlin2023orienternet}, uses geodetic north. To account for this difference, we offset the heading angles by the meridian convergence at the center of our topocentric system, which is approximately $0.8\degree$.

\PAR{Oxford Day-and-Night} We use the outdoor segments of the daytime sequences from four scenes\footnote{We do not use the \emph{oxford-robotics-institute} scene because its images are solely indoors.}: \emph{bodleian-library}, \emph{hb-allen-centre}, \emph{keble-college}, and \emph{observatory-quarter}.  As mentioned in the main paper (\cref{subsec:exp_ar}), before computing the error metrics, we align the ground-truth poses to the pose estimates of each method. This is motivated by the high local accuracy of the dataset, but the meter-level errors present in the geo-referenced poses (examples in \cref{fig:supp_oxford_align}). This procedure is akin to the standard evaluation protocol used in SLAM benchmarks \cite{fontan2025vslamlab,sturm2012benchmark,zuniga2020umavi,salas2015trajectory}, which also require the alignment between ground-truth and predictions. The main difference is that, in our case, the alignment must be robust, since estimates are obtained from single-view observations and thus can contain outliers/wrong estimates that contaminate the alignment \cite{lee2025wrong,Lee2025scores}.

Thus, we robustly compute, independently for each model and sequence of length $L$, a 3-DoF transformation $(\mathbf{R}\in\mathrm{SO}(2), \mathbf{t}\in\mathbb{R}^2)$ to align the positions and a scalar offset $\delta\in[0,2\pi)$ to align the heading angles. The 3-DoF transformation is obtained by solving the following truncated least-squares problem:
\begin{equation}\label{eq:supp_tls}
\min_{\mathbf{R}\in \mathrm{SO}(2),\; \mathbf{t}\in \mathbb{R}^2}
\sum_{i=1}^{L}
\min\left(
\left\lVert \mathbf{b}_i - \mathbf{R}\mathbf{a}_i - \mathbf{t} \right\rVert^2,\;
\bar{c}^{\,2}
\right)~,
\end{equation}
where $\mathbf{a}_i,\,\mathbf{b}_i\in\mathbb{R}^2$ denote the ground-truth and estimated positions, respectively. We set the outlier threshold to $\bar{c}=2$\,m and solve this problem using TEASER++ \cite{Yang2020tro-teaser}. Similarly, the offset for the headings is obtained by solving the following truncated least-absolute-deviations problem:
\begin{equation}\label{eq:supp_tlad}
\min_{\delta \in \mathbb{R}}
\sum_{i=1}^{L}
\min\left(
\left|\mathrm{wrap}\!\left(\beta_i-(\alpha_i+\delta)\right)\right|,\;
\bar{c}
\right).
\end{equation}
where $\alpha_i,\beta_i$ denote the ground-truth and estimated heading angles, respectively, and $\mathrm{wrap}(\cdot)$ computes the angular error by mapping differences to $(-\pi,\pi]$. We solve this problem using an implementation of the technique presented in TEAR \cite{Huang2024tear}, with the outlier threshold set to $\bar{c}=3\degree$. 

\section{Additional quantitative results}\label{sec:supp_quantitative}

\begin{table}[t]
    \centering
    \resizebox{1\linewidth}{!}{%
    \begin{tblr}{
        colspec = {*{4}l},
        row{1} = {font=\bfseries},
    }
        \toprule
        Alias & Loss function & Equations & Details \\
        \midrule
        
        Chunk$_{r=a}$
        & $\mathcal{L}_{\mathrm{GPS},~\mathrm{chunk}}$
        & Eq. \ref{eq:loss_xy_chunk}
        & chunk half-size $r=\pm a$\,m \\
        
        Chunk$_{r=a}$ $+$ $\Delta\Theta$
        & $\mathcal{L}_{\mathrm{GPS},~\mathrm{chunk}} + 0.5\mathcal{L}_{\Delta\Theta}$
        & Eqs. \ref{eq:loss_xy_chunk}, \ref{eq:loss_relrot}
        & chunk half-size $r=\pm a$\,m \\
        
        $\Delta\Theta+\normt_{\rdt=b}$
        & $0.1\mathcal{L}_{\Delta\Theta} + \mathcal{L}_{\normt}$
        & Eqs. \ref{eq:loss_relrot}, \ref{eq:loss_chunked_norm}
        & binning half-width $\rdt=\pm b$\,m \\
        
        $\Delta\Theta+\Delta XY$
        & $0.1\mathcal{L}_{\Delta\Theta} + \mathcal{L}_{\Delta XY}$
        & Eqs. \ref{eq:loss_relrot}, \ref{eq:loss_reltrans}
        & - \\
        
        Chunk$_{r=a}$ $+$ $\Delta\Theta+\Delta XY$
        & $\mathcal{L}_{\mathrm{GPS},~\mathrm{chunk}} + 0.5\mathcal{L}_{\Delta\Theta} + \mathcal{L}_{\Delta XY}$
        & Eqs. \ref{eq:loss_xy_chunk}, \ref{eq:loss_relrot}, \ref{eq:loss_reltrans}
        & chunk half-size $r=\pm a$\,m \\
        \bottomrule
    \end{tblr}
    }
    \caption{\textbf{\name aliases used in the supplementary.} 
    These checkpoints are repesentative of a practical setup where only GPS labels are available (Chunk$_{r=a}$) and when different types of relative-pose labels are also available:
    non-metric (Chunk$_{r=a}$ $+$ $\Delta\Theta$), metric ($\Delta\Theta+\normt_{\rdt=b}$) and coarsely geo-referenced ($\Delta\Theta+\Delta XY$).}
    \label{tab:supp_ckpt_summary}
\end{table}

\begin{table}[t]
\newcommand{\gnss}{GPS}
\newcommand{\onet}{OrienterNet \cite{sarlin2023orienternet}}
\newcommand{\auto}{\textbf{\name}}

\definecolor{best}{RGB}{178,255,178}
\definecolor{secd}{RGB}{255,235,158}

\centering
\resizebox{1\linewidth}{!}{%
\begin{tblr}{
    rowsep=2pt,
    colspec={l*{18}{c}},
    cell{1,2}{2,5,8,11,14,17}={c=3}{c},
    vline{2,5,8,11,14,17}={3-Z}{gray8},
    vline{2,5,8,11,14,17}={3}{abovepos=-2},
    vline{2,5,8,11,14,17}={Z}{belowpos=-2},
cell{8}{2} = {font=\bfseries},
cell{7}{2} = {cmd=\underline},
cell{8}{3} = {font=\bfseries},
cell{7}{3} = {cmd=\underline},
cell{7}{4} = {font=\bfseries},
cell{5}{4} = {cmd=\underline},
cell{8}{5} = {font=\bfseries},
cell{7}{5} = {cmd=\underline},
cell{8}{6} = {font=\bfseries},
cell{7}{6} = {cmd=\underline},
cell{8}{7} = {font=\bfseries},
cell{7}{7} = {cmd=\underline},
cell{8}{8} = {font=\bfseries},
cell{7}{8} = {cmd=\underline},
cell{8}{9} = {font=\bfseries},
cell{7}{9} = {cmd=\underline},
cell{8}{10} = {font=\bfseries},
cell{7}{10} = {cmd=\underline},
cell{8}{11} = {font=\bfseries},
cell{7}{11} = {cmd=\underline},
cell{5}{12} = {font=\bfseries},
cell{7}{12} = {cmd=\underline},
cell{5}{13} = {font=\bfseries},
cell{7}{13} = {font=\bfseries},
cell{8}{13} = {cmd=\underline},
cell{8}{14} = {font=\bfseries},
cell{7}{14} = {cmd=\underline},
cell{8}{15} = {font=\bfseries},
cell{7}{15} = {cmd=\underline},
cell{8}{16} = {font=\bfseries},
cell{5}{16} = {cmd=\underline},
cell{8}{17} = {font=\bfseries},
cell{7}{17} = {cmd=\underline},
cell{8}{18} = {font=\bfseries},
cell{7}{18} = {cmd=\underline},
cell{5}{19} = {font=\bfseries},
cell{8}{19} = {font=\bfseries},
cell{7}{19} = {cmd=\underline},
}
\toprule
Method  
& Lateral [m] & - & - 
& Longitudinal [m] & - & - 
& Orientation [\degree] & - & - 
& Lat. (seq) [m] & - & - 
& Long. (seq) [m] & - & - 
& Orient. (seq) [\degree] & - & - 
\\
\cmidrule[lr]{2-4} 
\cmidrule[lr]{5-7} 
\cmidrule[lr]{8-10} 
\cmidrule[lr]{11-13}
\cmidrule[lr]{14-16}
\cmidrule[lr]{17-19}
& R@1/3/5\,\textuparrow & - & - 
& R@1/3/5\,\textuparrow & - & - 
& R@1/3/5\,\textuparrow & - & - 
& R@1/3/5\,\textuparrow & - & - 
& R@1/3/5\,\textuparrow & - & - 
& R@1/3/5\,\textuparrow & - 
\\
\midrule
\onet &   37.7 &         67.3 &         77.4 &         17.6 &         42.6 &         53.7 &       15.7 &       42.1 &       57.1 & 75.2 &         95.3 &         98.5 &         47.9 &         92.9 &         95.1 &       70.8 &       93.8 &       98.7 \\
\auto \\
$\hookrightarrow$ Chunk$_{r=20}$ &  46.8 &         77.9 &         84.9 &         21.6 &         49.8 &         61.4 &       20.9 &       53.6 &       68.6 & 79.6 &         96.3 &         98.7 &         52.5 &         91.5 &         92.0 &       82.7 &       94.6 &      100.0 \\
$\hookrightarrow$ Chunk$_{r=20} + \Delta\Theta$ &  47.8 &         79.0 &         85.6 &         21.0 &         47.5 &         58.3 &       26.3 &       60.1 &       71.1  &  81.1 &         95.8 &         98.6 &         46.4 &         84.8 &         89.4 &       80.5 &       95.6 &      100.0 \\
$\hookrightarrow$ $\Delta\Theta+\normt_{\rdt=5}$ &       55.3 &         82.9 &         87.6 &         29.3 &         56.0 &         63.4 &       25.5 &       62.9 &       75.8  & 84.6 &         97.2 &         98.7 &         60.6 &         91.5 &         92.0 &       84.3 &       96.9 &       98.7 \\
$\hookrightarrow$  $\Delta\Theta+\Delta XY$ &       56.6 &         83.2 &         87.8 &         33.2 &         59.5 &         66.0 &       28.9 &       64.4 &       75.9 &  86.4 &         96.5 &         98.4 &         76.8 &         92.9 &         96.0 &       90.7 &       96.9 &       99.6 \\
\bottomrule
\end{tblr}
}
\caption{\textbf{Results per supervision type on KITTI \cite{geiger2012kitti} without orientation priors}. All methods use a ResNet-101 image backbone and are trained on the currently available sequences of MGL \cite{sarlin2023orienternet} (11 cities). The notation used and motivation for \name checkpoints is explained in \cref{tab:supp_ckpt_summary}. 
}
\label{tab:supp_kitti}
\end{table}

\begin{table}[t]
\newcommand{\gnss}{GPS}
\newcommand{\onet}{OrienterNet \cite{sarlin2023orienternet}}
\newcommand{\auto}{\textbf{\name}}

\definecolor{best}{RGB}{178,255,178}
\definecolor{secd}{RGB}{255,235,158}

\centering
\resizebox{0.9\linewidth}{!}{%
\begin{tblr}{
    rowsep=0.5pt,
    colspec={ll*{12}{c}},
    cell{1,2}{3,6,9,12}={c=3}{c},
    cell{3,9}{1}={r=6}{c, m, cmd=\rotatebox{90}}, 
    vline{3,6,9,12}={3-8,9-Z}{gray8},
    vline{3,6,9,12}={3,9}{abovepos=-2},
    vline{3,6,9,12}={8,Z}{belowpos=-2},
cell{8}{3} = {font=\bfseries},
cell{7}{3} = {cmd=\underline},
cell{8}{4} = {font=\bfseries},
cell{7}{4} = {cmd=\underline},
cell{8}{5} = {font=\bfseries},
cell{7}{5} = {cmd=\underline},
cell{8}{6} = {font=\bfseries},
cell{7}{6} = {cmd=\underline},
cell{8}{7} = {font=\bfseries},
cell{6}{7} = {cmd=\underline},
cell{8}{8} = {font=\bfseries},
cell{6}{8} = {cmd=\underline},
cell{8}{9} = {font=\bfseries},
cell{7}{9} = {cmd=\underline},
cell{8}{10} = {font=\bfseries},
cell{6}{10} = {cmd=\underline},
cell{8}{11} = {font=\bfseries},
cell{6}{11} = {cmd=\underline},
cell{8}{12} = {font=\bfseries},
cell{7}{12} = {cmd=\underline},
cell{8}{13} = {font=\bfseries},
cell{6}{13} = {cmd=\underline},
cell{8}{14} = {font=\bfseries},
cell{7}{14} = {cmd=\underline},
cell{14}{3} = {font=\bfseries},
cell{13}{3} = {cmd=\underline},
cell{14}{4} = {font=\bfseries},
cell{13}{4} = {cmd=\underline},
cell{11}{5} = {font=\bfseries},
cell{14}{5} = {cmd=\underline},
cell{13}{6} = {font=\bfseries},
cell{12}{6} = {cmd=\underline},
cell{12}{7} = {font=\bfseries},
cell{13}{7} = {cmd=\underline},
cell{12}{8} = {font=\bfseries},
cell{13}{8} = {cmd=\underline},
cell{14}{9} = {font=\bfseries},
cell{11}{9} = {cmd=\underline},
cell{14}{10} = {font=\bfseries},
cell{11}{10} = {cmd=\underline},
cell{11}{11} = {font=\bfseries},
cell{10}{11} = {cmd=\underline},
cell{14}{12} = {font=\bfseries},
cell{13}{12} = {cmd=\underline},
cell{14}{13} = {font=\bfseries},
cell{13}{13} = {cmd=\underline},
cell{14}{14} = {font=\bfseries},
cell{13}{14} = {cmd=\underline},
}
\toprule
& Method  
& XY [m] & - & - 
& Orientation [\degree] & - & - 
& XY (seq) [m] & - & - 
& Orient. (seq) [\degree] & - & - 
\\
\cmidrule[lr]{3-5} 
\cmidrule[lr]{6-8} 
\cmidrule[lr]{9-11} 
\cmidrule[lr]{12-14}
&
& R@1/3/5\,\textuparrow & - & - 
& R@1/3/5\,\textuparrow & - & - 
& R@1/3/5\,\textuparrow & - & - 
& R@1/3/5\,\textuparrow & - & - 
\\
\midrule
LaMAria \cite{Krishnan2025lamaria}
& \onet  &       1.9 &      10.9 &      18.2 &        4.8 &       13.4 &       20.7 &      35.8 &      76.4 &      84.6 &       50.8 &       83.8 &       89.0 \\
& \auto \\
& $\hookrightarrow$ Chunk$_{r=20}$ &     3.7 &      16.5 &      26.3 &        7.3 &       20.0 &       28.7   &   47.9 &      85.6 &      86.4 &       61.8 &       89.0 &       91.7 \\
& $\hookrightarrow$ Chunk$_{r=20}$ $+$ $\Delta\Theta$ &   2.9 &      17.6 &      27.3 &        8.7 &       22.5 &       31.9    &    41.4 &      85.1 &      89.9 &       58.4 &       87.0 &       92.0 \\
& $\hookrightarrow$ $\Delta\Theta+\normt_{\rdt=5}$ &       5.3 &      20.5 &      28.6 &        9.5 &       23.4 &       31.9 &      60.2 &      86.8 &      90.0 &       72.6 &       90.5 &       93.5 \\
& $\hookrightarrow$  $\Delta\Theta+\Delta XY$ &       8.1 & 25.2 & 33.0 & 10.5 & 26.4 & 35.5 & 72.6 & 92.1 & 92.6 & 77.8 & 92.8 & 94.0\\
\midrule[gray8]
\shortstack{Oxford \cite{wang2025oxford} \\ Day \& Night}
& \onet &       9.8 &      36.3 &      50.5 &       14.6 &       35.1 &       48.5 &      78.6 &      96.2 &      99.6 &       73.7 &       97.0 &       99.4 \\
& \auto \\
& $\hookrightarrow$ Chunk$_{r=20}$ &   10.4 &      36.8 &      52.2 &       17.8 &       41.6 &       54.4     &     76.6 &      94.8 &      98.4 &       69.5 &       93.0 &       98.4 \\
& $\hookrightarrow$ Chunk$_{r=20}$ $+$ $\Delta\Theta$ &   9.3 &      35.5 &      51.8 &       19.8 &       43.5 &       55.7  & 74.9 &      95.6 &      99.3 &       65.7 &       93.6 &       99.0 \\
& $\hookrightarrow$ $\Delta\Theta+\normt_{\rdt=5}$ &      11.5 &      41.4 &      55.9 &       20.5 &       44.9 &       57.3 &      81.0 &      96.1 &      98.7 &       76.0 &       99.2 &       99.7 \\
& $\hookrightarrow$  $\Delta\Theta+\Delta XY$ &      13.2 & 42.9 & 56.4 & 19.0 & 43.8 & 56.5 & 86.8 & 99.4 & 99.4 & 80.6 & 99.7 & 99.9 \\
\bottomrule
\end{tblr}
}
\caption{\textbf{Results per supervision type on LaMAria \cite{Krishnan2025lamaria} and Oxford Day \& Night \cite{wang2025oxford}}. All methods use a ResNet-101 image backbone and are trained on the currently available sequences of MGL \cite{sarlin2023orienternet} (11 cities). The notation used and motivation for \name checkpoints is explained in \cref{tab:supp_ckpt_summary}.
}
\label{tab:supp_lama_oxford}
\end{table}

\PAR{Supervision impact}
We expand the results over supervision types presented in \cref{subsec:exp_ablation} and in the right subfigure of \cref{fig:ablation} by providing separate results in KITTI (\cref{tab:supp_kitti}) and LaMAria and Oxford Day-and-Night (ODN) (\cref{tab:supp_lama_oxford}). 
The notation and details used for our checkpoints are explained in \cref{tab:supp_ckpt_summary}.
We also report sequential results, using a sequence length of 100 frames in KITTI and 50 frames in LaMAria and ODN. For reference, we also add metrics for the corresponding retrained version of OrienterNet with strong supervision \cite{sarlin2023orienternet}.

Conclusions remain consistent with the average results: performance correlates with the informativeness of ground-truth labels, specially on KITTI and LaMAria (performance is similar across checkpoints on ODN, presumably due to slight canceling of errors when performing the alignment). The weak chunk marginalization loss is more appropriate than the standard strong supervision, and it is further outperformed when relative pose labels are available.

\begin{table}[t!h]

\newcommand{\slic}{SliceMatch \cite{lentsch2023slicematch}}
\newcommand{\ccvp}{CCVPE \cite{xia2024ccvpe}}
\newcommand{\loct}{Loc$^2$ \cite{xia2026loc}}

\newcommand{\refi}{Shi \etal \cite{shi2022beyond}}
\newcommand{\boos}{Shi \etal \cite{shi2023boosting}}
\newcommand{\weak}{Shi \etal \cite{shi2024weakly}}
\newcommand{\fgtw}{FG$^2$ \cite{xia2025fg}}

\newcommand{\onet}{OrienterNet \cite{sarlin2023orienternet}}
\newcommand{\oloc}{OSMLoc \cite{liao2024osmloc}}

\centering
\resizebox{\textwidth}{!}{%
\begin{tblr}{
    colspec={lcll*{9}{c}},
    colsep=8pt,
    stretch=0pt,
    cell{1}{1-4}={r=2}{l},
    cell{1,2}{5,8,11}={c=3}{c},
    cell{3}{1,4}={r=4}{l},
    cell{3}{2}={r=4}{c},
    cell{7}{1,2}={r=2}{l},
    cell{9,13}{1,2,4}={r=4}{l},
    column{1,2,3} = {leftsep=2pt,rightsep=2pt},
    column{6,9,12} = {leftsep=0.5pt,rightsep=0.5pt},
    cell{12}{8}={font=\bfseries},
    cell{16}{5-7,9-13}={font=\bfseries},
    cell{12}{11}={cmd=\underline},
    cell{13}{7,9-10,12}={cmd=\underline},
    cell{15}{5-6,12-13}={cmd=\underline},
    cell{16}{8}={cmd=\underline},
}
\toprule
Map & \shortstack[c]{Training \\ dataset} & Method & \shortstack[c]{Image \\ encoder} & Lateral [m] & & & Longitudinal [m] & & & Orientation [$^\circ$] & & \\
&&&& R@1/3/5 $\uparrow$ & & & R@1/3/5 $\uparrow$ & & & R@1/3/5 $\uparrow$ & & \\
\midrule
Satellite & KITTI & 
    \boos & N/A & 57.7 & 86.8 & 91.2   & 14.2 & 34.6 & 45.0    & 99.0 & 100  & 100 \\
&&  \weak &     & 64.7 & 86.2 & -      & 11.8 & 34.8 & -       & 100  & 100  & 100 \\
&&  \fgtw &     & 37.9 & -    & 85.7   & 22.0 & -    & 60.8    & 23.0 & -    & 77.8 \\
&&  \loct &     & 45.3 & -    & 92.4   & 27.0 & -    & 68.3    & 26.0 & -    & 80.7 \\
\midrule
OSM & \shortstack[c]{MGL \\ (12 cities)} &
   \onet & ResNet-101 & 50.4 & 85.2 & 92.7 & 24.4 & 56.1 & 68.0 & 29.3 & 68.2 & 84.5 \\
&& \oloc & DINOv2-B   & 60.1 & 93.0 & 96.6 & 28.7 & 61.9 & 72.6 & 30.9 & 71.9 & 89.2 \\
\midrule[gray5,dashed]
OSM & \shortstack[c]{MGL \\ (11 cities)} &
   \onet                     & ResNet-101 & 48.3 & 82.0 & 91.4 & 20.8 & 50.3 & 62.6 & 25.5 & 62.0 & 79.8 \\
&& \textbf{\name} \\
&& $\hookrightarrow$ raw GPS &            & 51.5 & 88.8 & 94.1 & 29.6 & 61.4 & 71.2 & 29.7 & 69.0 & 86.2 \\
&& $\hookrightarrow$ w/ rel. poses &      & 64.7 & 91.6 & 95.1 & 36.3 & 65.1 & 71.3 & 37.8 & 80.6 & 93.3 \\
\midrule[gray5,dashed]
OSM & \shortstack[c]{MGL \\ (11 cities)} &
   \onet                      & DINOv2-B & 53.5 & 92.8 & 97.4 & 26.2 & 65.9 & 77.5 & 36.9 & 81.1 & 93.7 \\
&& \textbf{\name} \\
&& $\hookrightarrow$ raw GPS &           & 70.9 & 95.0 & 97.1 & 31.2 & 65.3 & 75.4 & 35.3 & 81.1 & 94.2 \\
&& $\hookrightarrow$ w/ rel. poses &     & 76.2 & 95.6 & 98.2 & 35.9 & 74.0 & 81.1 & 43.5 & 86.6 & 95.8 \\
\bottomrule
\end{tblr}
}
\caption{\textbf{Results on KITTI \cite{geiger2012kitti} with a $\pm10\degree$ orientation prior}. Following \cite{sarlin2023orienternet,shi2022beyond}, we restrict the orientation search space to $\pm10\degree$ around the ground-truth orientation. The results follow the same trends as in the evaluation without orientation priors reported in \cref{tab:kitti_main}: \name improves over strongly-supervised baselines \cite{sarlin2023orienternet,liao2024osmloc} even when trained only with raw GPS labels, and supervision with relative poses yields the best performance among OSM-based methods.}
\label{tab:supp_kitti_prior}
\end{table}

\PAR{KITTI evaluation with a $\pm10\degree$ orientation prior}
\cref{tab:supp_kitti_prior} reports results under the alternative standard protocol \cite{sarlin2023orienternet,shi2022beyond}, which restricts the pose search space to $\pm20$\,m and $\pm10\degree$ around the ground-truth pose. As in \cref{subsec:exp_driving}, we restrict the position search to a $40\!\times\!40$\,m area centered on the OSM tile, whose center is randomly sampled within $\pm20$\,m of the ground-truth position.
The conclusions are very similar to those obtained without an orientation prior: \name' weak supervision with either raw GPS labels or relative poses is also beneficial under this setting.
Note that satellite-based baselines \cite{shi2023boosting,shi2024weakly,xia2025fg,xia2026loc} are trained with this strong orientation prior, which can bias networks toward predicting orientations within this range.

\begin{table}[t]
\newcommand{\gnss}{GPS}
\newcommand{\onet}{OrienterNet \cite{sarlin2023orienternet}}
\newcommand{\olos}{OSMLoc-s \cite{liao2024osmloc}}
\newcommand{\olob}{OSMLoc-b \cite{liao2024osmloc}}
\newcommand{\auto}{\textbf{\name}}

\definecolor{best}{RGB}{178,255,178}
\definecolor{secd}{RGB}{255,235,158}

\centering
\resizebox{0.75\linewidth}{!}{%
\begin{tblr}{
    rowsep=0.5pt,
    colspec={l*{6}{c}},
    cell{1}{1}={r=2}{l},
    cell{1,2}{2,5}={c=3}{c},
}
\toprule
Method 
& XY [m] & - & - 
& Orientation [\degree] & - & - 
\\
\cmidrule[lr]{2-4} 
\cmidrule[lr]{5-7} 
& R@1/3/5\,\textuparrow & - & - 
& R@1/3/5\,\textuparrow & - & - 
\\
\midrule
Varying the chunk-size ($r$) in \cref{eq:loss_xy_chunk} \\
$\hookrightarrow$ Chunk$_{r=0}$    &  5.4 &      24.3 &      37.1 &       13.0 &       32.4 &       43.3 \\
$\hookrightarrow$ Chunk$_{r=5}$    &  8.4 &      32.9 &      44.4 &       14.9 &       37.3 &       49.4 \\
$\hookrightarrow$ Chunk$_{r=10}$   &  8.1 &      33.0 &      45.7 &       15.6 &       38.7 &       51.0 \\
$\hookrightarrow$ Chunk$_{r=20}$   &  8.2 &      31.6 &      45.0 &       15.3 &       38.4 &       50.6 \\
$\hookrightarrow$ Chunk$_{r=30}$   &  6.2 &      25.7 &      37.6 &       14.1 &       35.1 &       46.2 \\
\midrule[gray8]
\midrule[gray8]
Varying the binning-size ($\rdt$) in \cref{eq:loss_chunked_norm} \\
$\hookrightarrow$ $\Delta\Theta+\normt_{\rdt=1}$   &   10.5 &      35.7 &      46.5 &       17.5 &       41.6 &       52.7 \\
$\hookrightarrow$ $\Delta\Theta+\normt_{\rdt=5}$   &  11.3 &      37.5 &      48.2 &       18.5 &       43.7 &       55.0 \\
$\hookrightarrow$ $\Delta\Theta+\normt_{\rdt=10}$  &   9.5 &      34.1 &      44.9 &       17.8 &       41.0 &       52.1 \\
\midrule[gray8]
\midrule[gray8]
GPS + relative pose supervision  \\
$\hookrightarrow$ Chunk$_{r=20}$                          &       8.2 &      31.6 &      45.0 &       15.3 &       38.4 &       50.6 \\
$\hookrightarrow$ Chunk$_{r=5} + \Delta\Theta+\Delta XY$ &      11.0 &      37.5 &      48.1 &       19.1 &       43.6 &       54.2 \\
$\hookrightarrow$ Chunk$_{r=20} + \Delta\Theta+\Delta XY$ &      11.1 &      36.9 &      47.5 &       18.7 &       43.9 &       54.7 \\
$\hookrightarrow$ $\Delta\Theta+\Delta XY$                &      14.4 &      40.7 &      50.8 &       19.4 &       44.9 &       56.0 \\
\bottomrule
\end{tblr}
}
\caption{\textbf{Additional experiments.} We show average recalls across benchmarks for different setups: Varying the GPS neighborhood size $r$ in \cref{eq:loss_xy_chunk}, the norm binning size $\rdt$ in \cref{eq:loss_chunked_norm}, and combining GPS and relative pose supervision. Increasing $r$ up to 5\,m improves performance, while $\rdt$ has little impact. Finally, relative pose supervision remains more effective than GPS supervision alone.
}
\label{tab:supp_extra_ablations}
\end{table}

\PAR{Varying the chunk size $r$ for GPS-only supervision}
\cref{tab:supp_extra_ablations} shows the performance obtained by varying the marginalization chunk size $r$ used for GPS supervision with \cref{eq:loss_xy_chunk}. Directly using GPS ($r\!=\!0$), without marginalization over a local neighborhood, yields worse localization performance. 
Performance is similar for $r\!\in\![5,20]$\,m and thus consistently outperforms strong supervision. Performance decreases only for larger chunks ($r\!=\!30$\,m), arguably due to a weaker training signal.

\PAR{Varying the norm binning size $\rdt$}
\cref{tab:supp_extra_ablations} also shows that the binning size $\rdt$ used for relative distance supervision (\cref{eq:loss_chunked_norm}) is not a sensitive parameter, as similar performance is obtained across different values of $\rdt$.

\PAR{GPS + relative pose supervision}
Finally, \cref{tab:supp_extra_ablations} also tests the combination of GPS supervision (\cref{eq:loss_xy_chunk}) with relative pose supervision (\cref{eq:loss_relrot,eq:loss_reltrans}). Adding the latter improves performance over using GPS supervision alone. Interestingly, and contrary to what happens when using only GPS, increasing the size $r$ of the supervised GPS neighborhood yields better performance, achieving a closer, but still worse, accuracy to that obtained when just using relative pose supervision. This suggests that relative pose labels, when available and on their own, are already suitable for training visual localization models.

\begin{figure}[t]
    \def\plotH{1.15cm}
    \centering
    \begin{tblr}{
        colspec={*{6}c},
        colsep=1pt, 
        rowsep=1pt, 
        row{1,2,4}={belowsep=0.5pt},
        row{2,3,5}={abovesep=0.5pt},
        row{Y}={belowsep=0pt},
        row{Z}={abovesep=0pt},
        stretch=0pt,
        cell{2-Z}{1}={}{halign=c, valign=m,cmd={\scriptsize\rotatebox[origin=c]{90}}},
    }
    & Images
    & Input Map & Neural Map & Features Norm & Probabilities \\
    \name
    & \includegraphics[valign=m,height=\plotH]{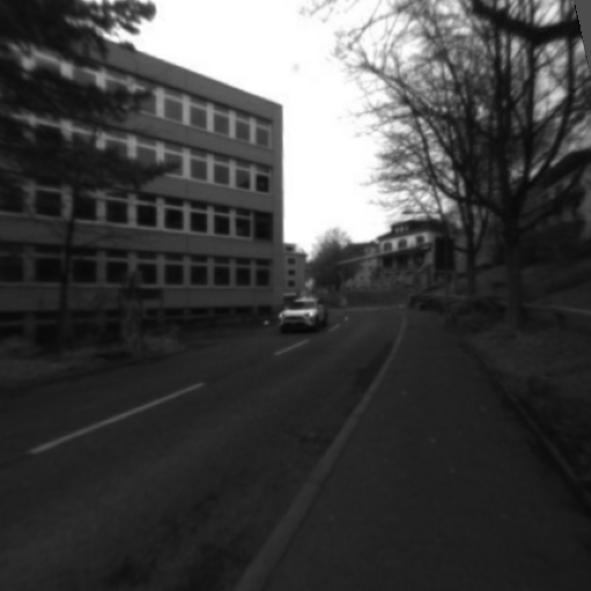} 
    & \includegraphics[valign=m,height=\plotH]{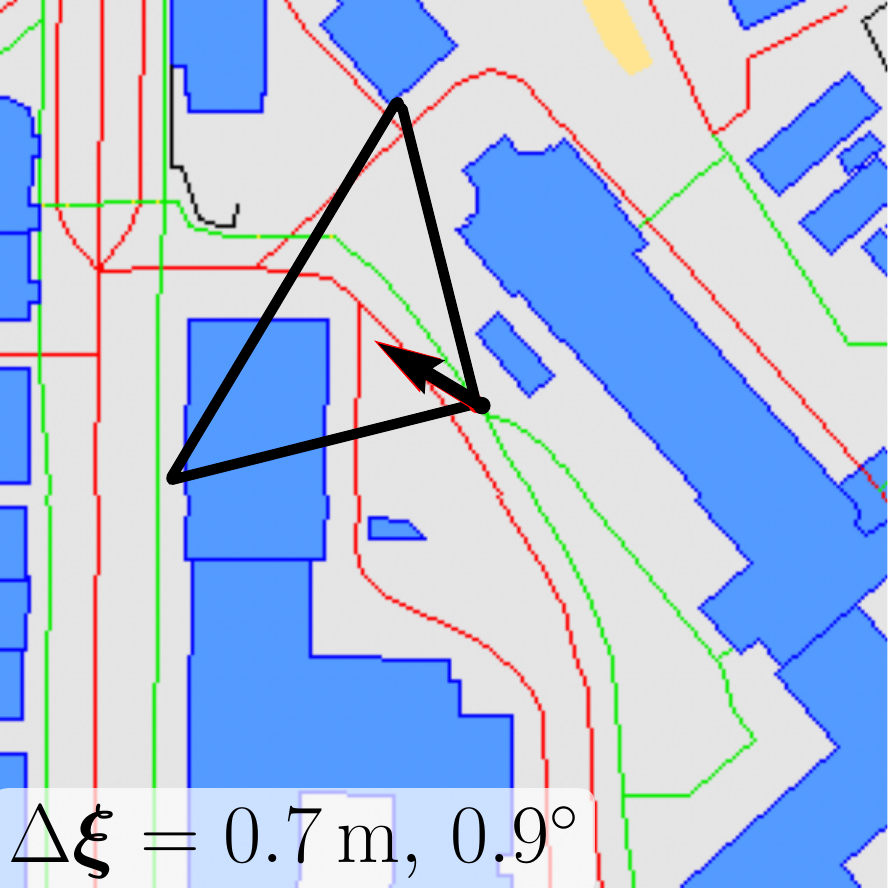} 
    & \includegraphics[valign=m,height=\plotH]{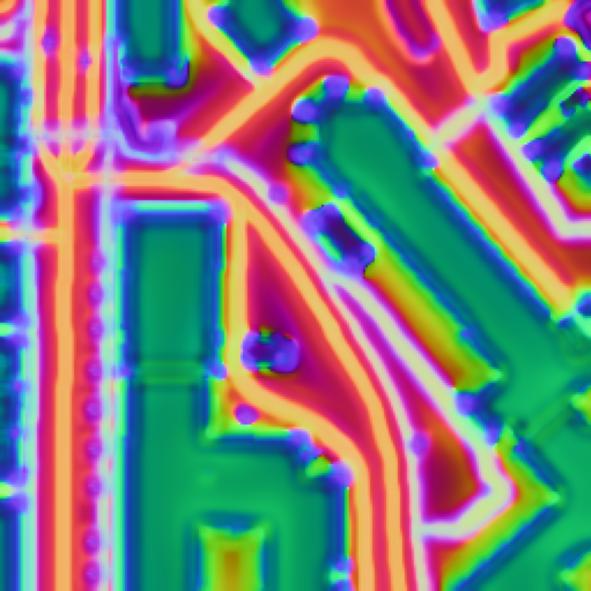} 
    & \includegraphics[valign=m,height=\plotH]{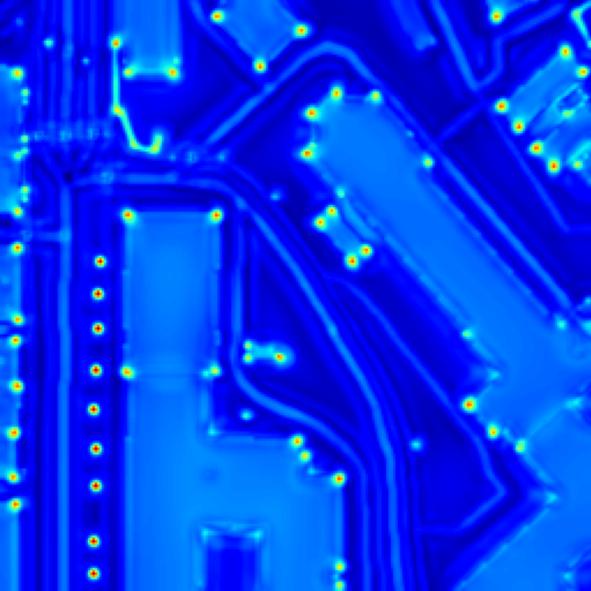} 
    & \includegraphics[valign=m,height=\plotH]{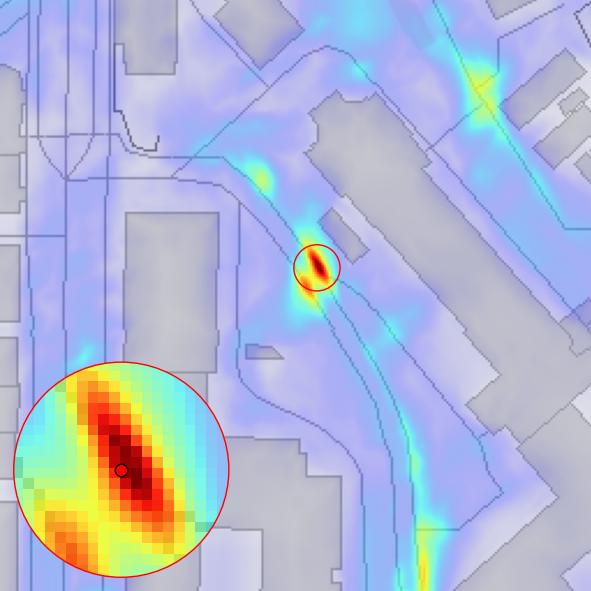} \\
    OrienterNet \cite{sarlin2023orienternet}
    & \includegraphics[valign=m,height=\plotH]{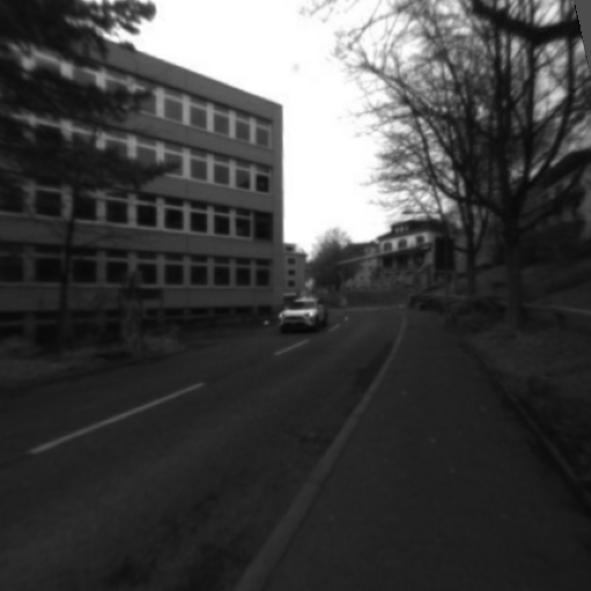} 
    & \includegraphics[valign=m,height=\plotH]{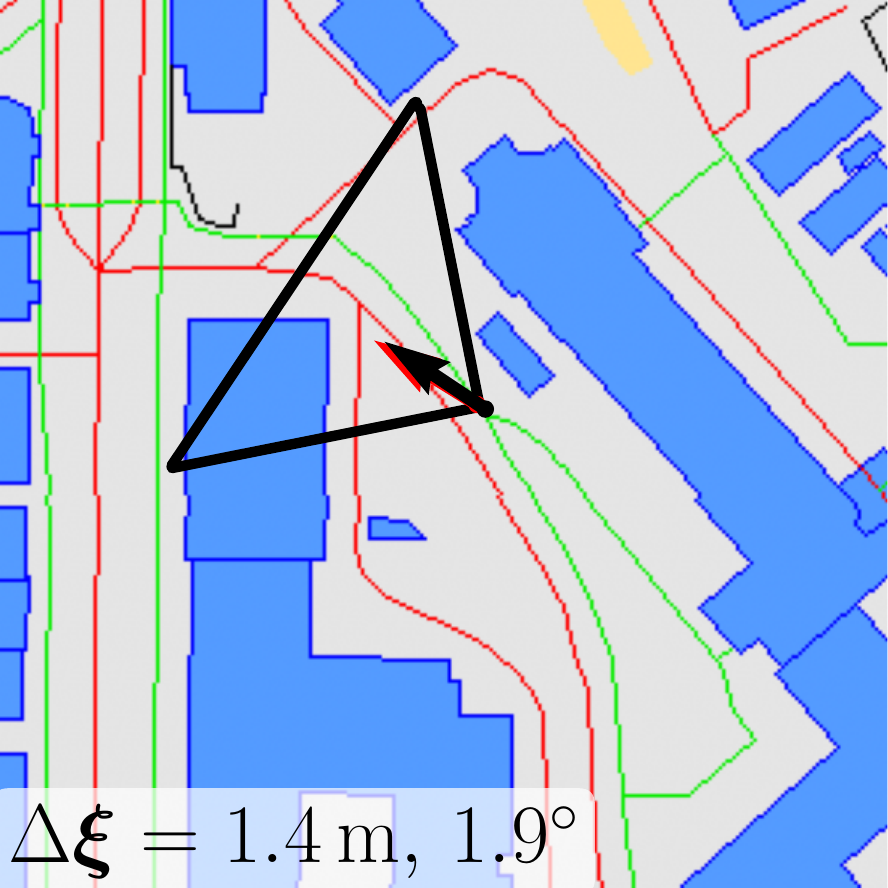} 
    & \includegraphics[valign=m,height=\plotH]{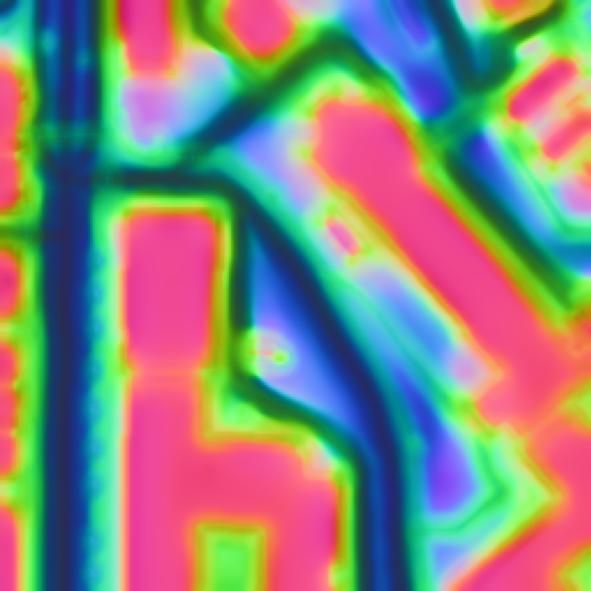} 
    & \includegraphics[valign=m,height=\plotH]{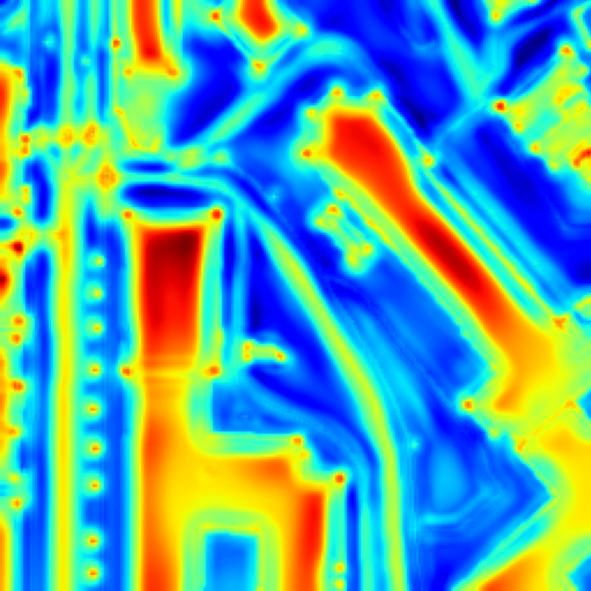} 
    & \includegraphics[valign=m,height=\plotH]{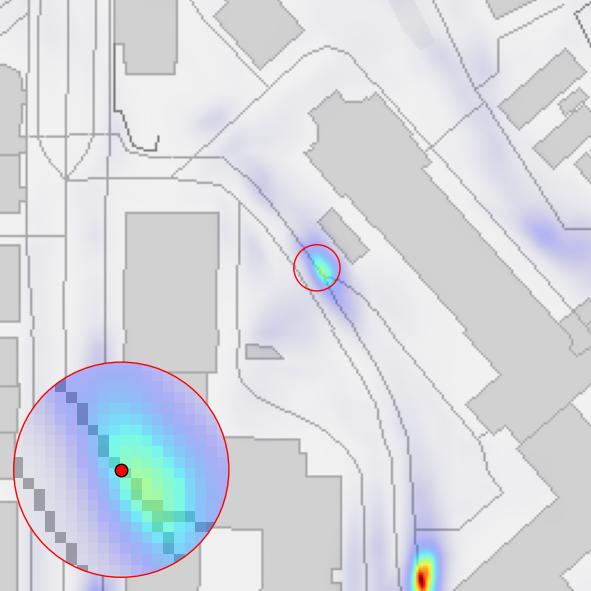} \\ 
    \name
    & \includegraphics[valign=m,height=\plotH]{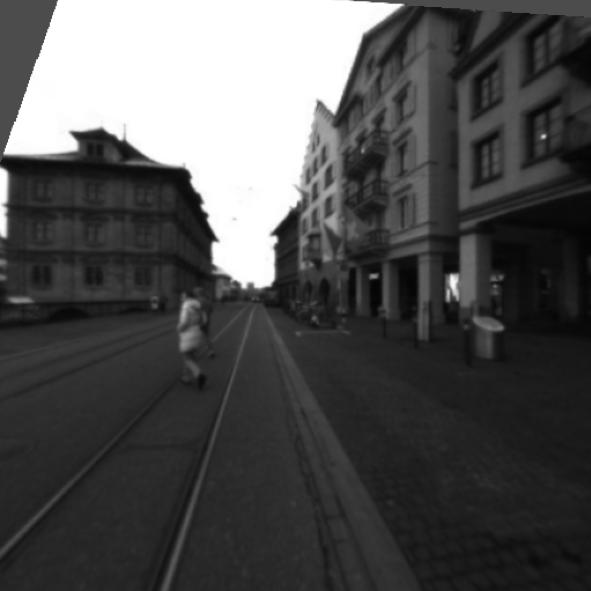} 
    & \includegraphics[valign=m,height=\plotH]{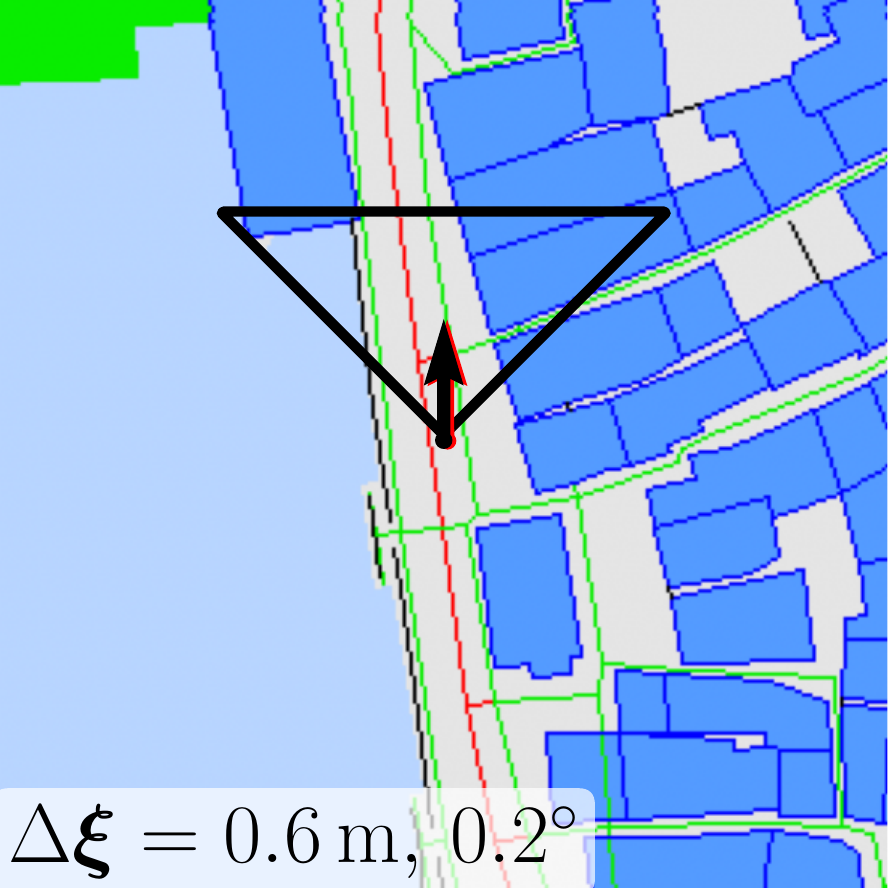} 
    & \includegraphics[valign=m,height=\plotH]{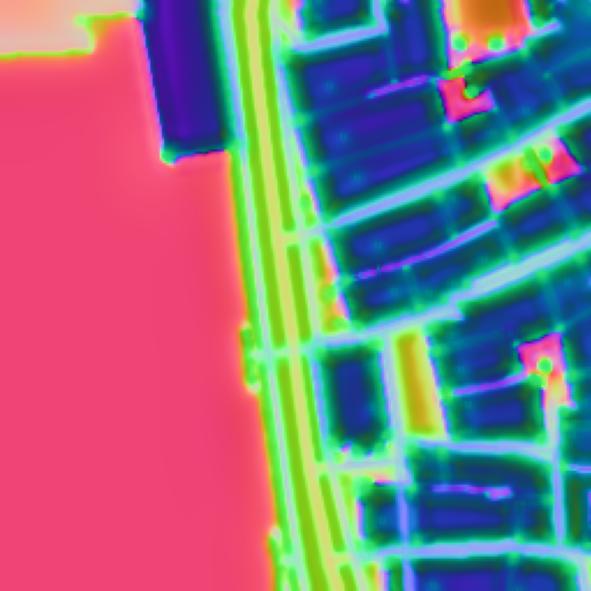} 
    & \includegraphics[valign=m,height=\plotH]{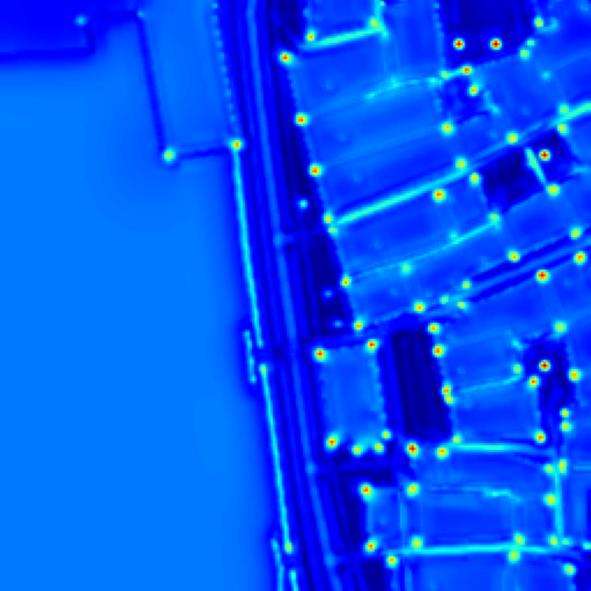} 
    & \includegraphics[valign=m,height=\plotH]{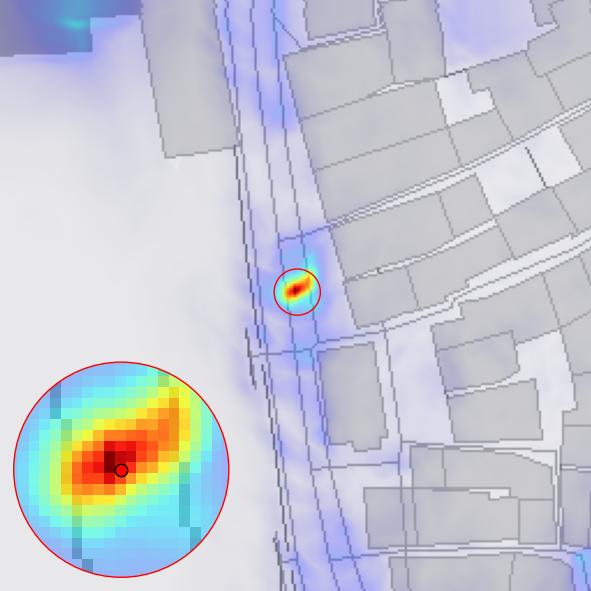} \\ 
    OSMLoc \cite{liao2024osmloc}
    & \includegraphics[valign=m,height=\plotH]{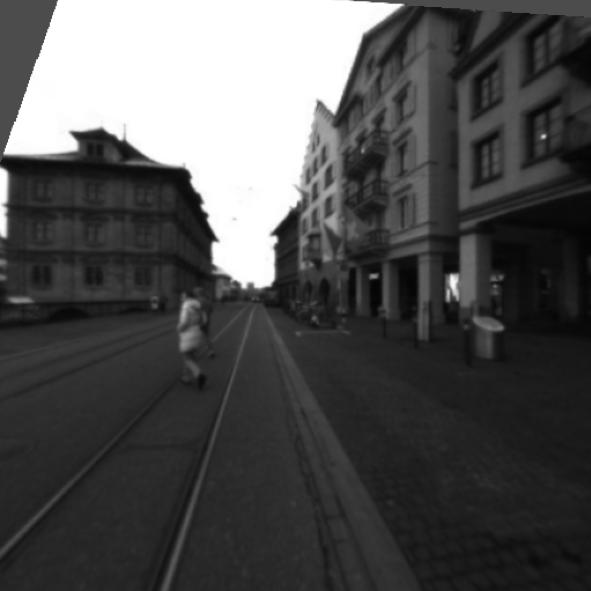} 
    & \includegraphics[valign=m,height=\plotH]{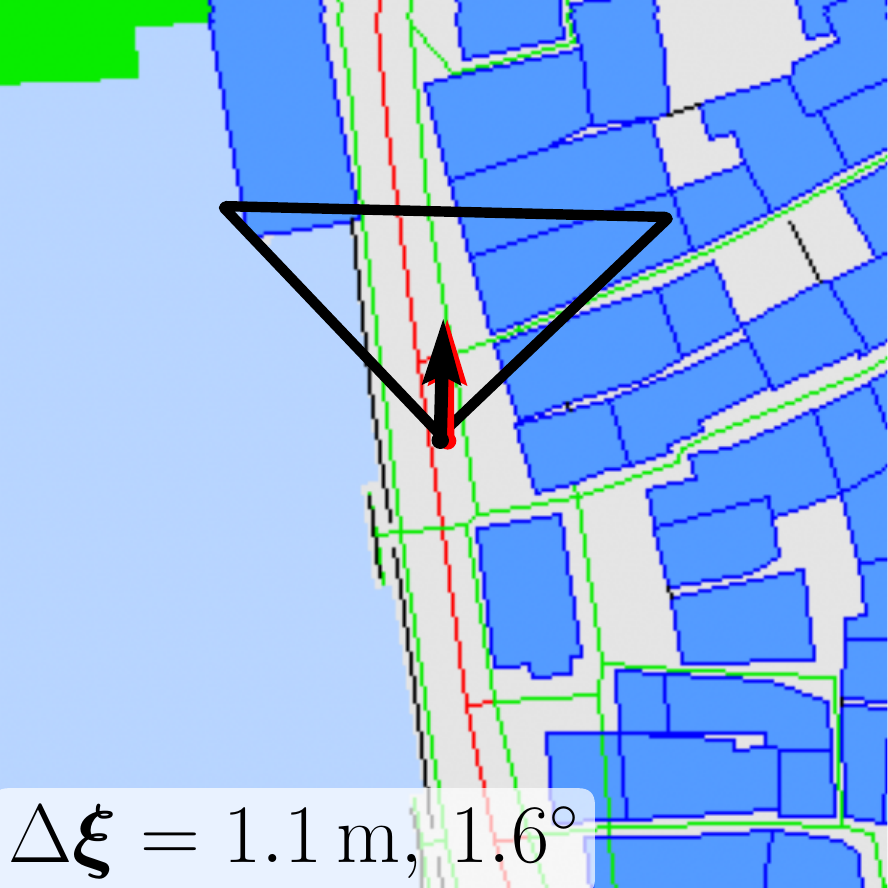} 
    & \includegraphics[valign=m,height=\plotH]{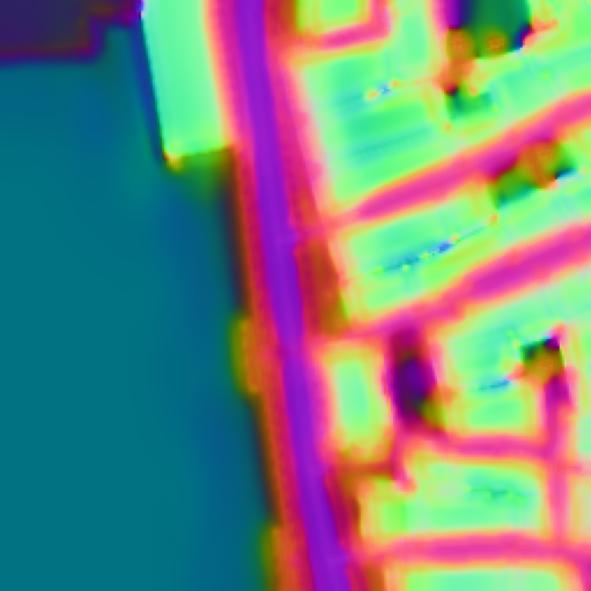} 
    & \includegraphics[valign=m,height=\plotH]{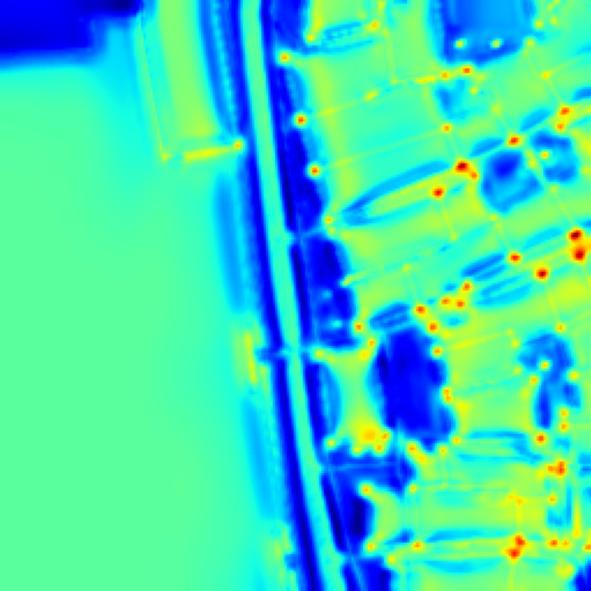} 
    & \includegraphics[valign=m,height=\plotH]{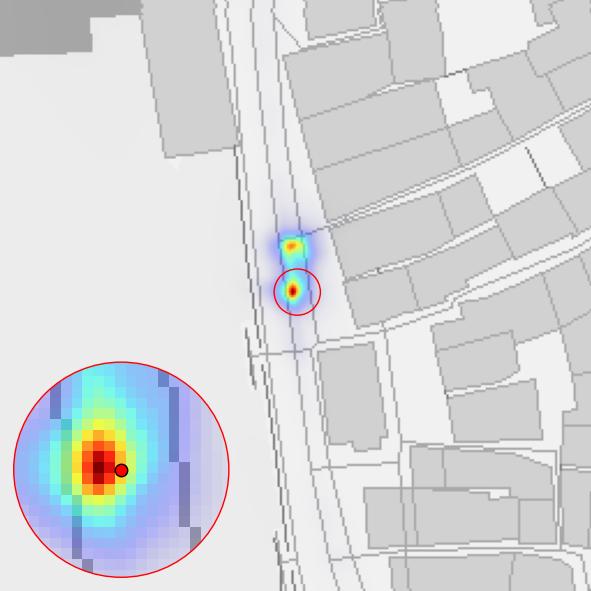} \\ 
    \name
    & \includegraphics[valign=m,height=\plotH]{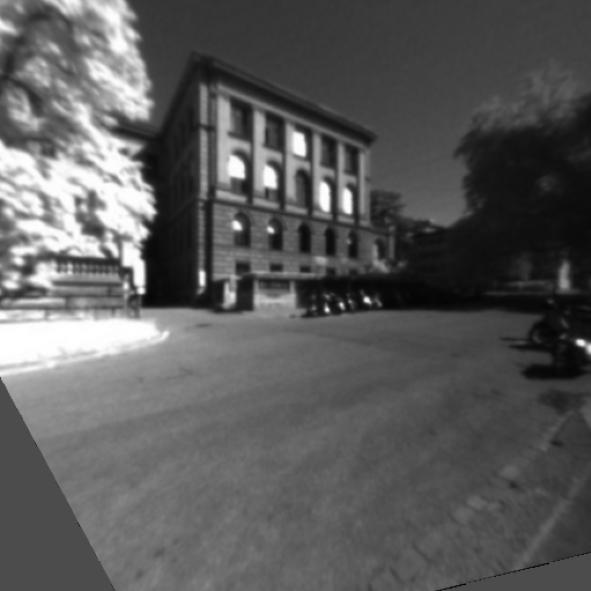} 
    & \includegraphics[valign=m,height=\plotH]{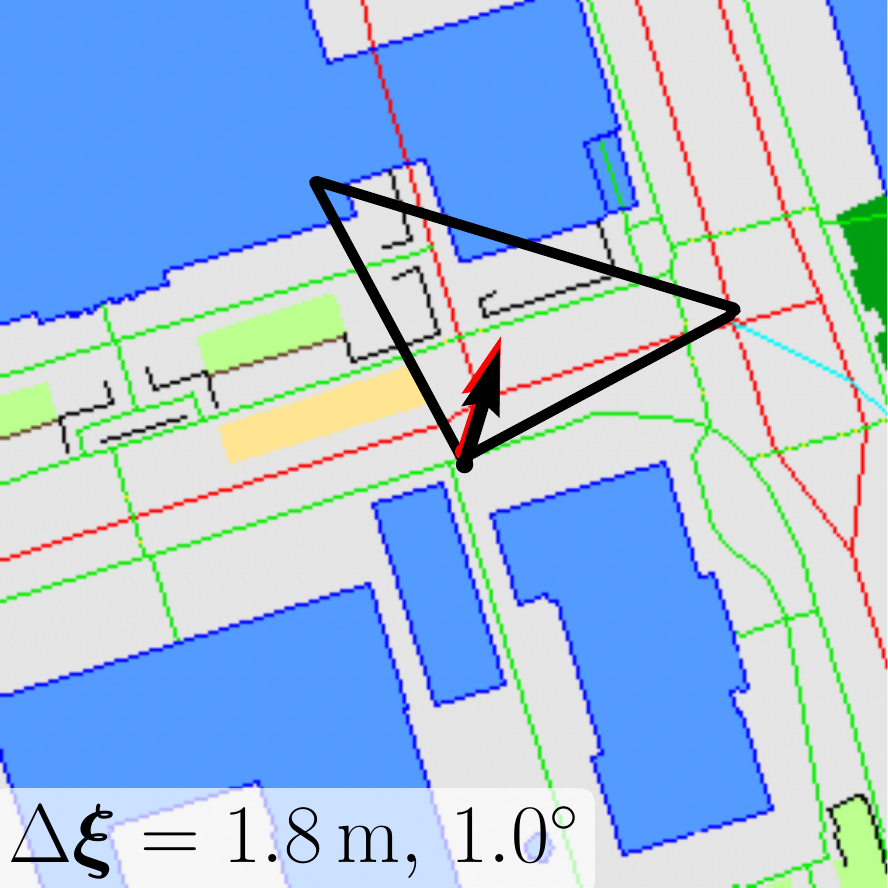} 
    & \includegraphics[valign=m,height=\plotH]{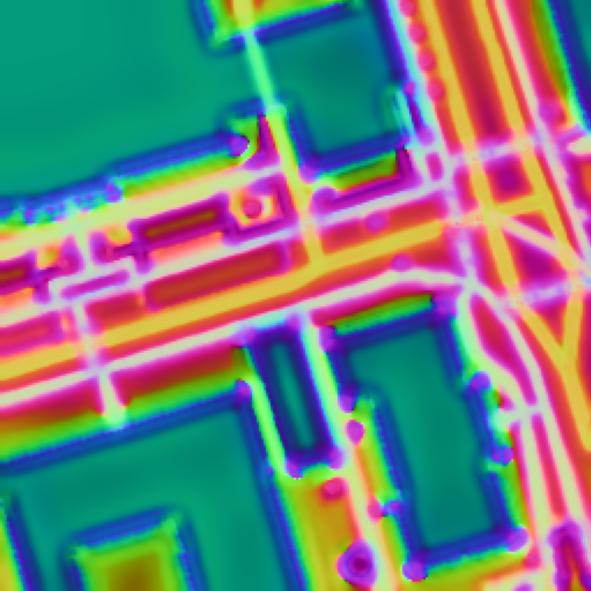} 
    & \includegraphics[valign=m,height=\plotH]{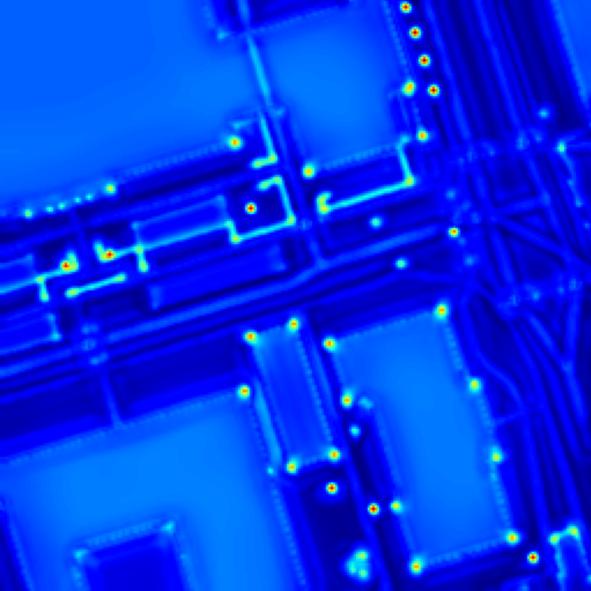} 
    & \includegraphics[valign=m,height=\plotH]{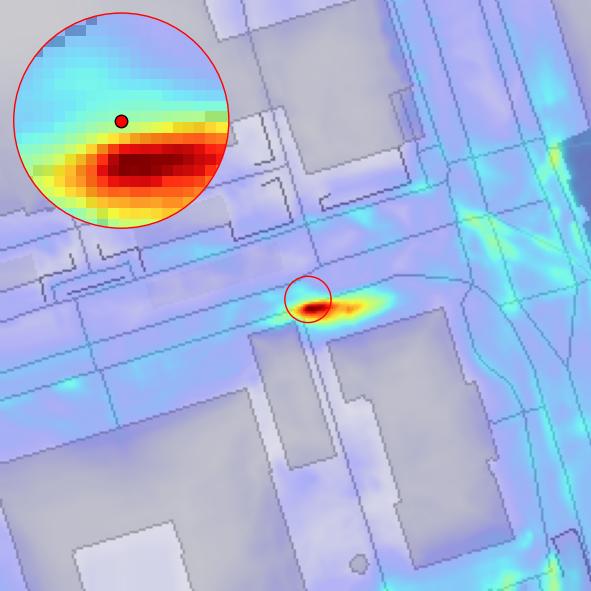} \\ 
    OrienterNet \cite{sarlin2023orienternet}
    & \includegraphics[valign=m,height=\plotH]{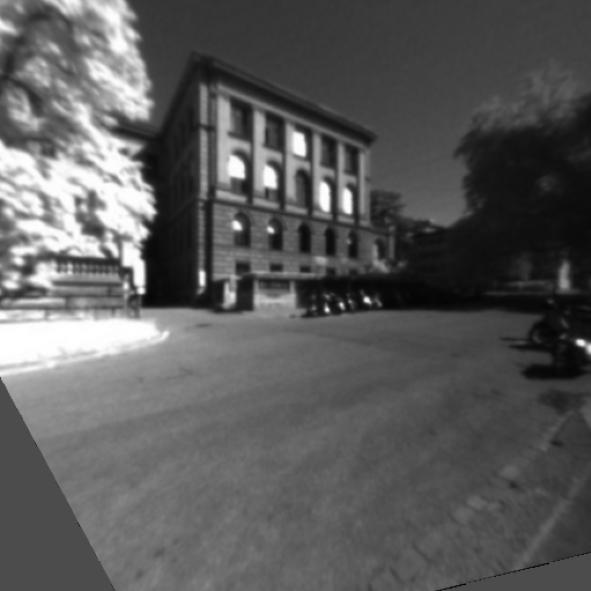} 
    & \includegraphics[valign=m,height=\plotH]{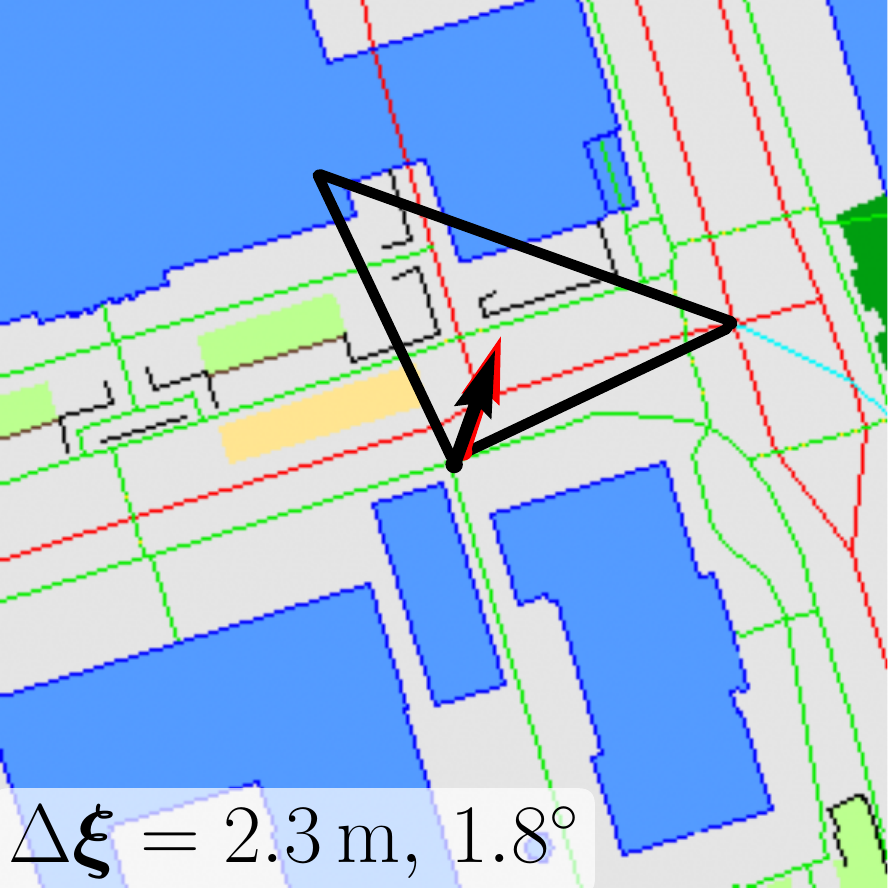} 
    & \includegraphics[valign=m,height=\plotH]{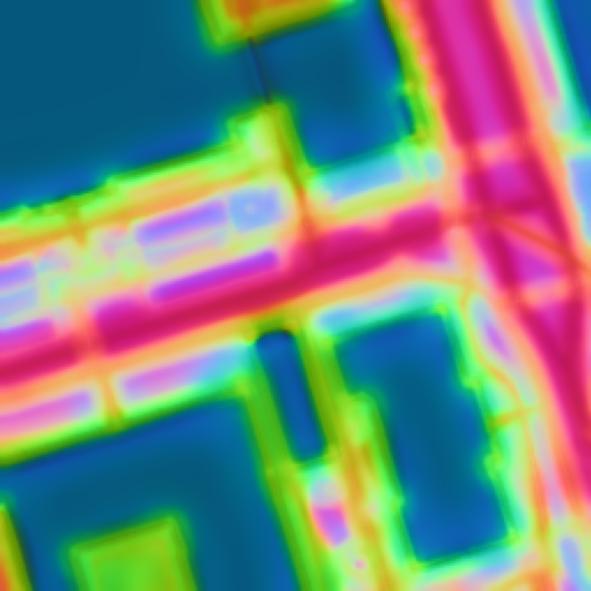} 
    & \includegraphics[valign=m,height=\plotH]{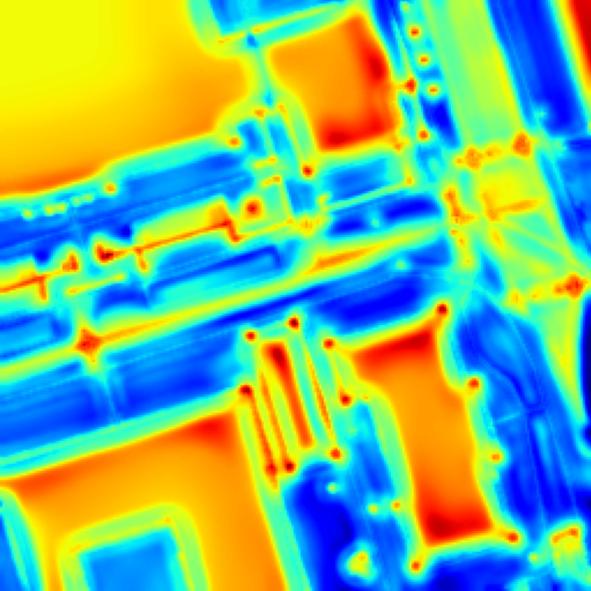} 
    & \includegraphics[valign=m,height=\plotH]{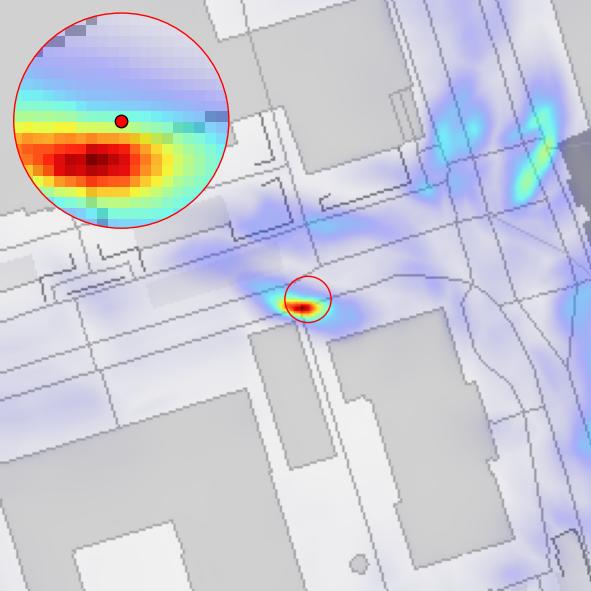} \\
    \end{tblr}
    \caption{\textbf{Qualitative results on LaMAria \cite{Krishnan2025lamaria}.} 
    On the map, we show the GT and predicted pose with a red \includegraphics[width=2mm]{figures/qual/red_arrow.png}, and black \includegraphics[width=2mm]{figures/qual/black_arrow.png} arrow, respectively. For each prediction, we also show its BEV frustum and its error, $\Delta\pose$.}
    \label{fig:supp_qual_lamaria}
\end{figure}

\begin{figure}[t]
    \def\plotH{1.15cm}
    \centering
    \begin{tblr}{
        colspec={*{6}c},
        colsep=1pt, 
        rowsep=1pt, 
        row{1,2,4}={belowsep=0.5pt},
        row{2,3,5}={abovesep=0.5pt},
        row{Y}={belowsep=0pt},
        row{Z}={abovesep=0pt},
        stretch=0pt,
        cell{2-Z}{1}={}{halign=c, valign=m,cmd={\scriptsize\rotatebox[origin=c]{90}}},
    }
    & Images
    & Input Map & Neural Map & Features Norm & Probabilities \\
    \name
    & \includegraphics[valign=m,height=\plotH]{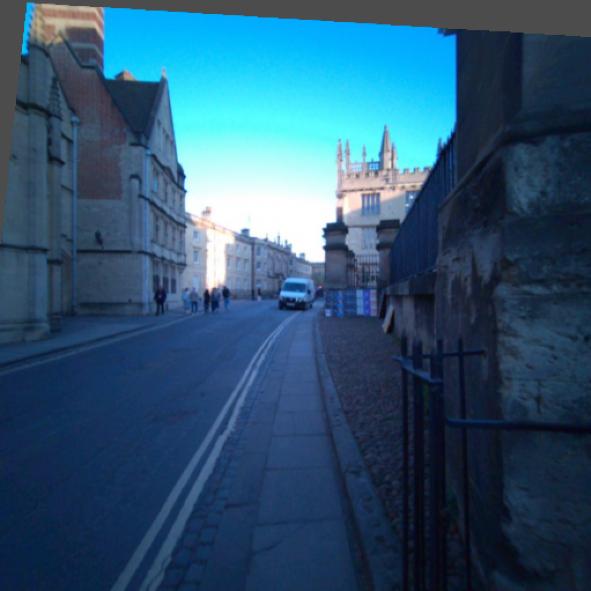} 
    & \includegraphics[valign=m,height=\plotH]{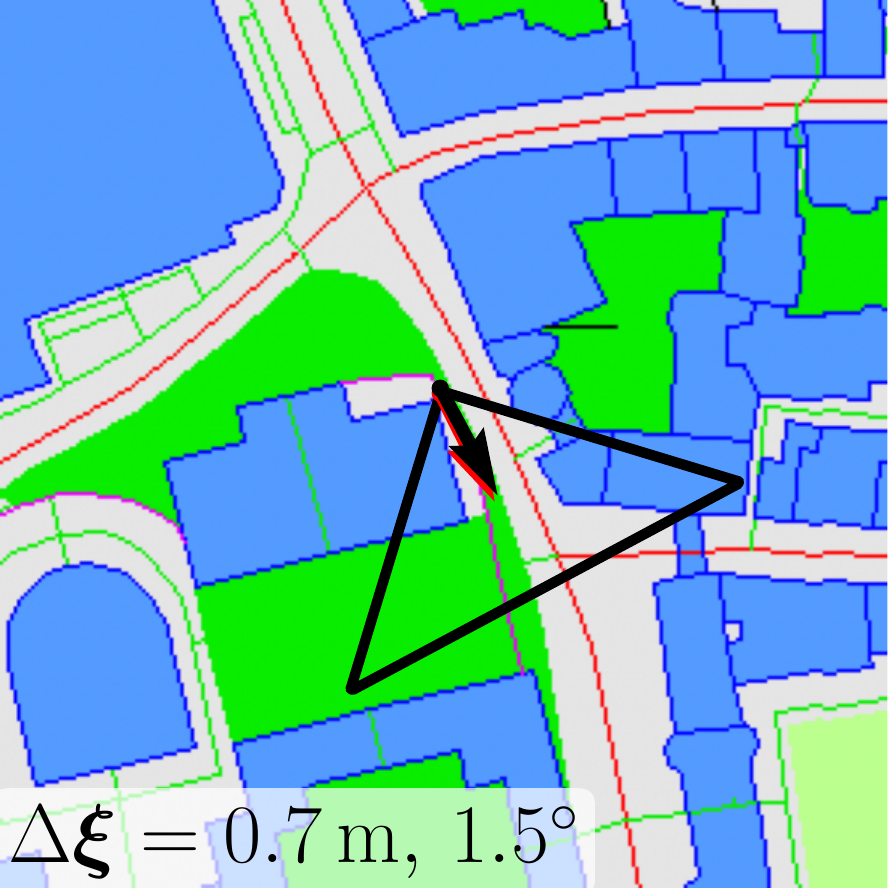} 
    & \includegraphics[valign=m,height=\plotH]{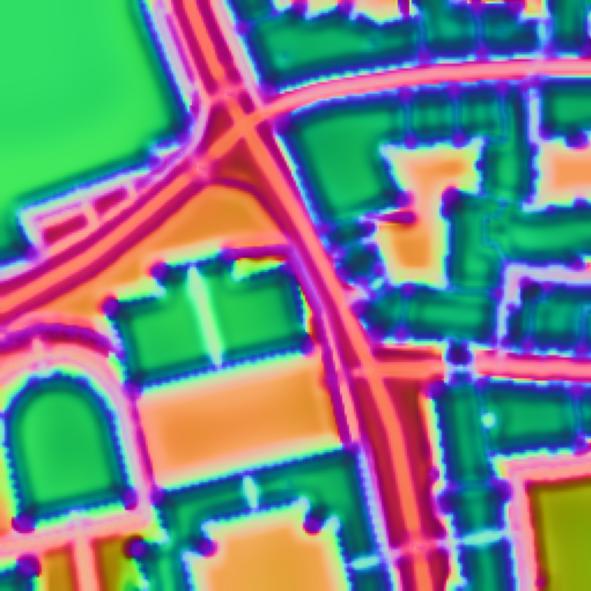} 
    & \includegraphics[valign=m,height=\plotH]{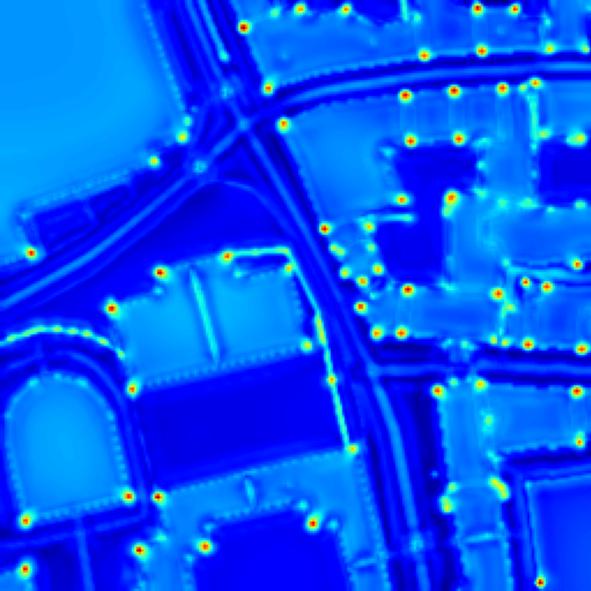} 
    & \includegraphics[valign=m,height=\plotH]{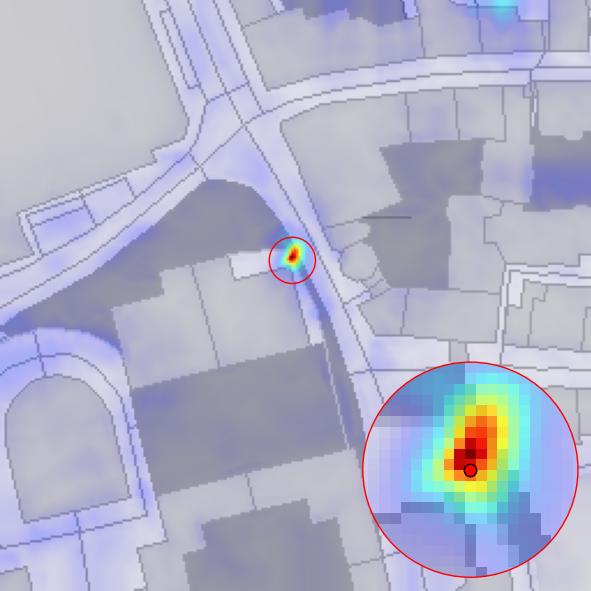} \\
    OrienterNet \cite{sarlin2023orienternet}
    & \includegraphics[valign=m,height=\plotH]{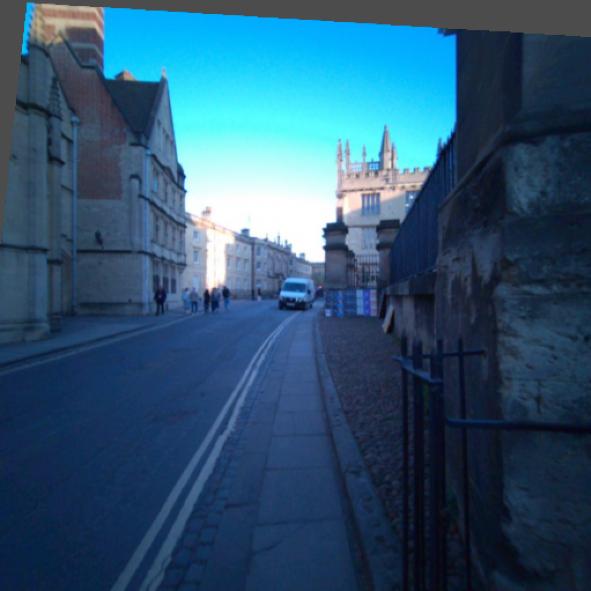} 
    & \includegraphics[valign=m,height=\plotH]{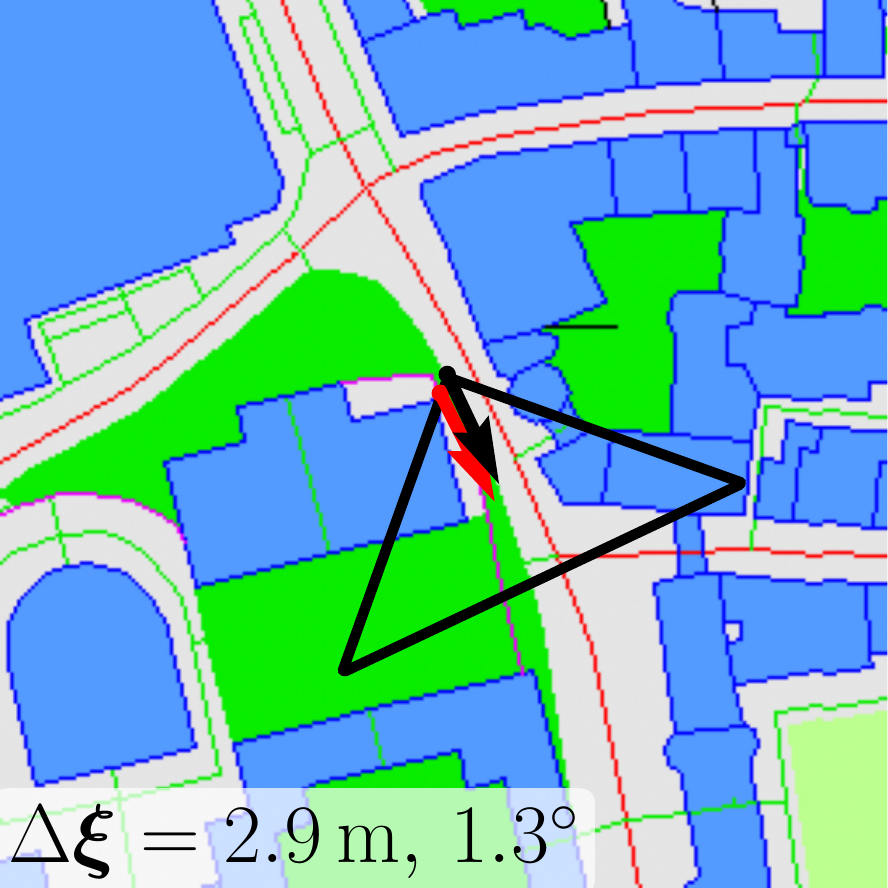} 
    & \includegraphics[valign=m,height=\plotH]{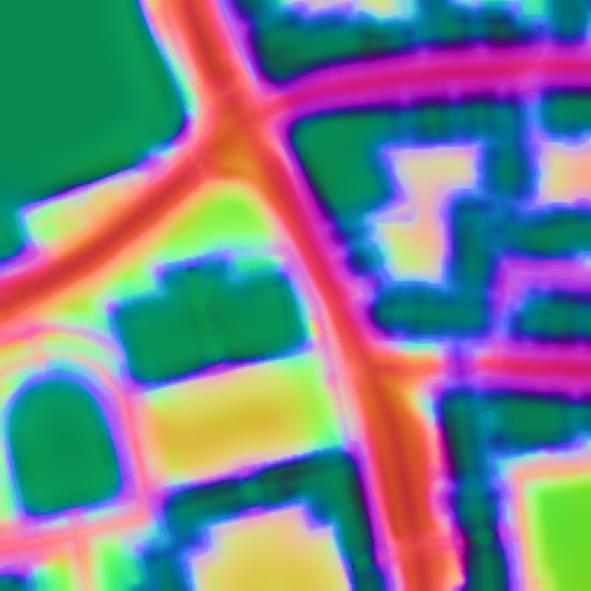} 
    & \includegraphics[valign=m,height=\plotH]{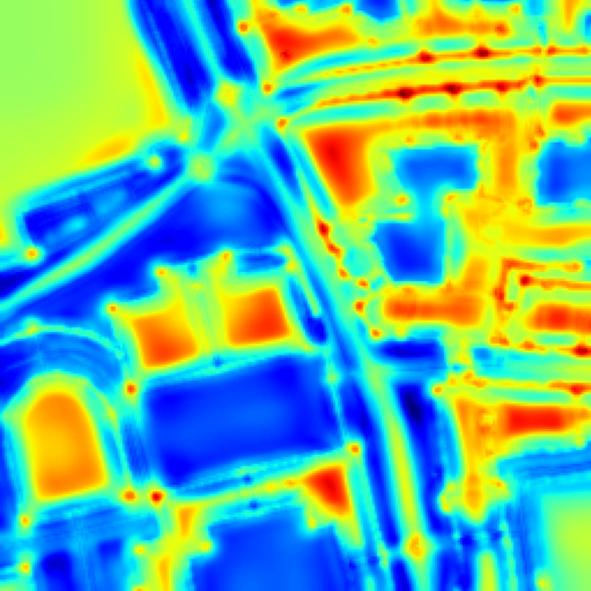} 
    & \includegraphics[valign=m,height=\plotH]{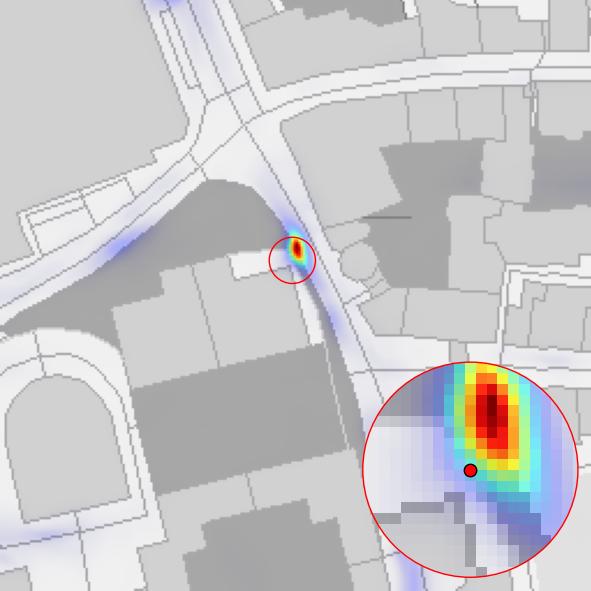} \\
    \name
    & \includegraphics[valign=m,height=\plotH]{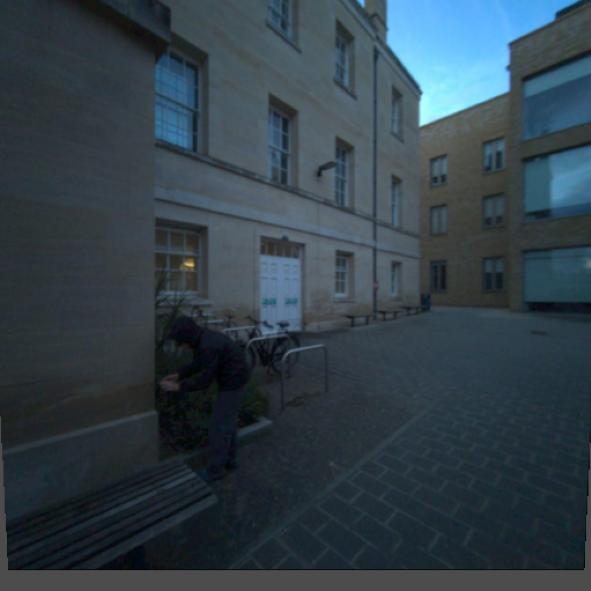} 
    & \includegraphics[valign=m,height=\plotH]{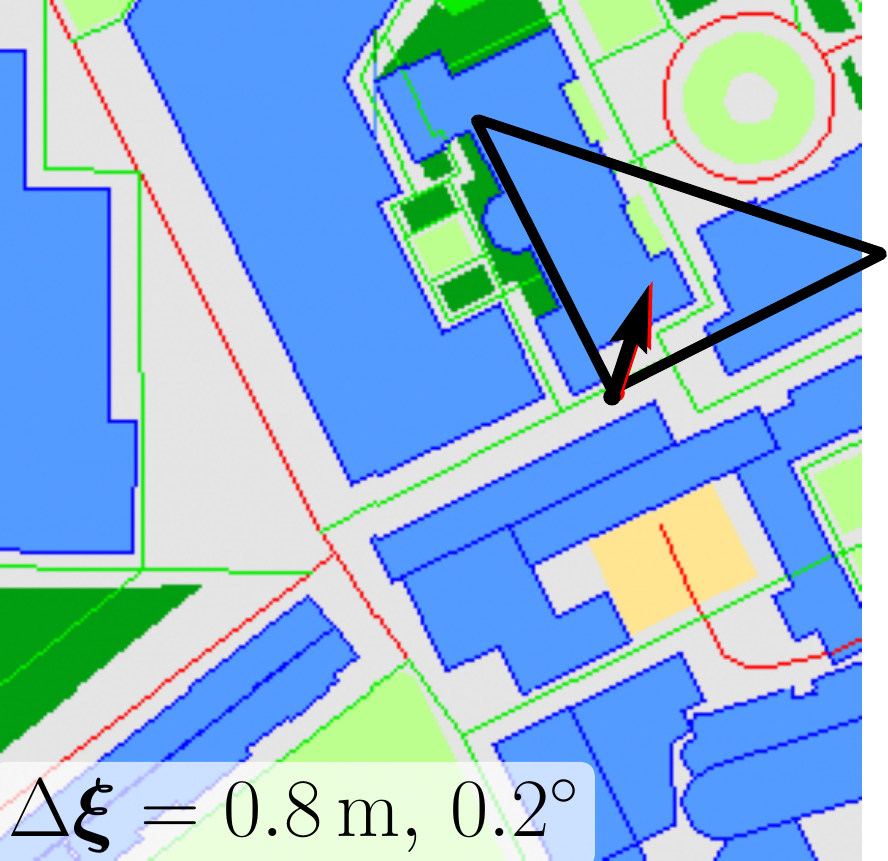} 
    & \includegraphics[valign=m,height=\plotH]{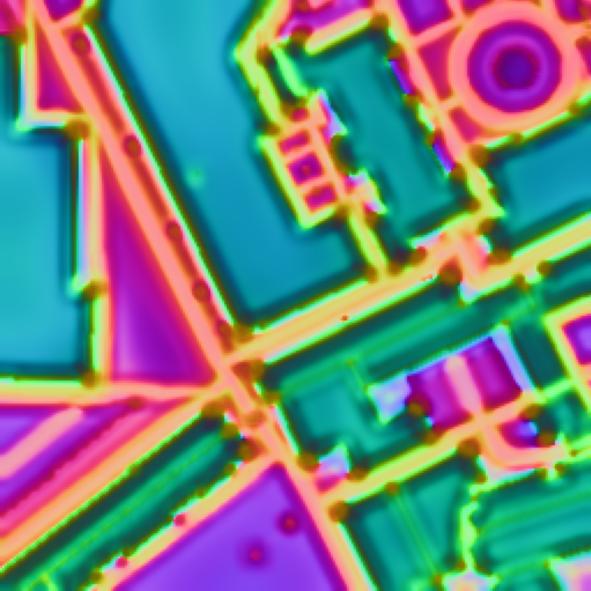} 
    & \includegraphics[valign=m,height=\plotH]{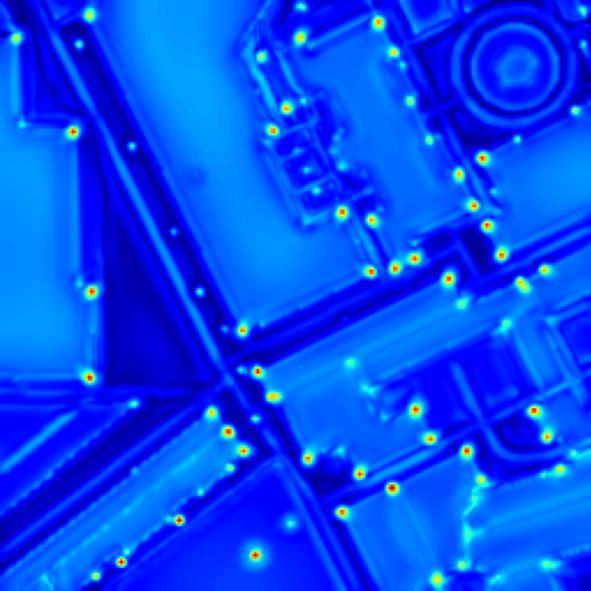} 
    & \includegraphics[valign=m,height=\plotH]{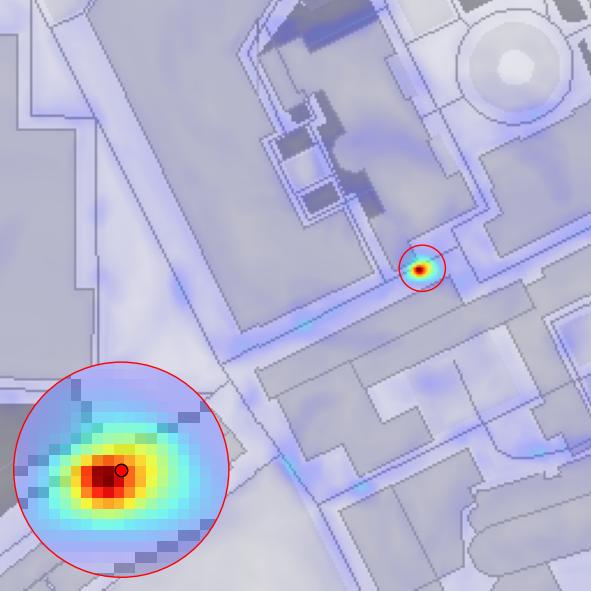} \\
    OSMLoc \cite{liao2024osmloc}
    & \includegraphics[valign=m,height=\plotH]{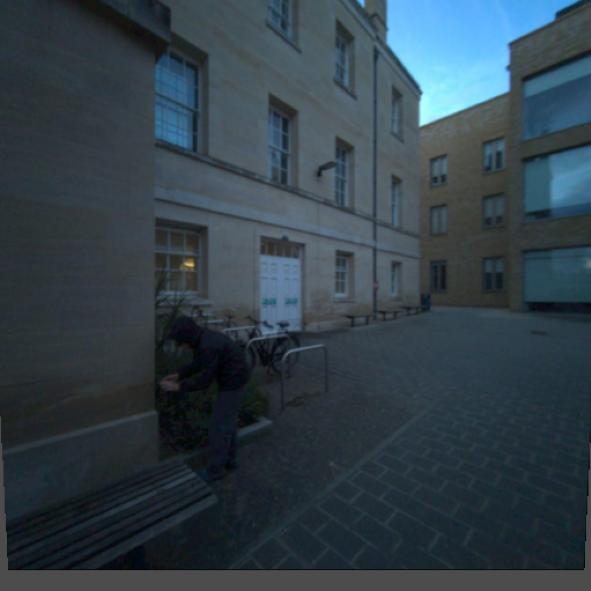} 
    & \includegraphics[valign=m,height=\plotH]{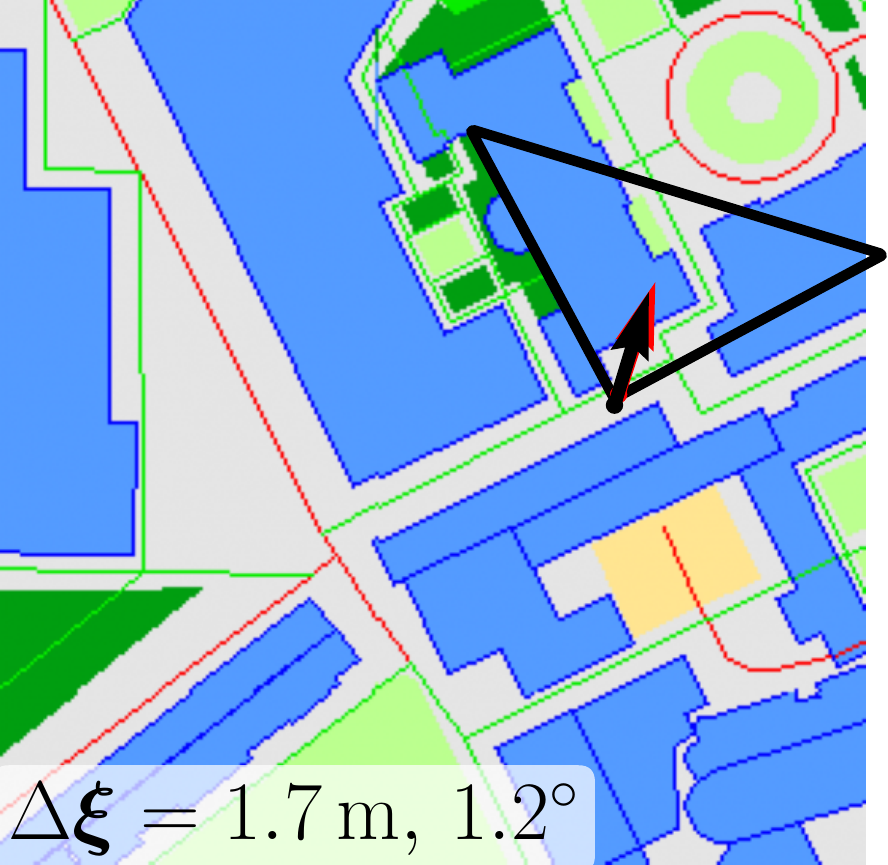} 
    & \includegraphics[valign=m,height=\plotH]{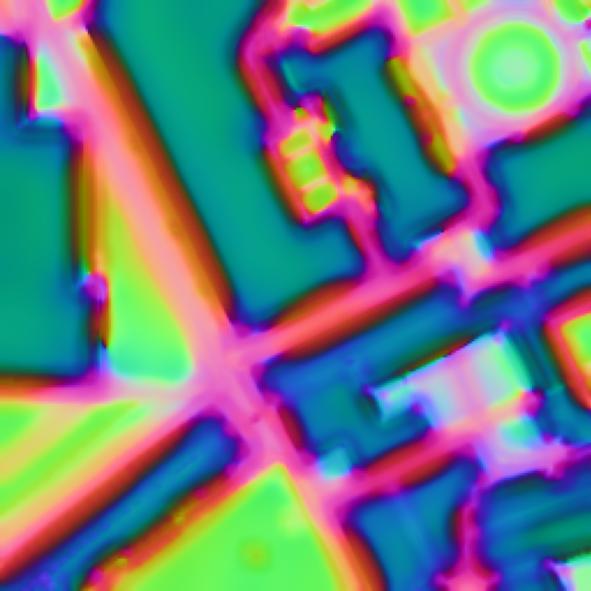} 
    & \includegraphics[valign=m,height=\plotH]{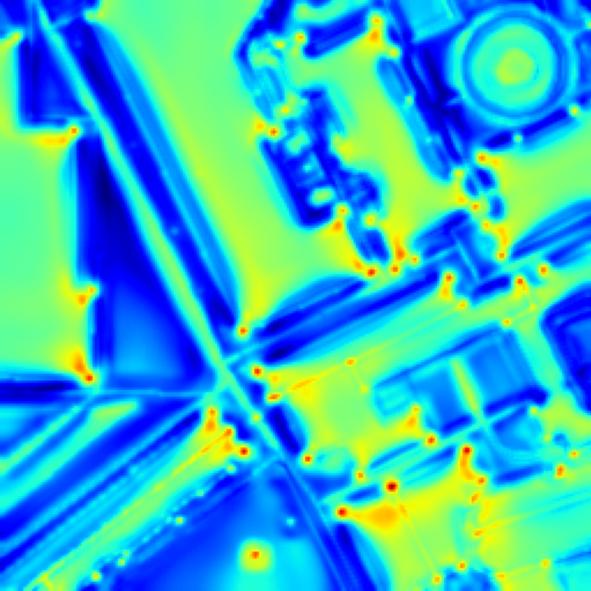} 
    & \includegraphics[valign=m,height=\plotH]{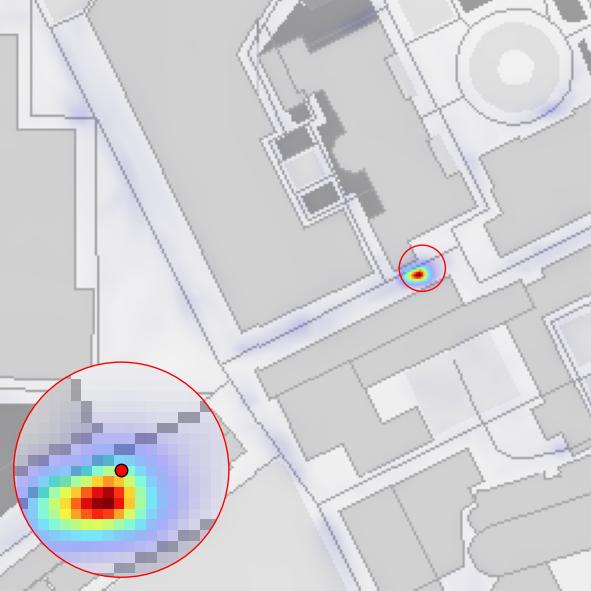} \\
    \name
    & \includegraphics[valign=m,height=\plotH]{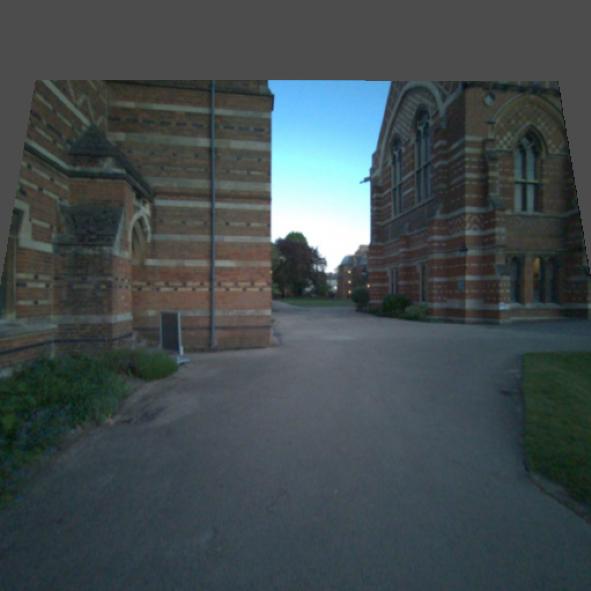} 
    & \includegraphics[valign=m,height=\plotH]{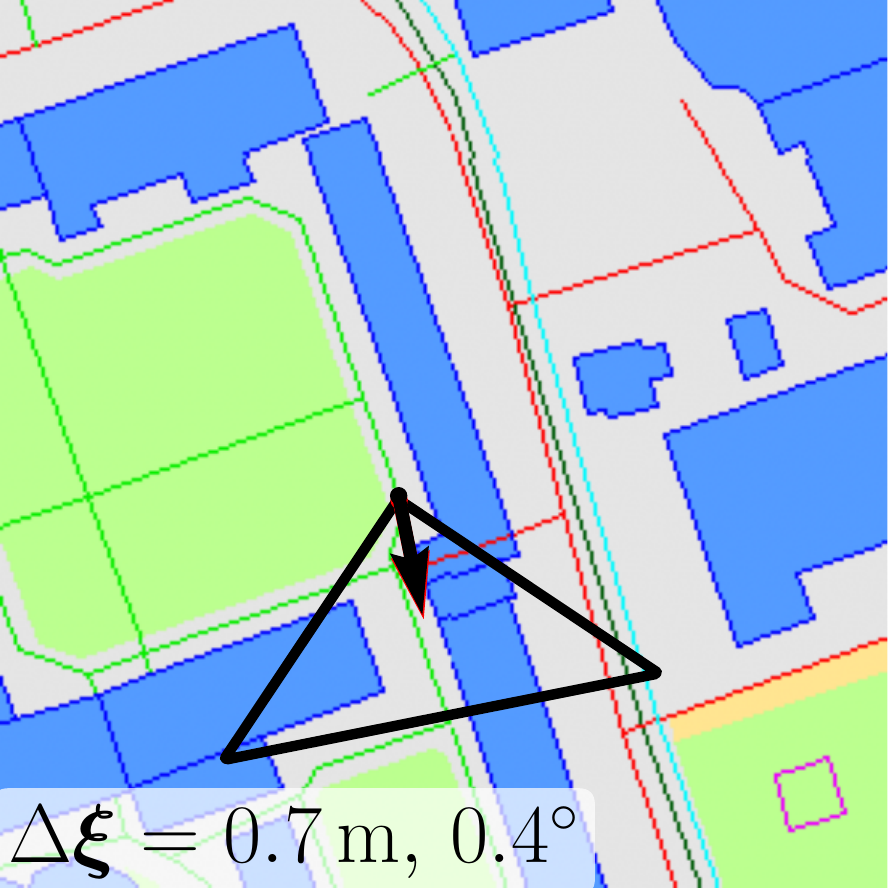} 
    & \includegraphics[valign=m,height=\plotH]{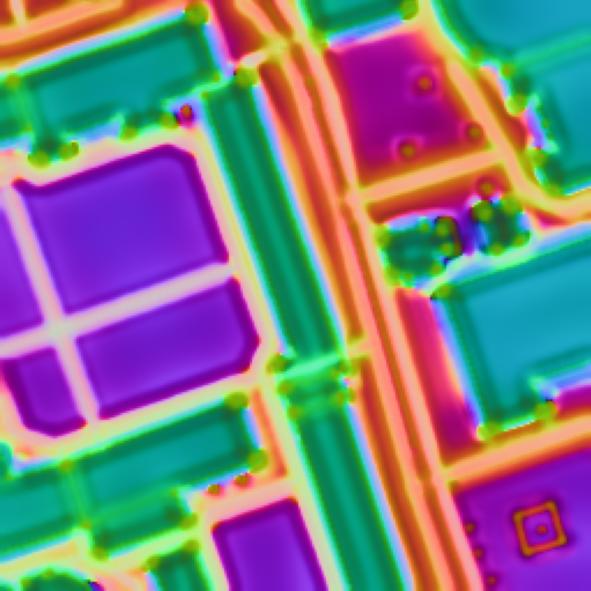} 
    & \includegraphics[valign=m,height=\plotH]{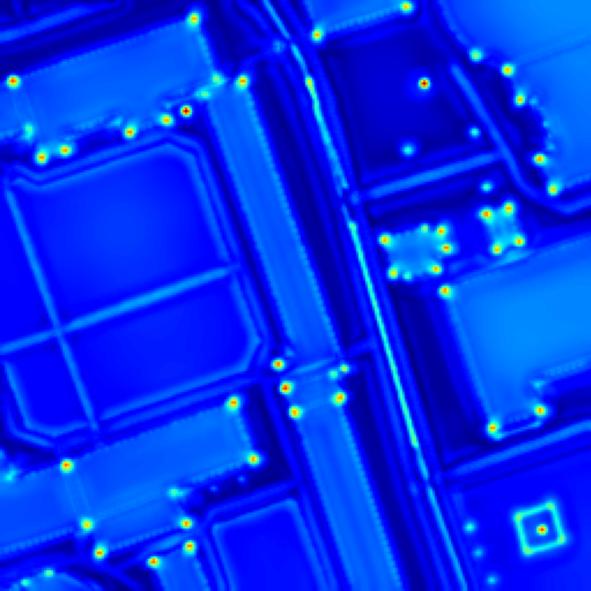} 
    & \includegraphics[valign=m,height=\plotH]{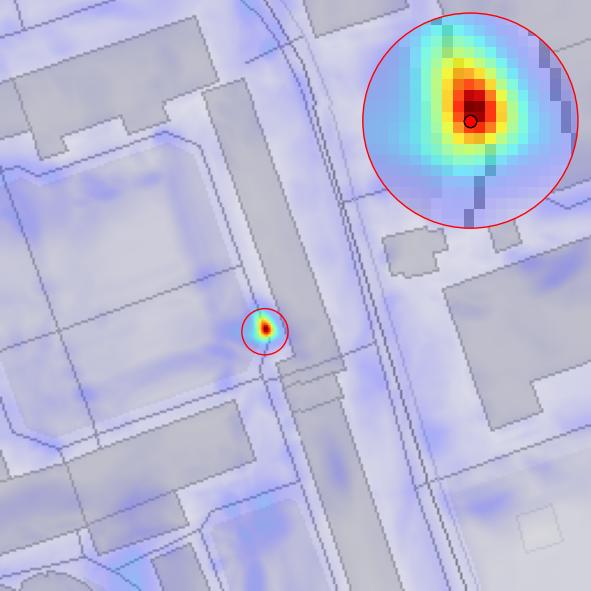} \\ 
    OrienterNet \cite{sarlin2023orienternet}
    & \includegraphics[valign=m,height=\plotH]{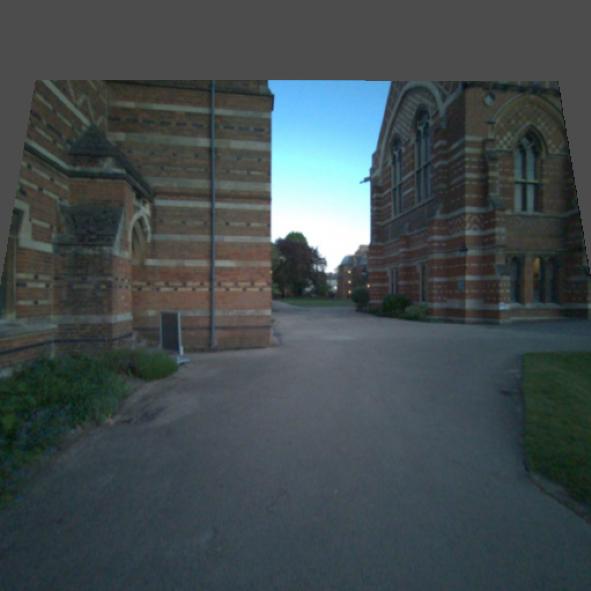} 
    & \includegraphics[valign=m,height=\plotH]{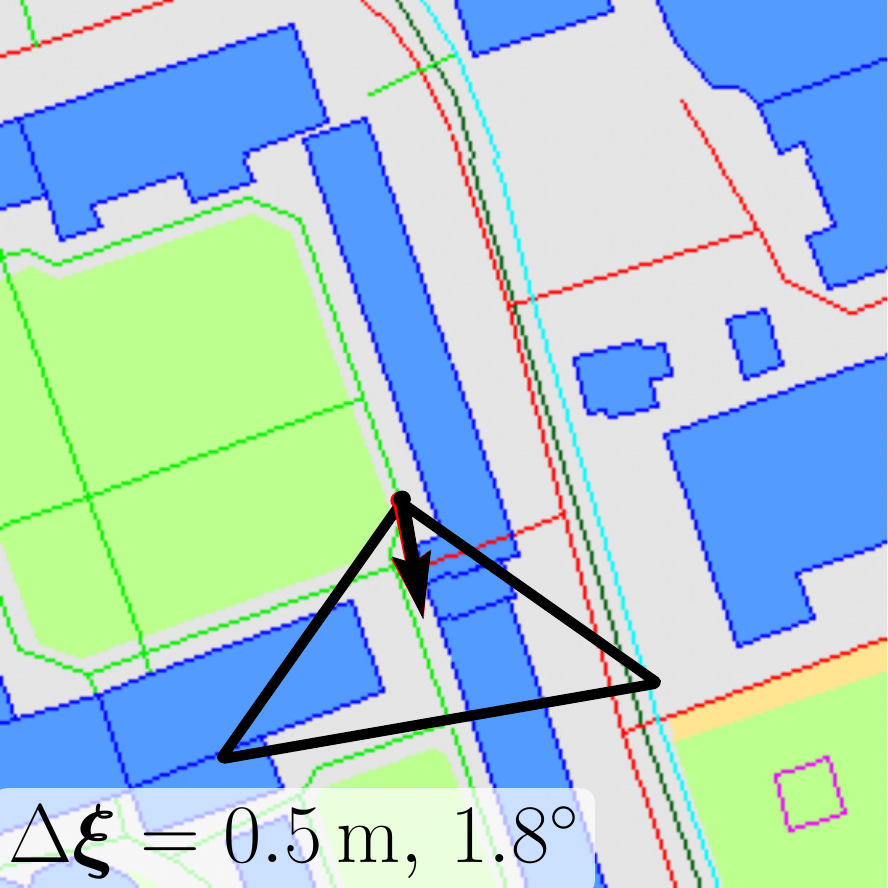} 
    & \includegraphics[valign=m,height=\plotH]{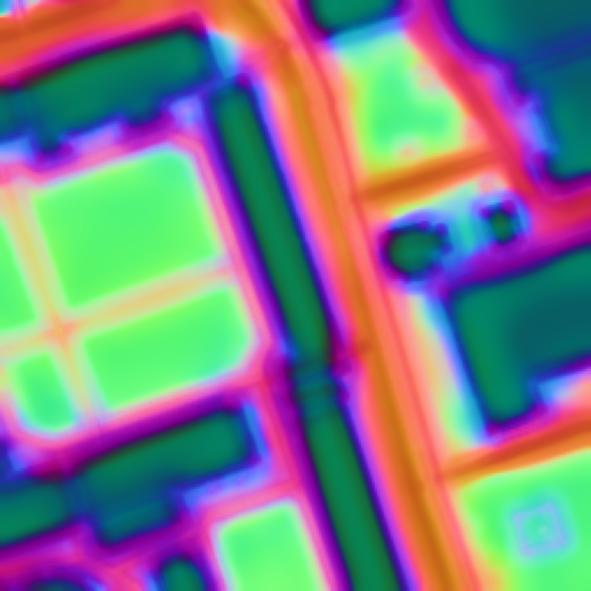} 
    & \includegraphics[valign=m,height=\plotH]{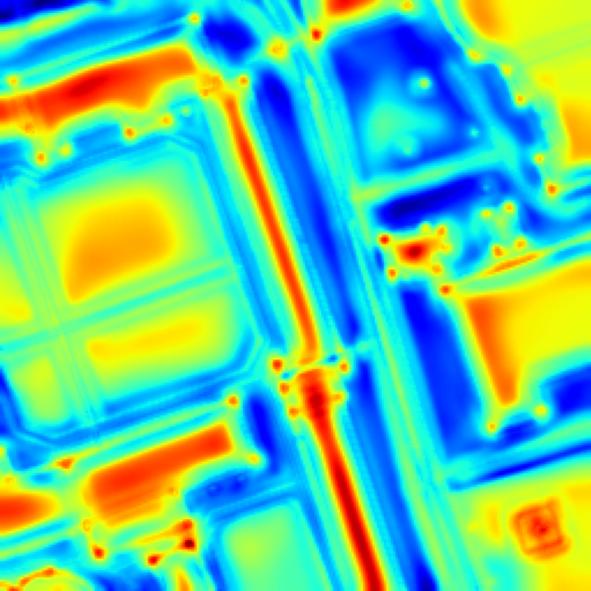} 
    & \includegraphics[valign=m,height=\plotH]{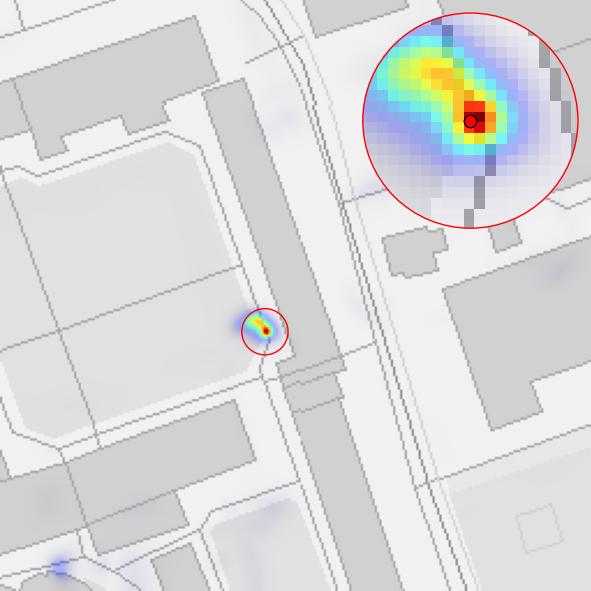} \\
    \end{tblr}
    \caption{\textbf{Qualitative results on Oxford Day \& Night (ODN) \cite{wang2025oxford}.}
    On the map, we show the GT and predicted pose with a red \includegraphics[width=2mm]{figures/qual/red_arrow.png}, and black \includegraphics[width=2mm]{figures/qual/black_arrow.png} arrow, respectively. For each prediction, we also show its BEV frustum and error, $\Delta\pose$. Since the geo-referenced poses in ODN can be offset (\cref{fig:supp_oxford_align}), and in order to have the same GT with which to compare and compute $\Delta\pose$, for just these examples we precompute an alignment using an independent \name model, trained with Chunk$_{r=20}+\Delta\Theta+\Delta XY$.}
    \label{fig:supp_qual_oxford}
\end{figure}

\begin{figure}[t]
    \def\plotH{0.87cm}
    \def\plotHI{1.15cm}
    \centering
    \begin{tblr}{
        colspec={*{6}c},
        colsep=1pt, 
        rowsep=1pt, 
        row{1,2,4}={belowsep=0.5pt},
        row{2,3,5}={abovesep=0.5pt},
        row{Y}={belowsep=0pt},
        row{Z}={abovesep=0pt},
        stretch=0pt,
        row{1}={}{cmd=\scriptsize},
        cell{2-Z}{1}={}{halign=c, valign=m,cmd={\tiny\rotatebox[origin=c]{90}}},
    }
    & Images
    & Input Map & Neural Map & Norms & Probs. \\
    \name
    & \includegraphics[valign=m,height=\plotH]{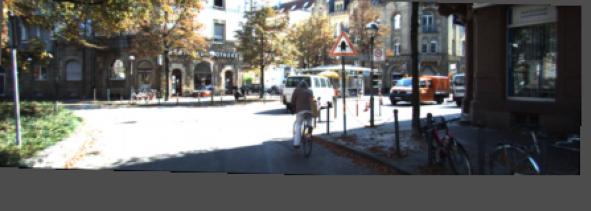} 
    & \includegraphics[valign=m,height=\plotH]{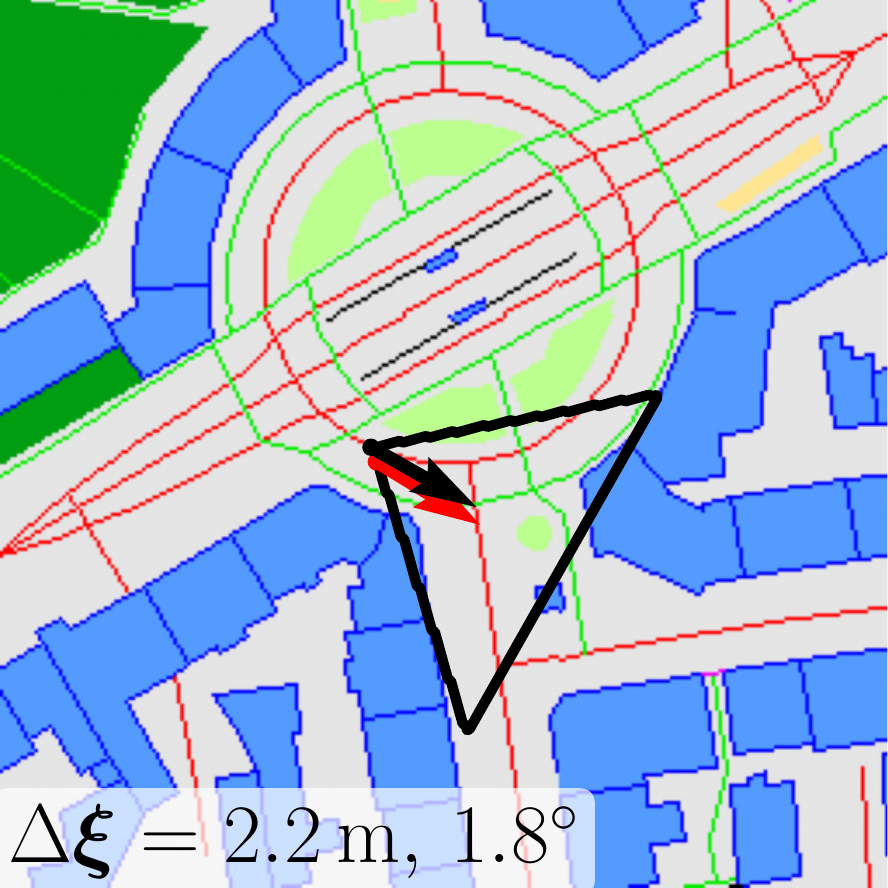} 
    & \includegraphics[valign=m,height=\plotH]{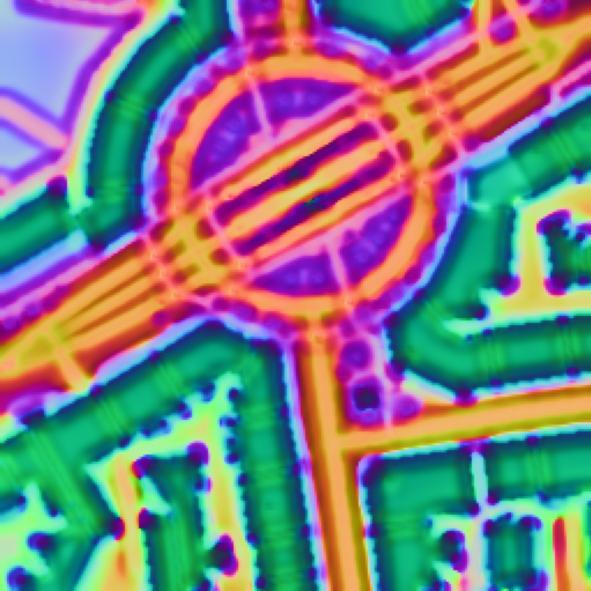} 
    & \includegraphics[valign=m,height=\plotH]{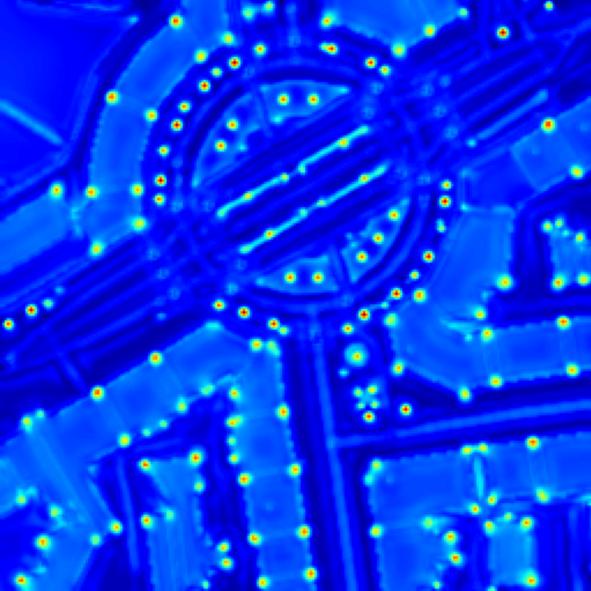} 
    & \includegraphics[valign=m,height=\plotH]{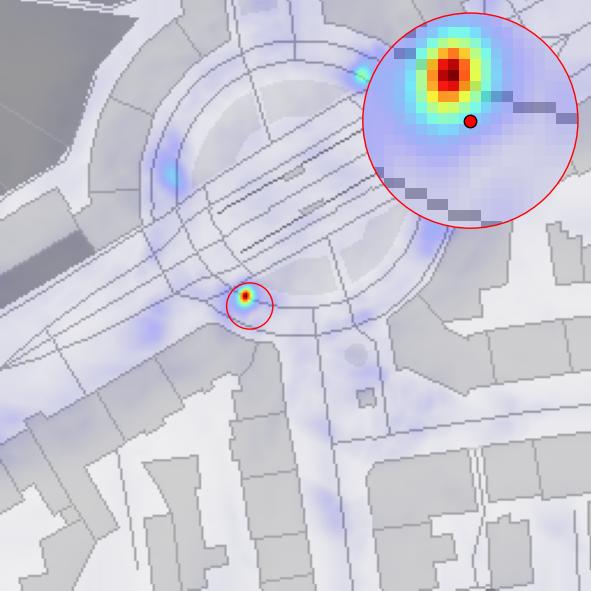} \\
    OrienterNet \cite{sarlin2023orienternet}
    & \includegraphics[valign=m,height=\plotH]{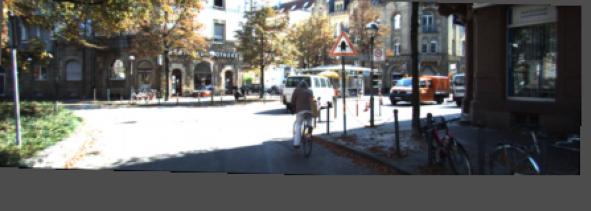} 
    & \includegraphics[valign=m,height=\plotH]{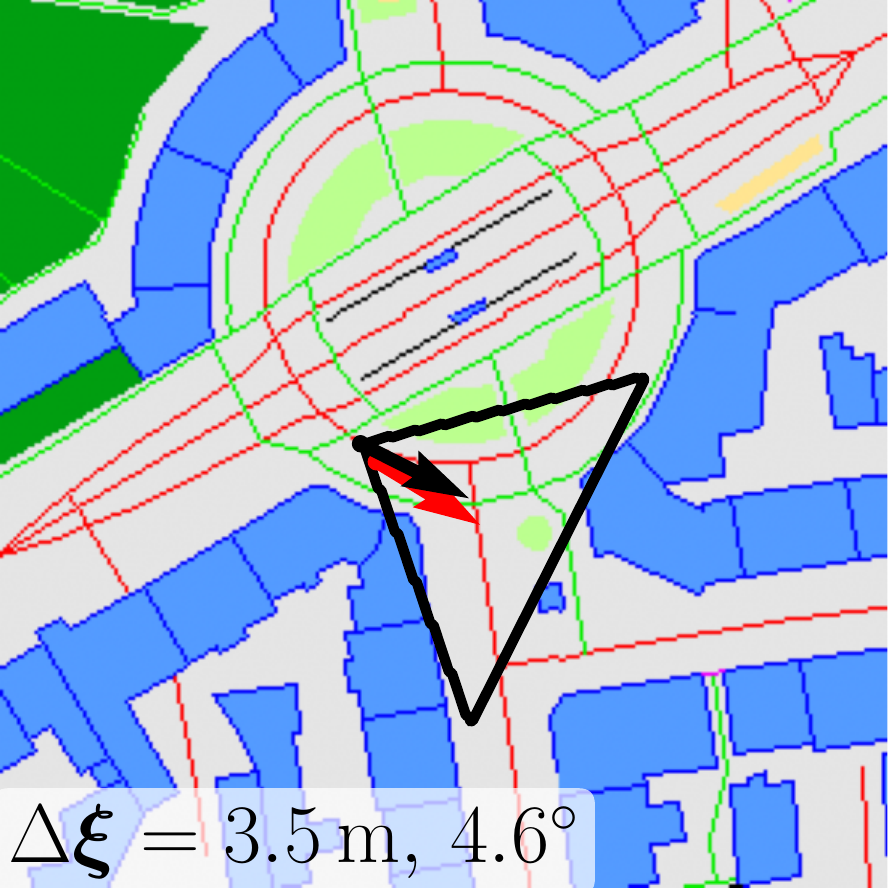} 
    & \includegraphics[valign=m,height=\plotH]{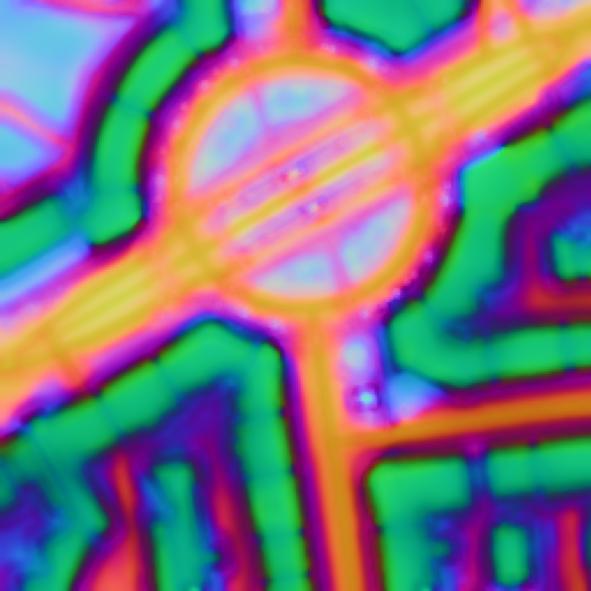} 
    & \includegraphics[valign=m,height=\plotH]{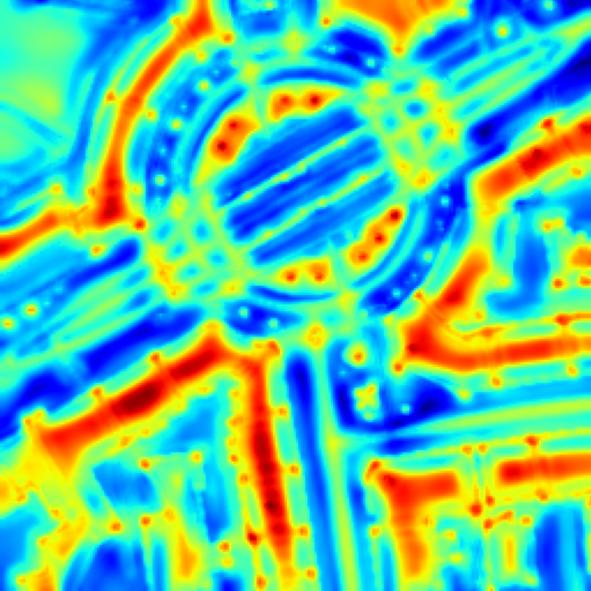} 
    & \includegraphics[valign=m,height=\plotH]{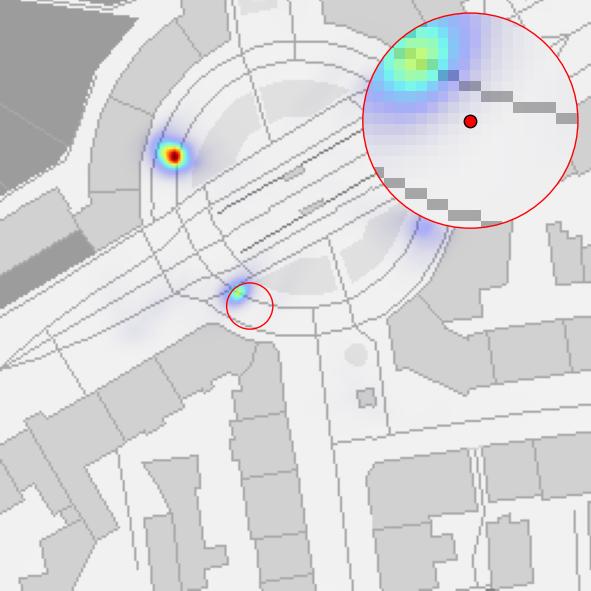} \\
    \name
    & \includegraphics[valign=m,height=\plotH]{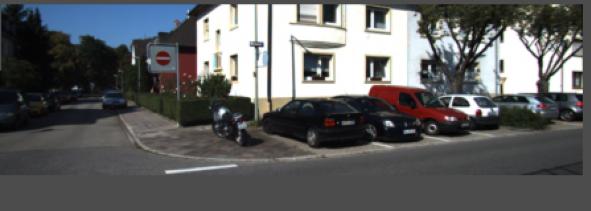} 
    & \includegraphics[valign=m,height=\plotH]{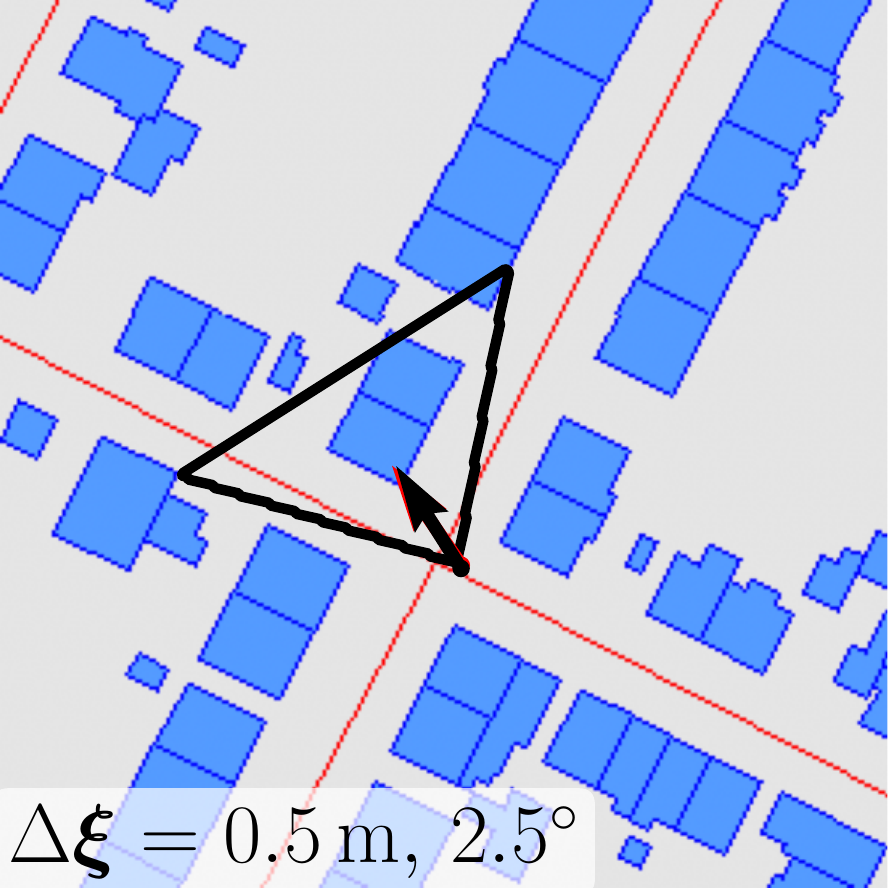} 
    & \includegraphics[valign=m,height=\plotH]{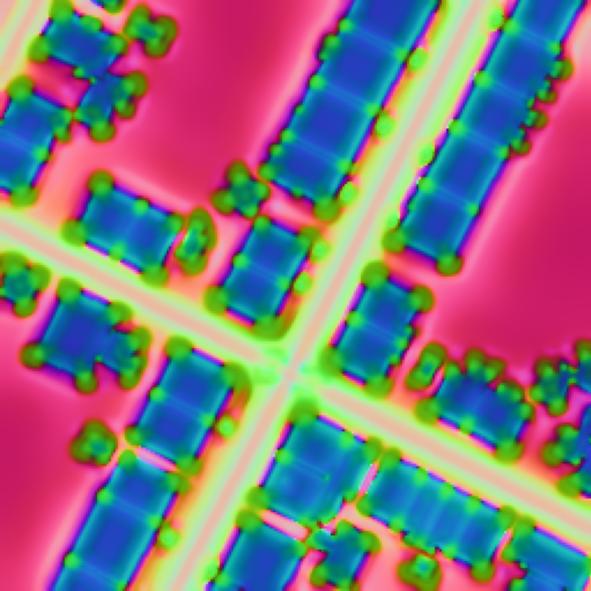} 
    & \includegraphics[valign=m,height=\plotH]{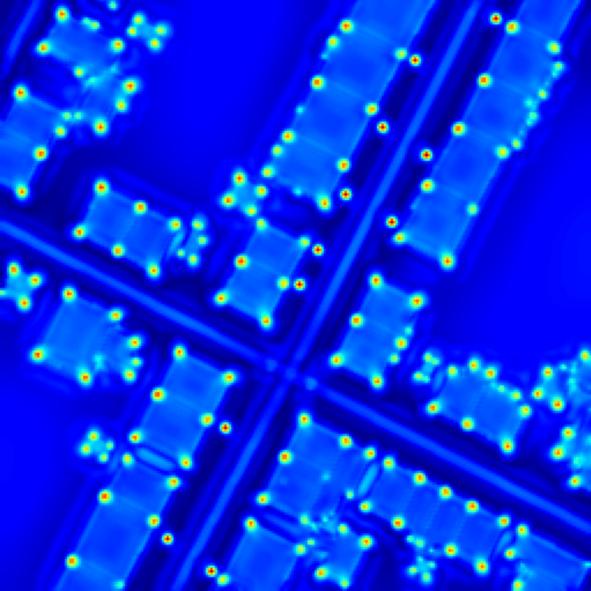} 
    & \includegraphics[valign=m,height=\plotH]{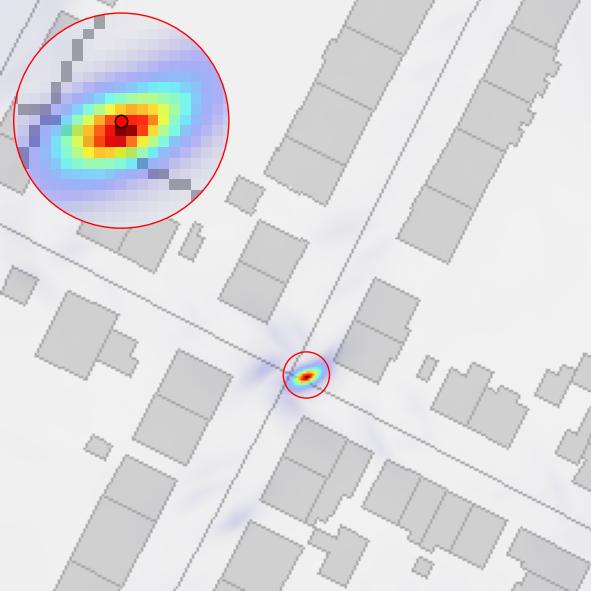} \\
    OSMLoc \cite{liao2024osmloc}
    & \includegraphics[valign=m,height=\plotH]{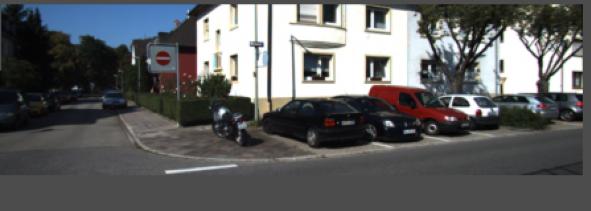} 
    & \includegraphics[valign=m,height=\plotH]{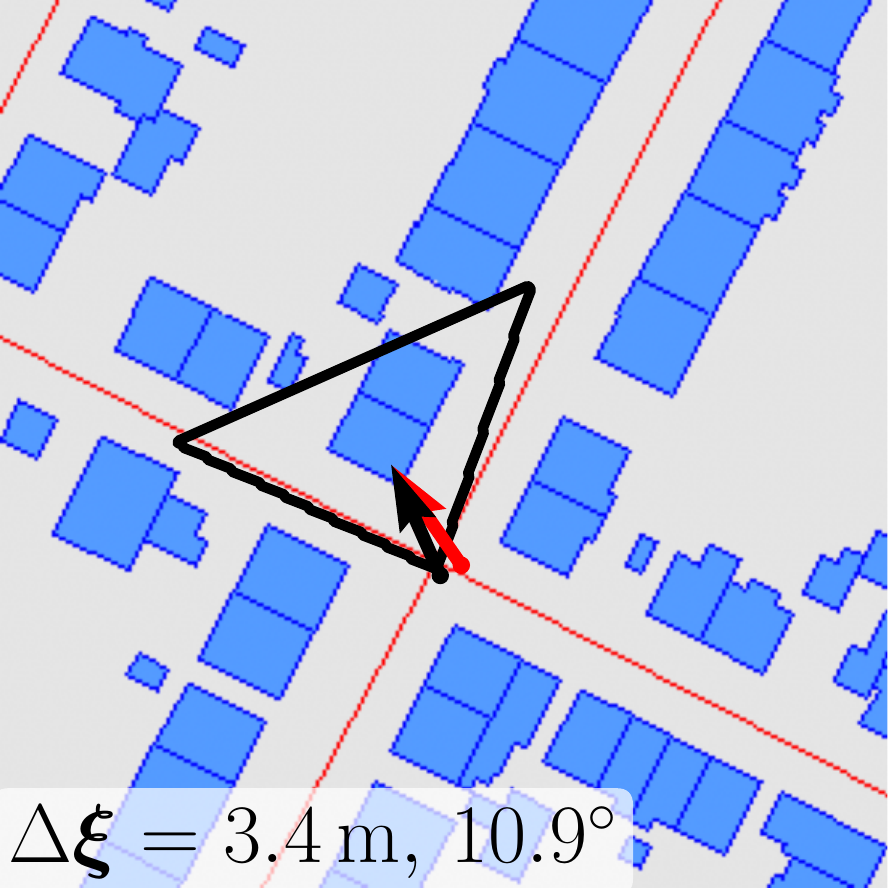} 
    & \includegraphics[valign=m,height=\plotH]{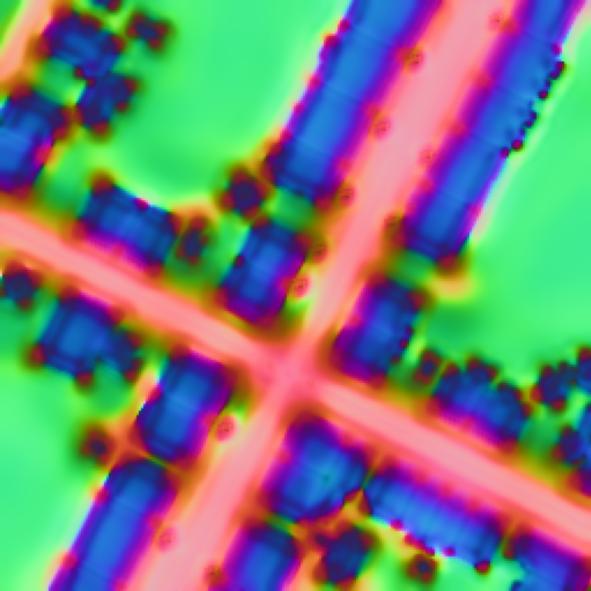} 
    & \includegraphics[valign=m,height=\plotH]{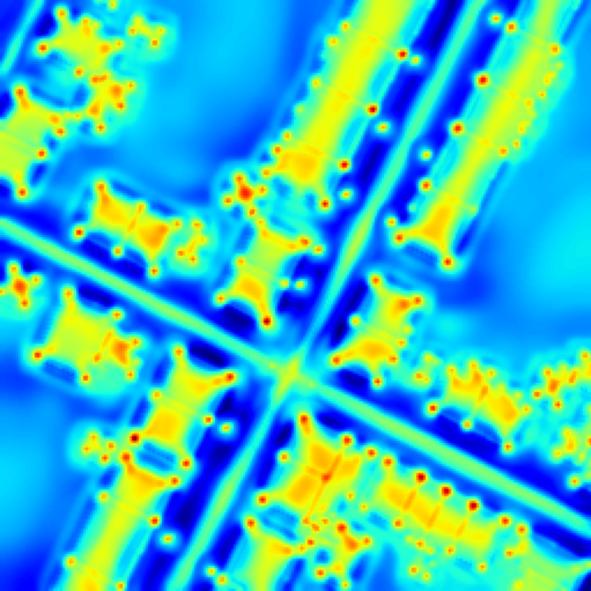} 
    & \includegraphics[valign=m,height=\plotH]{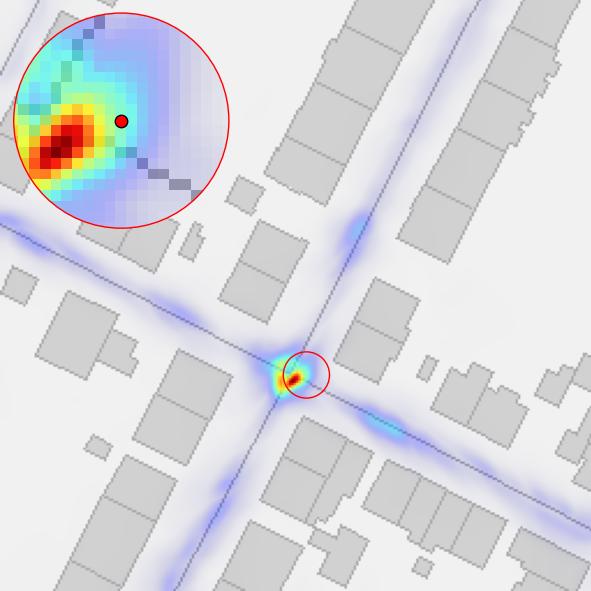} \\
    \name
    & \includegraphics[valign=m,height=\plotH]{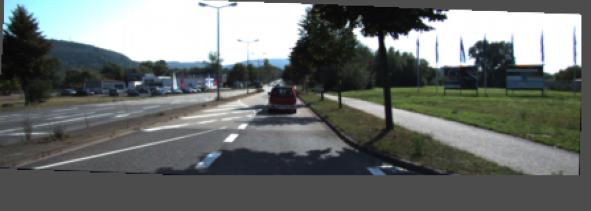} 
    & \includegraphics[valign=m,height=\plotH]{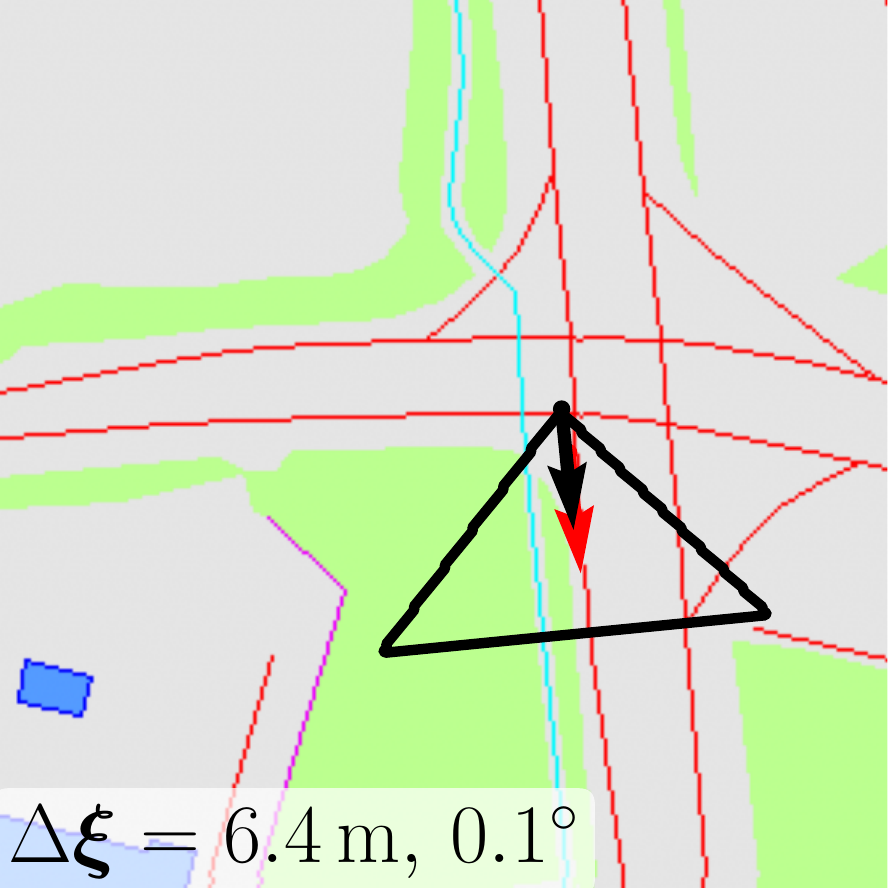} 
    & \includegraphics[valign=m,height=\plotH]{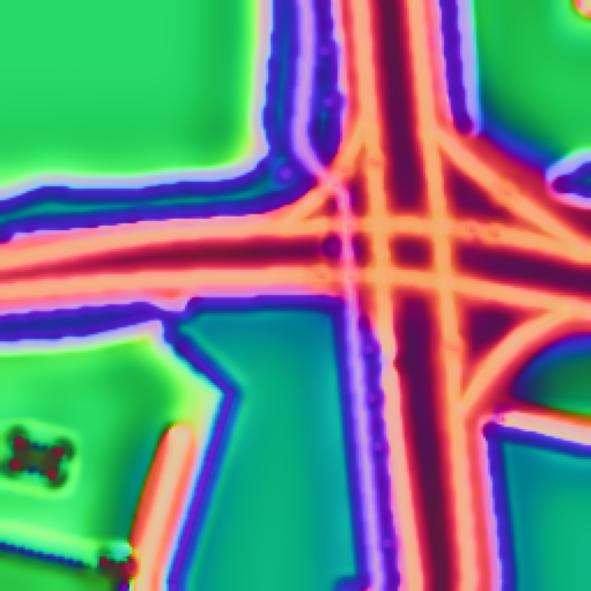} 
    & \includegraphics[valign=m,height=\plotH]{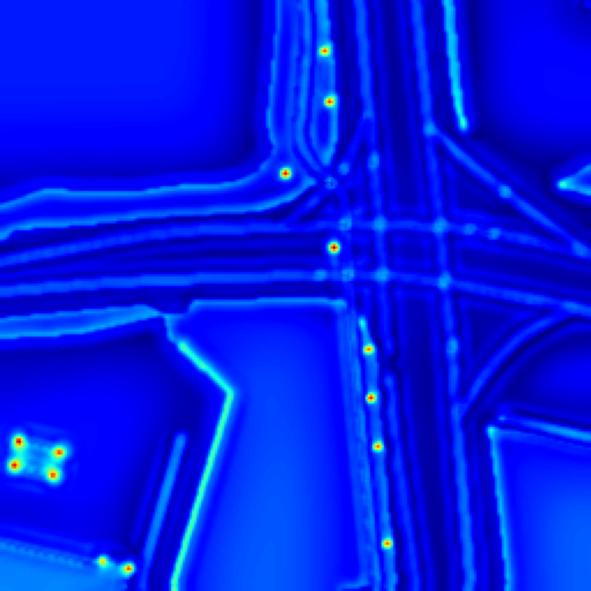} 
    & \includegraphics[valign=m,height=\plotH]{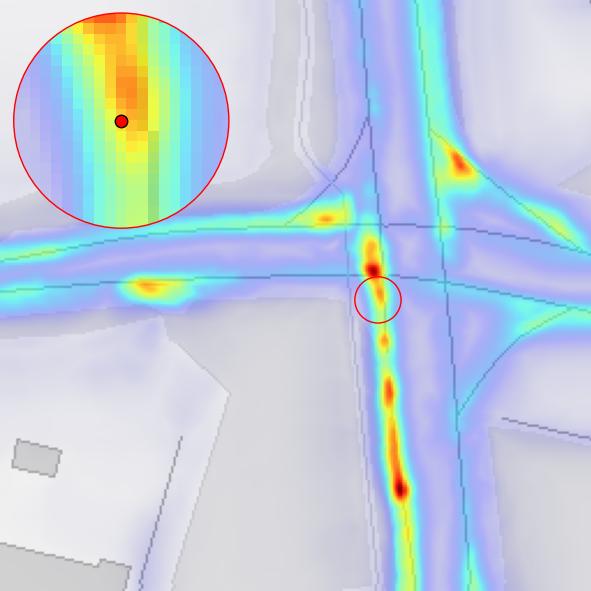} \\
    OrienterNet \cite{sarlin2023orienternet}
    & \includegraphics[valign=m,height=\plotH]{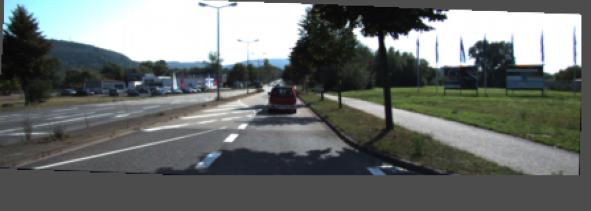} 
    & \includegraphics[valign=m,height=\plotH]{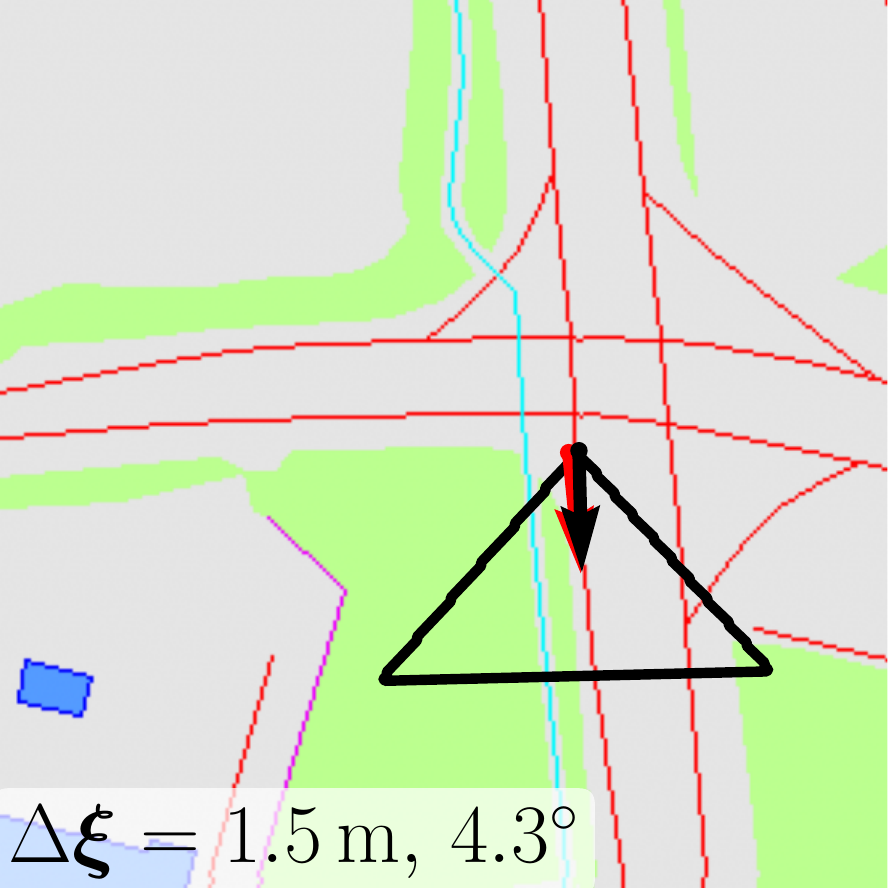} 
    & \includegraphics[valign=m,height=\plotH]{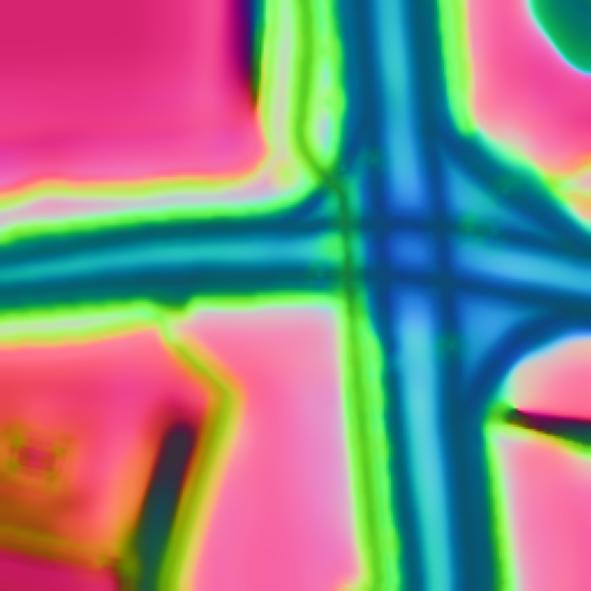} 
    & \includegraphics[valign=m,height=\plotH]{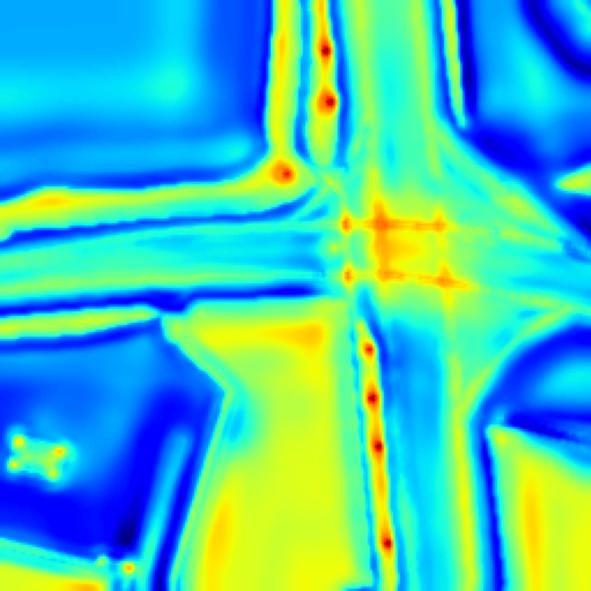} 
    & \includegraphics[valign=m,height=\plotH]{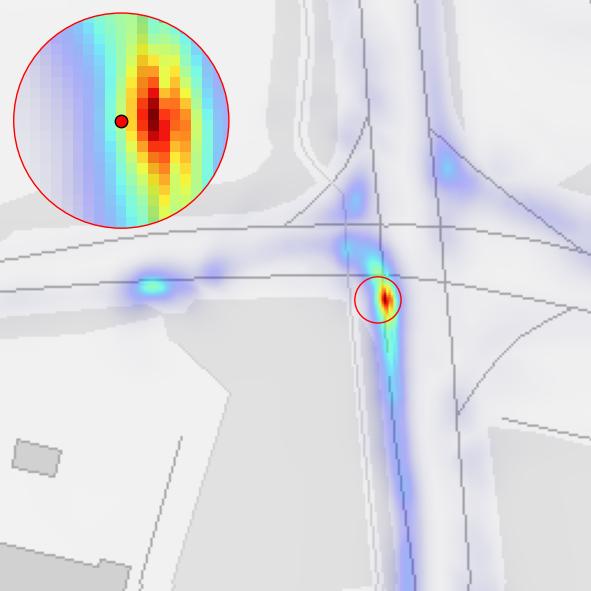} \\ 
    \end{tblr}
    \caption{\textbf{Qualitative results on KITTI \cite{geiger2012kitti}.} 
    On the map, we show the GT and predicted pose with a red \includegraphics[width=2mm]{figures/qual/red_arrow.png}, and black \includegraphics[width=2mm]{figures/qual/black_arrow.png} arrow, respectively. For each prediction, we also show its BEV frustum and its error, $\Delta\pose$.}
    \label{fig:supp_qual_kitti}
\end{figure}

\begin{figure}[t]
    \def\plotH{1.15cm}
    \def\bevH{0.58cm}
    \centering
    \resizebox{1\linewidth}{!}{%
    \begin{tblr}{
        colspec={*{6}c},
        colsep=1pt, 
        rowsep=1pt, 
        column{4}={rightsep+=5pt},
        column{5}={leftsep+=5pt},
        cell{1-Z}{1}={}{halign=c, valign=m,cmd={\scriptsize\rotatebox[origin=c]{90}}},
    }
    Inputs 
    & \includegraphics[valign=m,height=\plotH]{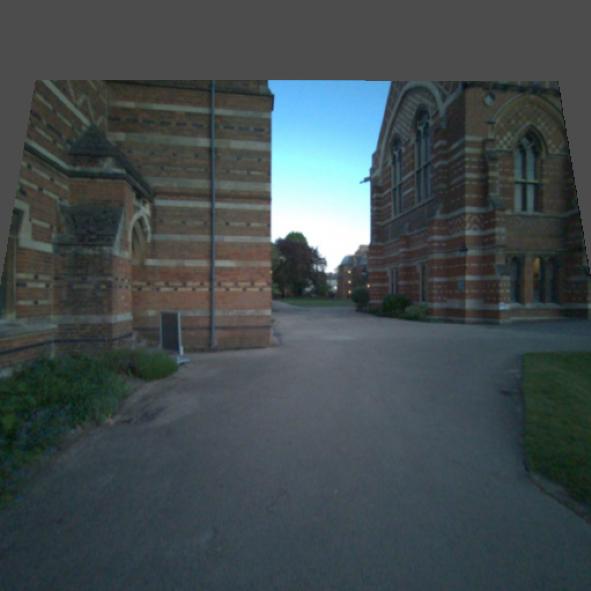}
    & \includegraphics[valign=m,height=\plotH]{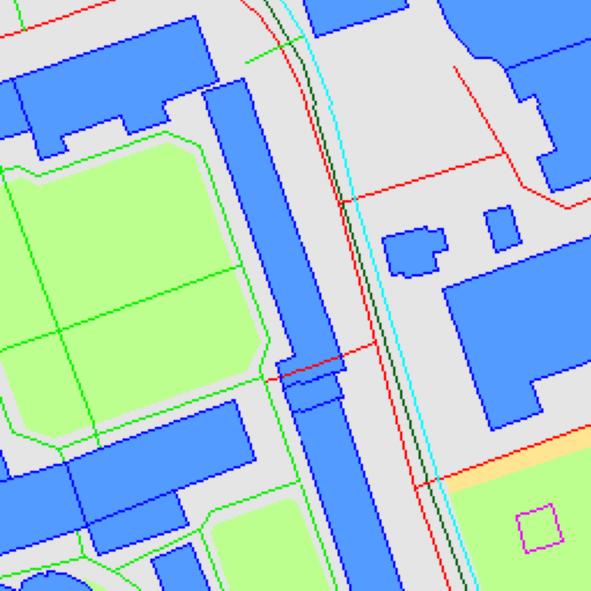}
    &
    & \includegraphics[valign=m,height=\plotH]{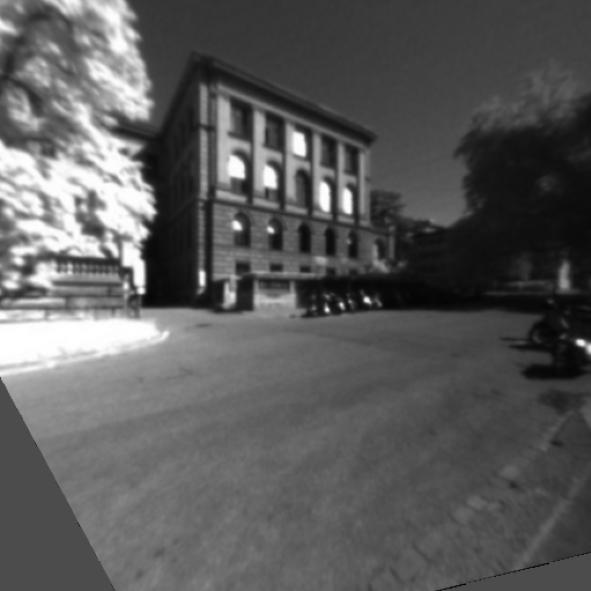}
    & \includegraphics[valign=m,height=\plotH]{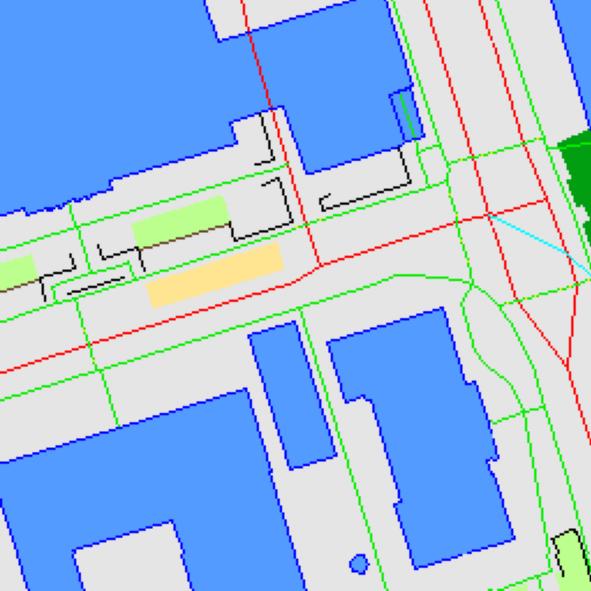} 
    \\
    OrienterNet \cite{sarlin2023orienternet}
    & \includegraphics[valign=m,height=\bevH]{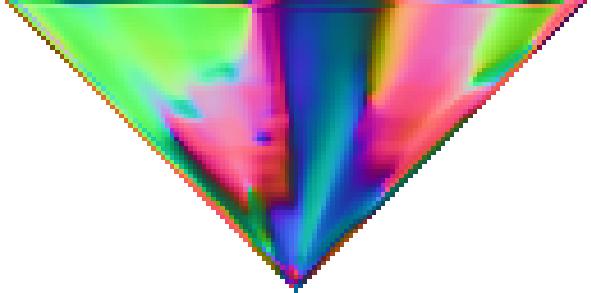} 
    & \includegraphics[valign=m,height=\plotH]{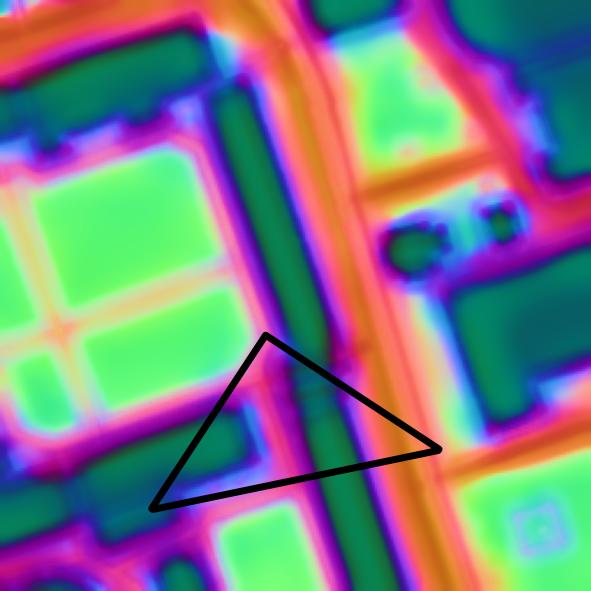} 
    & \includegraphics[valign=m,height=\plotH]{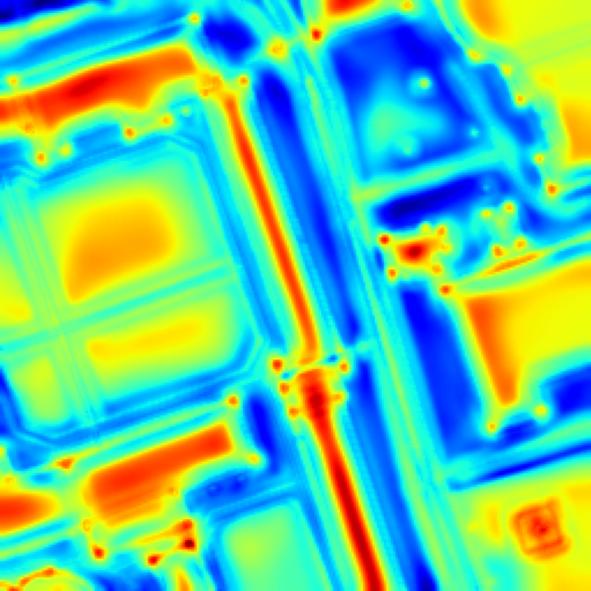} 
    & \includegraphics[valign=m,height=\bevH]{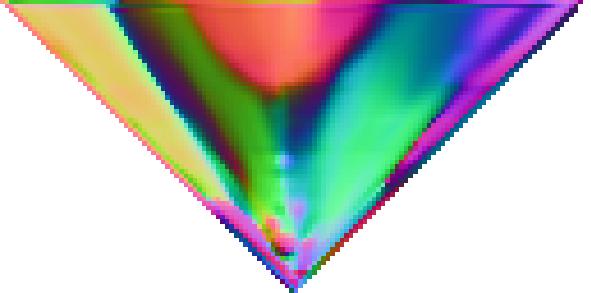} 
    & \includegraphics[valign=m,height=\plotH]{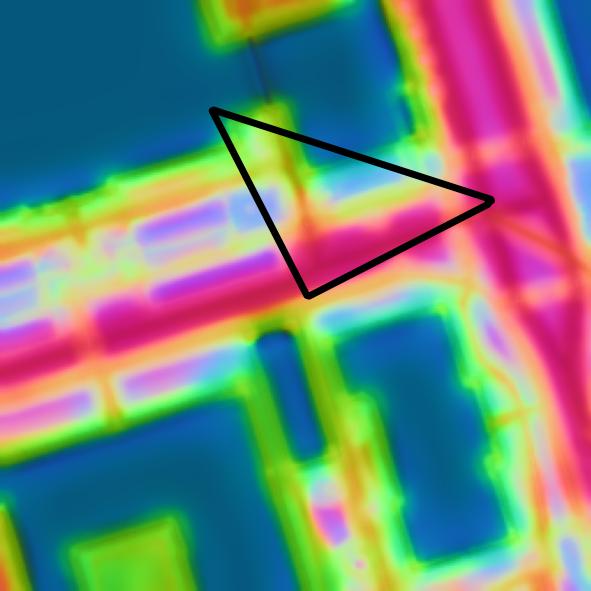} 
    & \includegraphics[valign=m,height=\plotH]{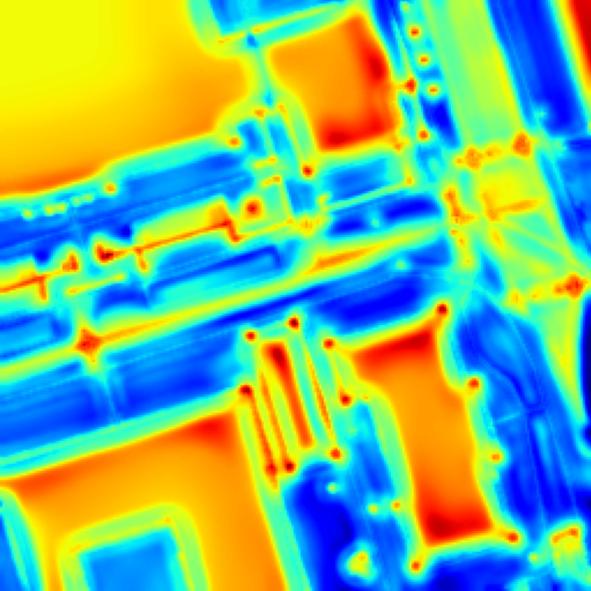} 
    \\
    Chunk$_{r=5}$
    & \includegraphics[valign=m,height=\bevH]{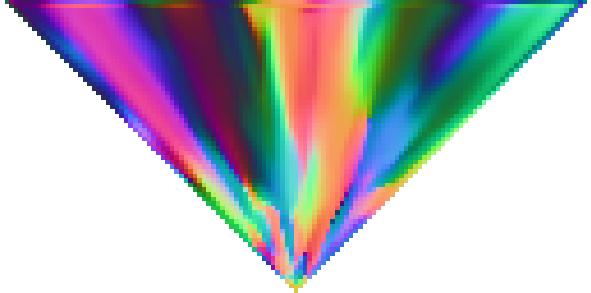} 
    & \includegraphics[valign=m,height=\plotH]{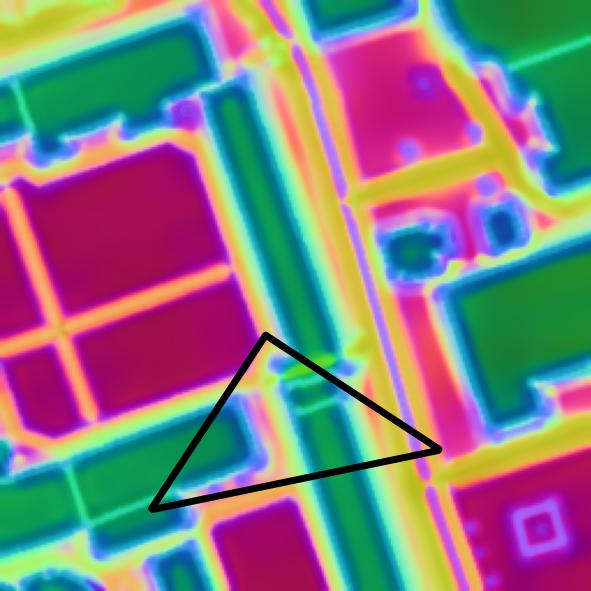} 
    & \includegraphics[valign=m,height=\plotH]{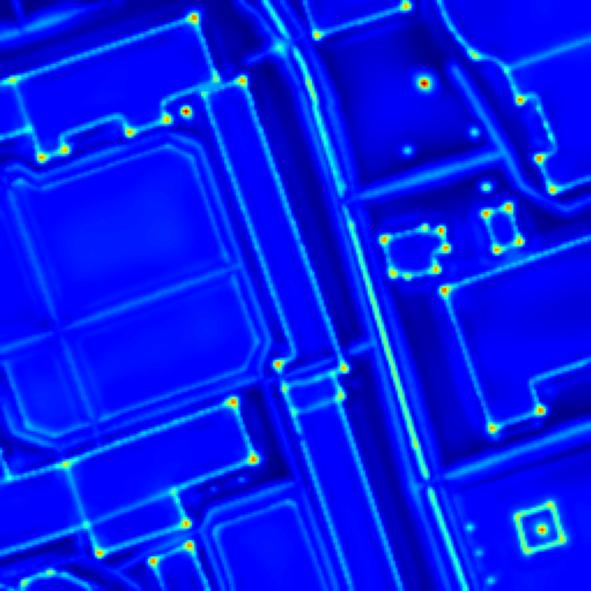}
    & \includegraphics[valign=m,height=\bevH]{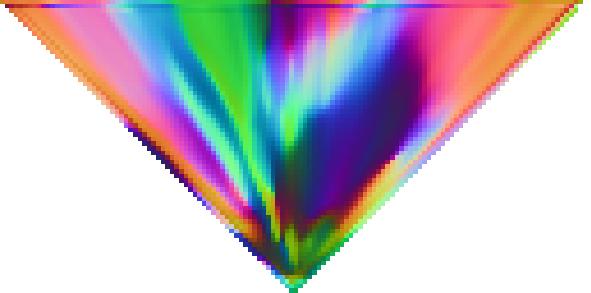} 
    & \includegraphics[valign=m,height=\plotH]{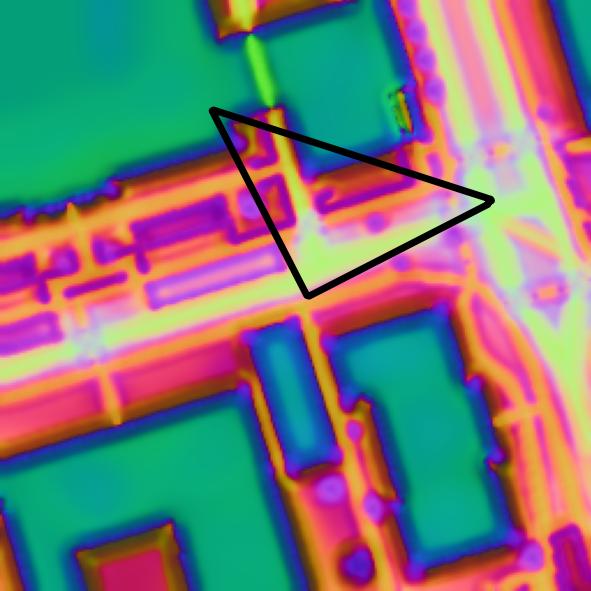} 
    & \includegraphics[valign=m,height=\plotH]{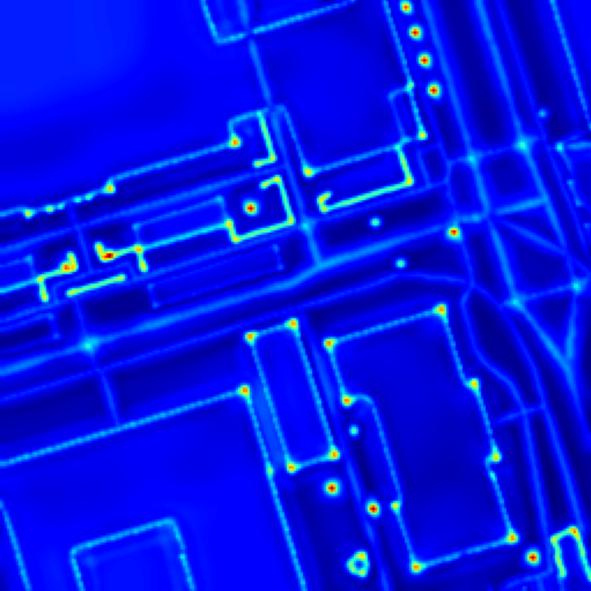} 
    \\ 
    Chunk$_{r=5}$ $+$ $\Delta\Theta$
    & \includegraphics[valign=m,height=\bevH]{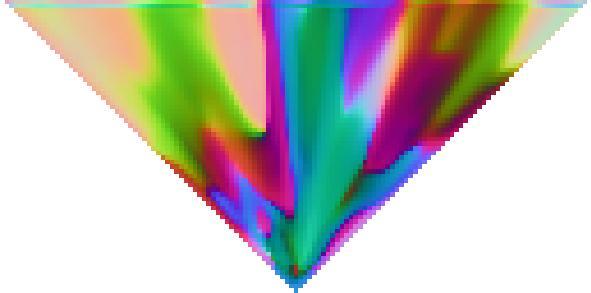} 
    & \includegraphics[valign=m,height=\plotH]{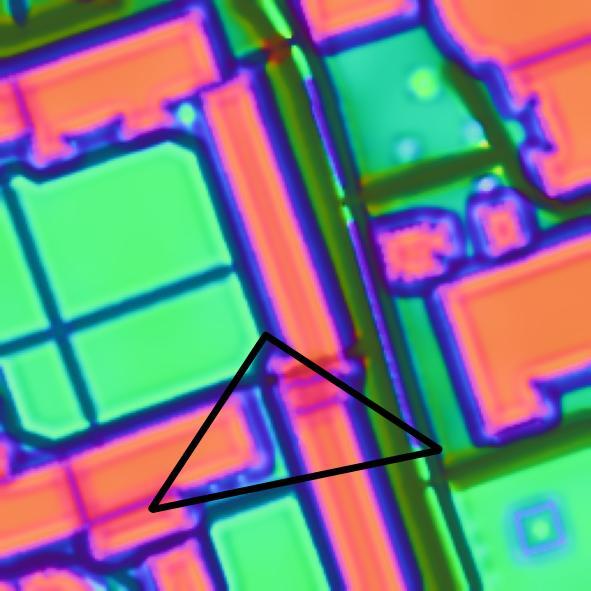} 
    & \includegraphics[valign=m,height=\plotH]{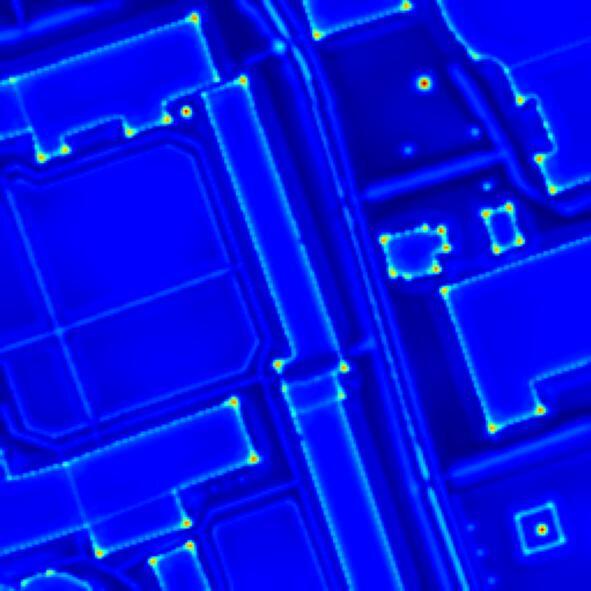}
    & \includegraphics[valign=m,height=\bevH]{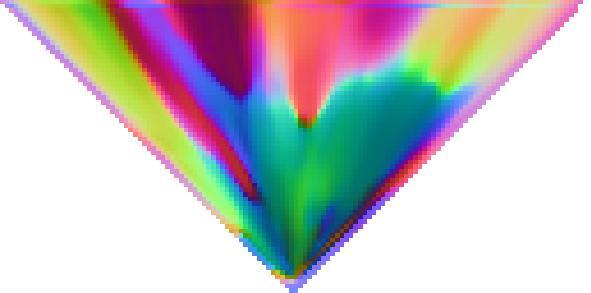} 
    & \includegraphics[valign=m,height=\plotH]{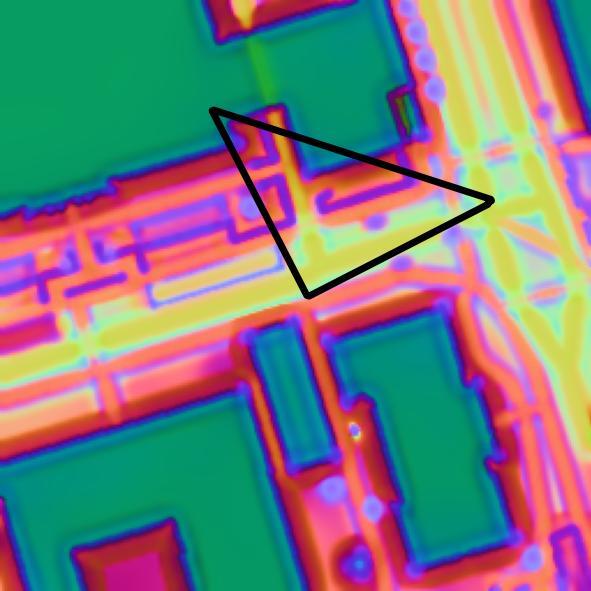} 
    & \includegraphics[valign=m,height=\plotH]{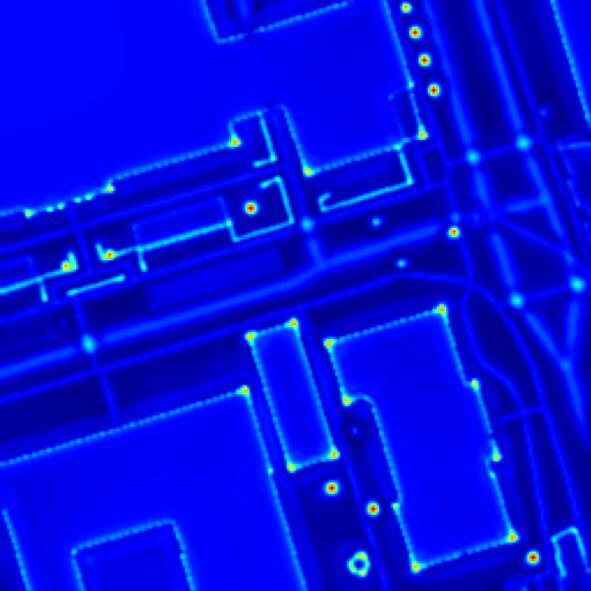} 
    \\
    $\Delta\Theta+\normt_{\rdt=5}$
    & \includegraphics[valign=m,height=\bevH]{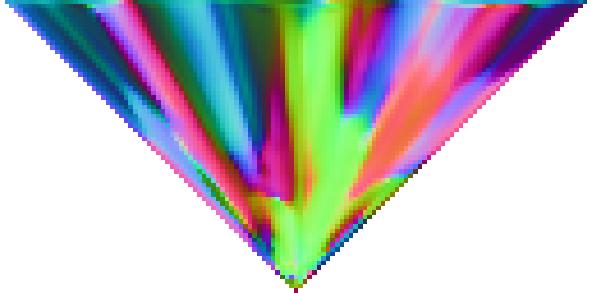} 
    & \includegraphics[valign=m,height=\plotH]{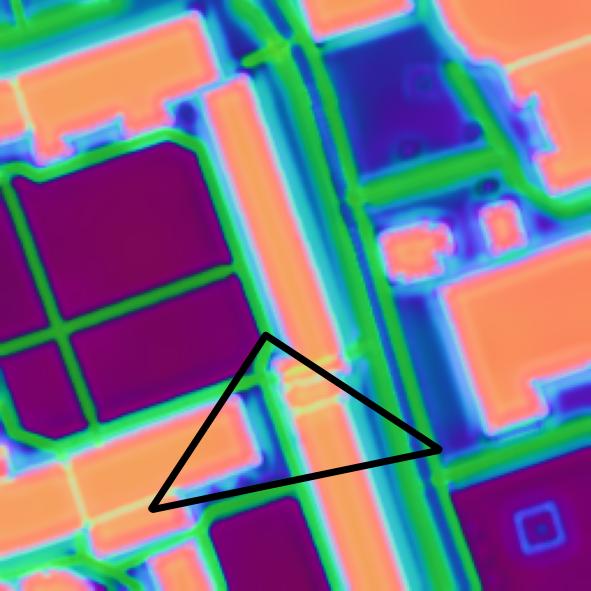} 
    & \includegraphics[valign=m,height=\plotH]{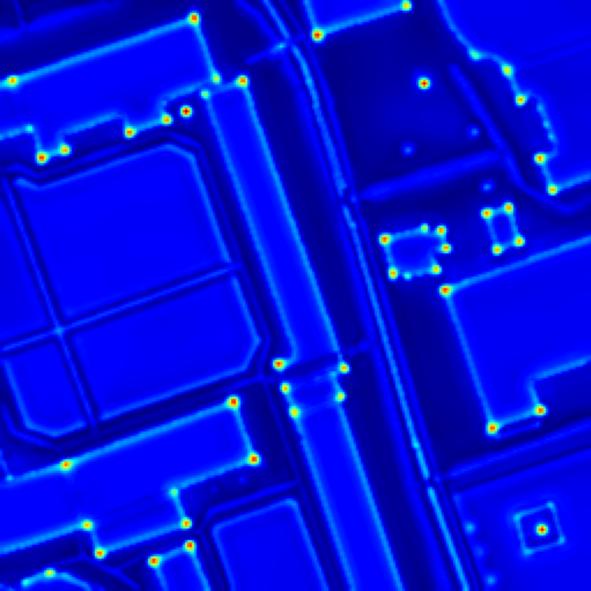}
    & \includegraphics[valign=m,height=\bevH]{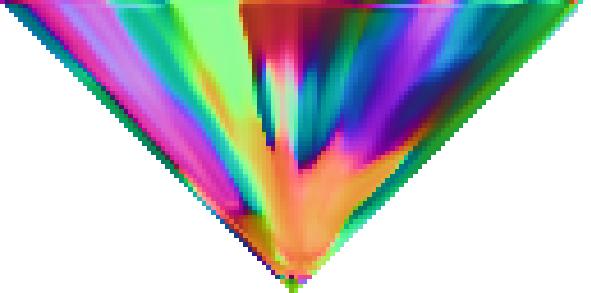} 
    & \includegraphics[valign=m,height=\plotH]{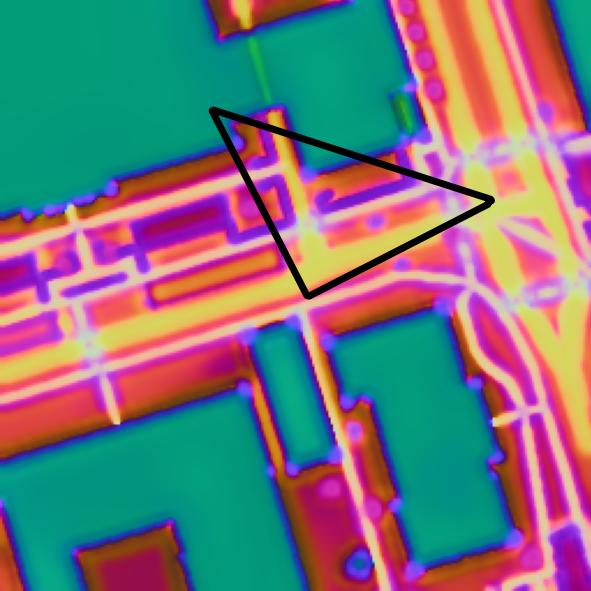} 
    & \includegraphics[valign=m,height=\plotH]{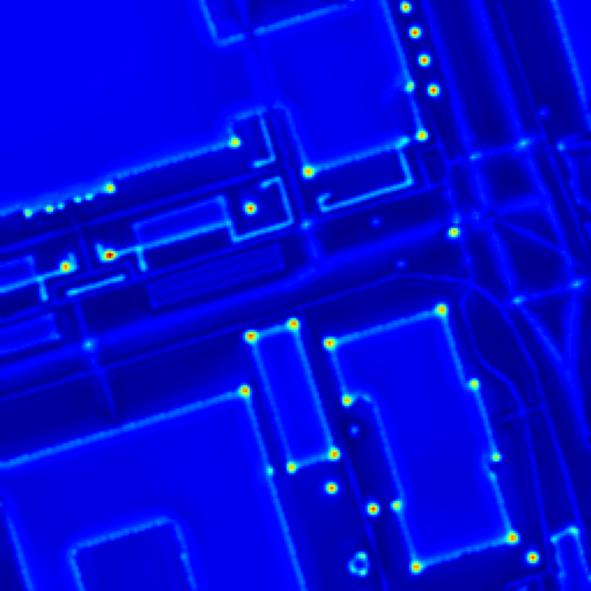} 
    \\ 
    $\Delta\Theta+\Delta XY$
    & \includegraphics[valign=m,height=\bevH]{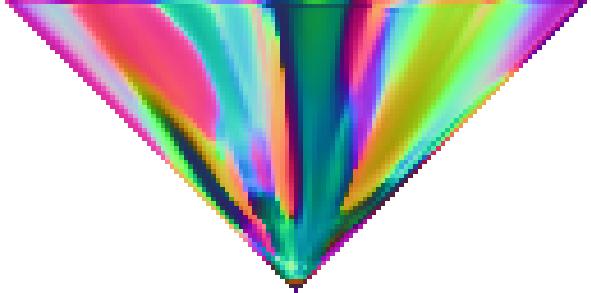} 
    & \includegraphics[valign=m,height=\plotH]{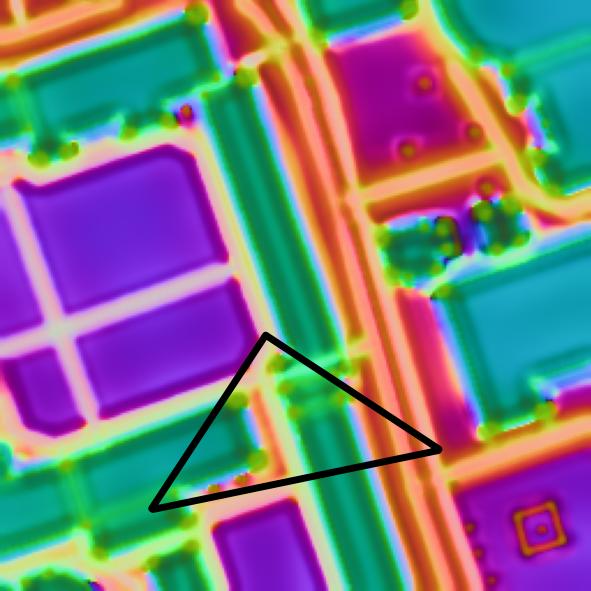} 
    & \includegraphics[valign=m,height=\plotH]{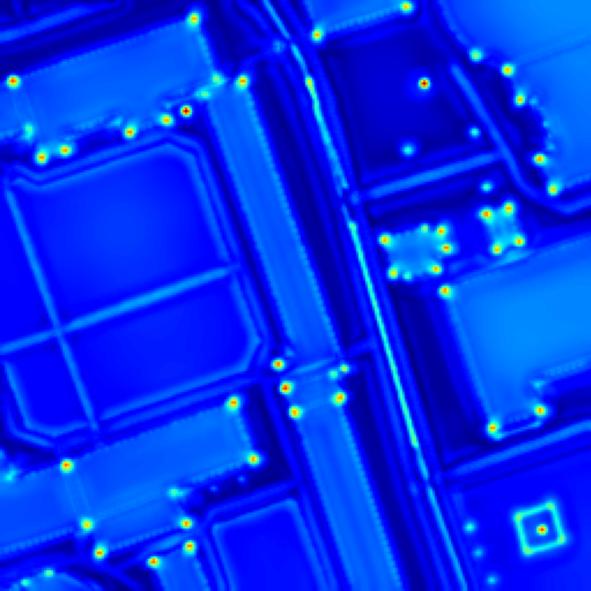}
    & \includegraphics[valign=m,height=\bevH]{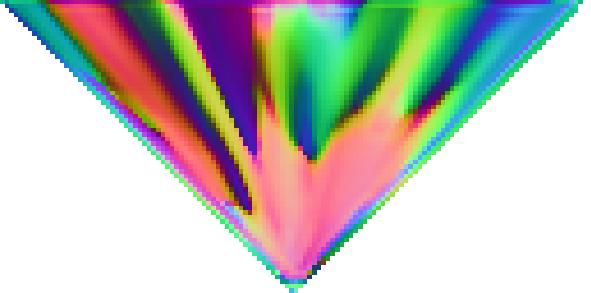} 
    & \includegraphics[valign=m,height=\plotH]{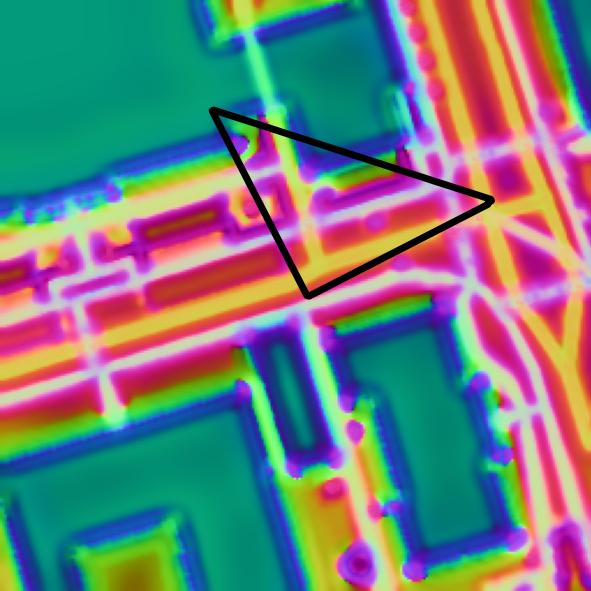} 
    & \includegraphics[valign=m,height=\plotH]{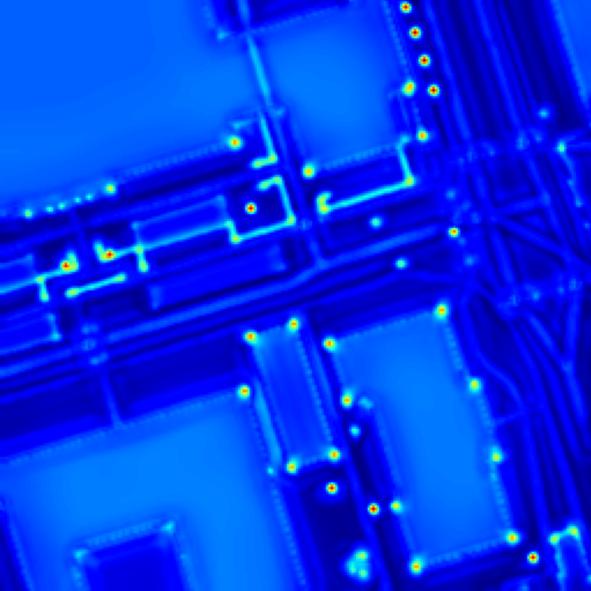} 
    \\
    \end{tblr}
    }
    \caption{\textbf{Qualitative results across supervision types.}
    We show example BEV and map features, as well as their norms, obtained with our different weak supervisions.
    The behavior is consistent across them: we obtain sharper map features than those of strongly supervised baselines \cite{sarlin2023orienternet,liao2024osmloc}, becoming even sharper when using relative poses. This suggests that weak labels, either relative poses or just assuming that the pose lies anywhere within a GPS neighborhood, are less noisy than absolute ones.}
    \label{fig:supp_qual_ckpts}
\end{figure}

\section{Additional qualitative results}\label{sec:supp_qualitative}

\PAR{Comparison across methods}
\cref{fig:supp_qual_oxford,fig:supp_qual_lamaria,fig:supp_qual_kitti} show qualitative results and comparisons with OrienterNet \cite{sarlin2023orienternet} and OSMLoc \cite{liao2024osmloc} across LaMAria \cite{Krishnan2025lamaria}, Oxford Day-and-Night \cite{wang2025oxford} and KITTI \cite{geiger2012kitti}. For the visualizations we use our version of \name trained with relative pose supervision ($\Delta\Theta+\Delta XY$ on \cref{tab:supp_ckpt_summary}). 

\name learns sharper map features, where even small alleys and narrow roads can be distinguished. The norm of these features is proportional to the magnitude of the cross-correlation scores, which indicates that \name learns to focus on distinctive keypoints visible in the input images, such as building corners and road intersections. \name also learns to ignore uninformative regions for fine-grained localization, such as large bodies of water. This contrasts with strongly supervised baselines \cite{sarlin2023orienternet,liao2024osmloc}, which learn smoother feature maps, arguably due to inaccuracies in the ground-truth labels, and instead focus on larger areas that are mostly useful for coarse localization.

\PAR{Comparison across types of weak supervision}
\cref{fig:supp_qual_ckpts} compares results across our different proposed weak supervision types, confirming that the qualitative behavior described above is consistent across supervision types.

\clearpage

\end{document}